\documentclass[11pt]{article}
\usepackage[letterpaper,margin=1in]{geometry}
\usepackage[T1]{fontenc}
\usepackage[expansion=false,protrusion=false]{microtype}
\usepackage[table]{xcolor}
\usepackage{tikz}
\usepackage{eso-pic}
\usepackage{graphicx}
\usepackage{subcaption}
\usepackage{xurl}
\usepackage[
  colorlinks=true,
  linkcolor=black,
  citecolor=black,
  urlcolor=blue
]{hyperref}
\usepackage{cleveref}
\usepackage{pgfplots}
\pgfplotsset{compat=1.18}
\newcommand{\hf}[2]{\href{https://huggingface.co/datasets/#1}{\texttt{#2}}}

\newcommand{\webdataset}[2]{\href{#1}{\texttt{#2}}}
\definecolor{soofipurple}{RGB}{231,206,246}  %
\definecolor{archgreen}{RGB}{212,237,218}    %
\definecolor{osrcamber}{RGB}{249,230,201}    %
\definecolor{osrcblue}{rgb}{0.82,0.91,0.98}%
\definecolor{openwt}{RGB}{236,237,239}       %
\usepackage{booktabs}
\usepackage{longtable}
\usepackage{array}
\usepackage{siunitx}   %
\usepackage{pdflscape} %
\usepackage{placeins}
     \newcommand{\ourmodel}{Soofi S}
\newcommand{\numbaselines}{15}   %
\newcommand{\nummodels}{16}      %
\begin{document}
\noindent\rule{\textwidth}{0.6pt}
\vspace{2.2em}
\begin{center}
  \includegraphics[width=3.5in]{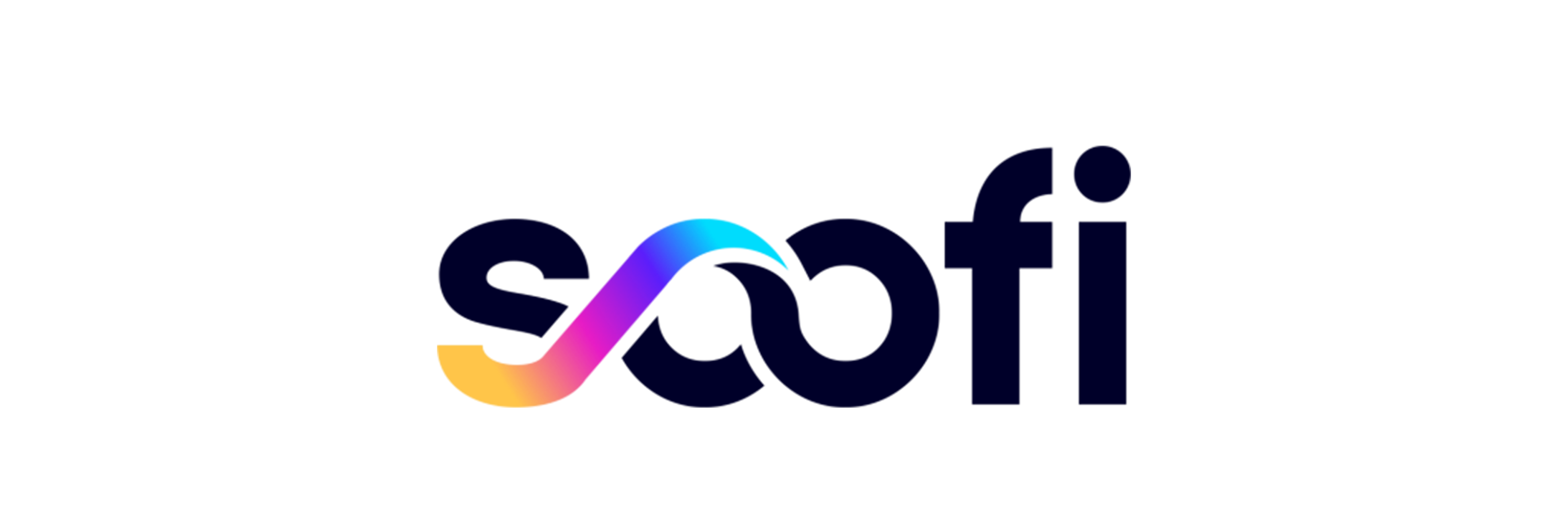}
\end{center}
\vspace{1.6em}
{\fontsize{20}{24}\selectfont\scshape
A Sovereign, Open-Source
Foundation Model for German and English\par}
\vspace{0.9em}
{\scshape\large Soofi S Pretraining Report v1.0\par}
\vspace{2.4em}
{\large\bfseries The Soofi-Team\textsuperscript{*}}\par
\vspace{1.4em}
{\bfseries Core Team:} Benedikt Droste\textsuperscript{10}, David Fitzek\textsuperscript{3,9}, Ruben H{\"a}rle\textsuperscript{5}, Lukas Helff\textsuperscript{2,5,11}, Maximilian Idahl\textsuperscript{10}, Alex Jude\textsuperscript{3,9}, Abbas Goher Khan\textsuperscript{3}, Maurice Kraus\textsuperscript{5}, Timm Ruland\textsuperscript{3,9}, Richard Rutmann\textsuperscript{3,9}, \\Sebastian~Sztwiertnia\textsuperscript{5}
\par\vspace{1.1em}
{\bfseries Contributors:} Markus Frey\textsuperscript{3,9}, Daniil Gurgurov\textsuperscript{2}, Jan Pfister\textsuperscript{6}, Tom Röhr\textsuperscript{7},
Sebastian von Rohrscheidt\textsuperscript{7}

\par\vspace{1.1em}
{\bfseries Advisors:} Jörg Bienert\textsuperscript{1}, Nicolas Flores-Herr\textsuperscript{3}, Simon Gottschalk\textsuperscript{8}, Andreas Hotho\textsuperscript{6}, Kristian Kersting\textsuperscript{2,5,11}, Joachim Köhler\textsuperscript{3}, Alexander Löser\textsuperscript{7}, Wolfgang Nejdl\textsuperscript{8}, Simon Ostermann\textsuperscript{2}, Jan Plogsties\textsuperscript{4}, Björn Plüster\textsuperscript{10}, Patrick Putzky\textsuperscript{12}
\par\vspace{1.1em}
{\bfseries Technical Leads:} Mehdi Ali\textsuperscript{3,9}, Michael Fromm\textsuperscript{3,9}, Max Lübbering\textsuperscript{3,9}

\vspace{2.4em}
{\bfseries Affiliations:}
\textsuperscript{1}KI Bundesverband,
\textsuperscript{2}DFKI,
\textsuperscript{3}Fraunhofer IAIS,
\textsuperscript{4}Fraunhofer IIS,
\textsuperscript{5}Technische Universität Darmstadt,
\textsuperscript{6}Universität Würzburg,
\textsuperscript{7}Berliner Hochschule für Technik,
\textsuperscript{8}L3S Research Center,
\textsuperscript{9}Lamarr,
\textsuperscript{10}ellamind,
\textsuperscript{11}{hessian.AI},
\textsuperscript{12}{Merantix Momentum}
\par\vspace{1.0em}
{\bfseries Coordination \& Funding:} Consortium coordinated by the KI Bundesverband.
Funded by the German Federal Ministry for Economic Affairs and Energy (BMWE).
\vfill
\noindent\rule{2in}{0.4pt}\par
\vspace{0.4em}
{\footnotesize\textsuperscript{$*$}Authors are listed alphabetically. Detailed Contributions in Appendix \ref{app:contributions}.}
\vspace{1.2em}
\begin{center}1\end{center}
\clearpage
\pagestyle{plain}
\setcounter{page}{2}
\setlength{\parskip}{0.6em}
{\large\bfseries Abstract}\par
\vspace{0.4em}
We present Soofi S 30B-A3B, a sovereign, open-source Mixture-of-Experts (MoE) hybrid Mamba Transformer foundation model for German and English. Its hybrid design activates only 3B of 30B parameters per token and keeps the inference cache near-constant as context grows, giving it a decisive throughput advantage over dense models for long-context, high-concurrency deployment. %
Pretrained on roughly 27 trillion tokens with deliberately up-weighted German, Soofi S matches dense 14 to 27B models on aggregate English and German benchmarks while achieving the best code aggregates in both languages among \nummodels{} open base models, and outperforms every European sovereign baseline in our comparison, including ones far larger in active parameters. Among fully open models, \ourmodel{} obtains the highest English and German evaluation scores, ahead of Olmo~3~32B and Apertus~70B. Soofi S was built end-to-end on the German Industrial AI Cloud, a sovereign HPC-scale AI infrastructure operated by Deutsche Telekom in Munich. Soofi S will be released under highly permissive, open-access terms: weights, selected intermediate checkpoints\footnote{\url{https://huggingface.co/Soofi-Project}}, full per-source data accounting, hyperparameters, and training and evaluation code. Where source licenses permit, data-construction artifacts are released under permissive licenses; commercially
licensed sources are documented with aggregate statistics and exact mixture
accounting.
\vspace{1.0em}
\section{Introduction}

\begin{figure}[ht]
  \centering
  \makebox[\textwidth][c]{\includegraphics[width=6.5in]{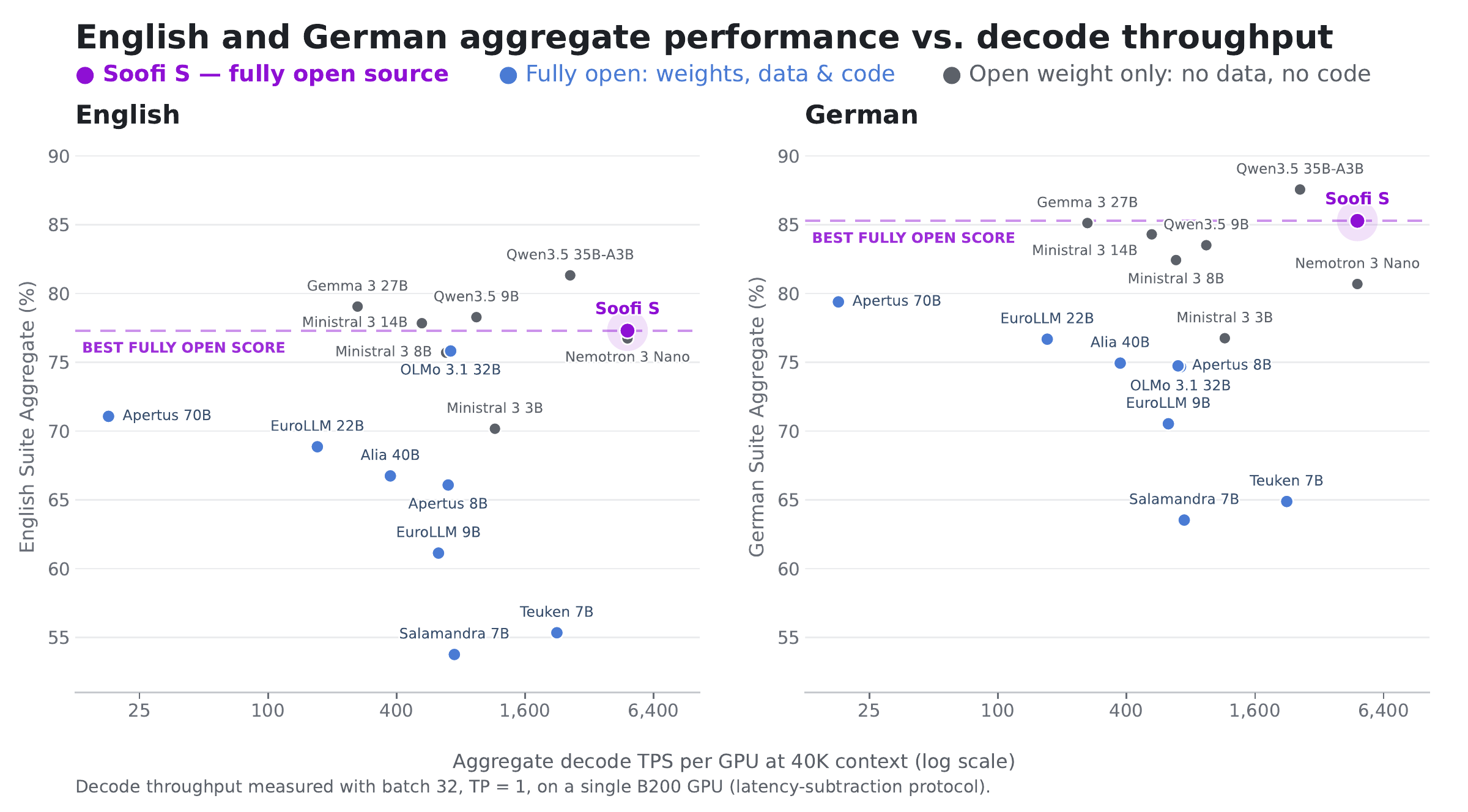}}
\caption{\textbf{English and German aggregate performance versus decode
    throughput.}
    \ourmodel{} obtains the highest English and German aggregate scores
    among the fully open base models in this comparison. \ourmodel{} and
    Nemotron~3~Nano, which share the same hybrid Mamba--MoE architecture,
    reach the highest measured decode throughput at 4.8k TPS/GPU.
    The left panel reports the mean of our English benchmark suite.
    The right panel reports the mean of our German benchmark suite.
    Both panels share a common vertical scale so the two suites can be
    compared directly.
    Both aggregates exclude GPQA and the withdrawn held-out benchmark group (see \Cref{subsec:contamination}).
    The horizontal axis shows measured aggregate decode throughput per GPU
    at a 40K-token context length. Higher evaluation scores and greater
    throughput are better.}
  \label{fig:serving-efficiency}
  \end{figure}

Open language models have improved at remarkable speed, yet three gaps remain
conspicuous for anyone deciding what to actually deploy. 

The first is openness in substance rather than name: despite a proliferation of capable models, the majority of releases remain weight-only releases, documenting their training with little more than
an aggregate token count and omitting the data, recipes, and decisions needed
to reproduce or audit them. 

The second is language: general-purpose
multilingual models are either English-centric or spread their capacity thinly
across dozens of languages, leaving German underrepresented relative to its economic
and scientific weight. Dedicated European efforts to
date~\cite{ali2024teuken,gonzalezagirre2025salamandra,martins2024eurollm,hernandezcano2025apertus}
have prioritized openness and language coverage over frontier capability.

The third gap is the one that most directly governs the deployment cost, and where our chosen architecture is aimed. At economic concurrency the price of
generation is set not by how many parameters a model nominally contains, nor
even by how many it activates per token, but by memory bandwidth: every decoded
token must re-read the model weights and, for a Transformer, the attention cache
of every sequence in the batch. As contexts grow into the tens or hundreds of
thousands of tokens and many requests are served in parallel, this key--value
(KV) cache comes to dominate, and full-attention dense models slow down
accordingly. A model that keeps its per-sequence state small and near-constant
in context length therefore enjoys a structural advantage that compounds in
exactly the regime \texttt{long context, high concurrency} that matters most in
production.

\ourmodel{} 30B-A3B addresses all three at once: It is a Mixture-of-Experts (MoE)
hybrid Mamba Transformer~\cite{nvidia2025nemotron3,nvidia2025nemotronh} trained
to excel in both German and English and will be released radically open: not weights alone,
but the full set of artifacts required to audit every stage of training and,
where source licenses permit, rebuild the data mixture, in the spirit of recent fully open
efforts~\cite{hernandezcano2025apertus,olmo3_2025}. Architecturally it adopts
the openly published Nemotron~3~Nano reference
design~\cite{nvidia2025nemotron3} (\Cref{sec:architecture}): Mamba-2
layers~\cite{dao2024mamba2} carry most of the sequence mixing with a fixed-size
recurrent state, only 6 of its 52 layers maintain a KV cache, and sparse MoE
layers activate just 3.2 of 31.6 billion parameters per token---the capacity of
a 30B network at roughly the inference cost of a 3B one.
 
\paragraph{Contributions.} We summarize our contributions below:
\begin{itemize}
        \item \textbf{Strongest fully open German--English model.} \ourmodel{} is the strongest fully open model in our evaluation on English and German benchmarks, and matches or outperforms every European sovereign baseline in our comparison on every German benchmark in our suite, while matching dense 14--27B international models on English and German aggregate performance at a fraction of their active-parameter cost (\Cref{sec:base-evals}).
    \item \textbf{Full data transparency.} We release the complete pretraining
          corpus statistics (\Cref{tab:phase1-sources}, \Cref{tab:phase2-sources}, \Cref{tab:lc-budget}) and reproducible construction scripts\footnote{\label{note1}\url{https://github.com/soofi-project/Soofi-Pretraining}} with \emph{per-source and per-language token accounting}, distinguishing dataset-card estimates from tokenizer-exact consumed-token counts. We also provide the
          German\,:\,English\,:\,code mixing ratio, and the rationale behind
          it, in contrast to reports that disclose only an aggregate token
          count.
    \item \textbf{Reproducible recipe.} We publish the full learning-rate
          schedule (Warmup--Stable--Decay), optimizer, all hyperparameters, the
          per-phase token budgets, and the phase boundaries, so a third party
          can rebuild the run.
    \item \textbf{Long-context serving efficiency.} The hybrid
          Mamba--MoE design keeps the per-sequence cache near-constant in
          context length, yielding measured aggregate decode TPS/GPU
          $8$--$9\times$ that of dense 14--27B models at 40K context and batch
          32, and aggregate decode TPS that stays essentially flat from 4K to 256K where
          full-attention models degrade (\Cref{fig:serving-throughput-scaling}, \Cref{subsec:efficiency}).
    \item \textbf{Documented design.} We report the data and design
          ablations behind each choice, exposing the \emph{why} rather than only
          the final configuration.
    \item \textbf{Complete artifact release.} We will release selected \ourmodel{} 30B-A3B Base checkpoints\footnote{\url{https://huggingface.co/Soofi-Project}}, training\footnote{\url{https://github.com/soofi-project/Soofi-Pretraining}} and evaluation code\footnote{\url{https://github.com/ellamind/base-eval}}\footnote{\url{https://github.com/ellamind/eval-hive}}, all under permissive licenses for transparent audit and
          extension. 
\end{itemize}

The remainder of this report documents \ourmodel{} 30B-A3B Base end to end.
\Cref{sec:architecture} presents the model and the training recipe: the
Nemotron~3~Nano reference architecture we adopt and the rationale for adopting
it, the Warmup--Stable--Decay optimization schedule, all hyperparameters and
per-phase token budgets, the training dynamics of the run, and the compute
infrastructure. \Cref{sec:pretraining-data} details the three-phase,
26.68T-token German--English data curriculum, with full per-source token
accounting for every stage. \Cref{sec:base-evals} then evaluates the resulting
base model against \numbaselines{} open models of comparable or larger active
size along two axes: capability across parallel English and German benchmarks,
and serving efficiency. Section~\ref{sec:related} situates the work relative to
prior open and European efforts, and Section~\ref{sec:conclusion} concludes.

\section{Model Architecture and Training}

\begin{figure}[t]
    \centering
    \includegraphics[width=\linewidth]{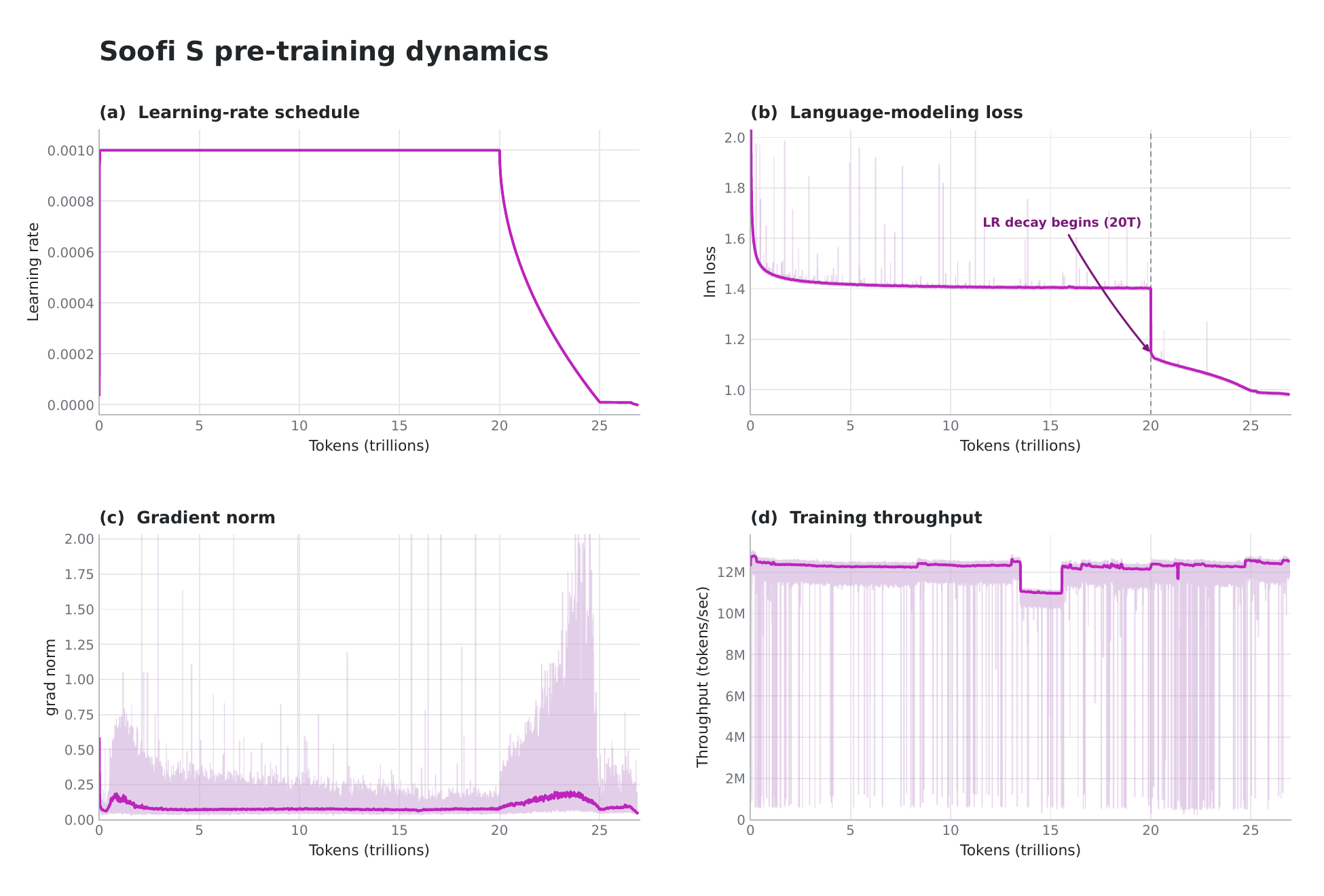}
    \caption[Training dynamics versus consumed tokens.]{%
        \textbf{Training dynamics over the full \({\sim}27\text{T}\)-token run.}
        The quantity is plotted against the number of consumed tokens
        (in trillions); the pretraining-to-annealing transition occurs at
        ${\sim}20\text{T}$ tokens. The solid line is a rolling median over
        $1{,}000$ steps and the faint trace is the raw per-step signal.
    }
    \label{fig:training-dynamics}
\end{figure}

\label{sec:architecture}
\ourmodel{} 30B-A3B Base adopts the hybrid Mamba--Transformer
Mixture-of-Experts (MoE) reference architecture of
Nemotron~3~Nano~\cite{nvidia2025nemotron3,nvidia2025nemotronh}: a 52-layer
network interleaving 23 Mamba-2 sequence-mixing layers~\cite{dao2024mamba2},
23 granular MoE layers with shared
experts~\cite{shazeer2017moe,fedus2022switch,dai2024deepseekmoe,krajewski2024granularity},
and 6 Grouped-Query Attention (GQA) layers~\cite{ainslie2023gqa} distributed
sparsely through the network depth. 

The model totals ${\sim}31.6$B parameters,
of which only ${\sim}3.2$B are active per forward pass (${\sim}3.6$B including
embeddings), and only the 6 GQA layers maintain a KV cache. Because we reuse
the reference design without modification, we refer the reader to the Nemotron
reports~\cite{nvidia2025nemotronh,nvidia2025nemotron3} for the design
motivation and the ablations behind each architectural choice;
Table~\ref{tab:architecture} records the exact configuration for
reproducibility.
\paragraph{Why a reference architecture.}
Reusing an established, openly specified architecture rather than designing a
bespoke one was a deliberate decision, on three grounds. First,
\emph{deployability}: the Nemotron~3~Nano architecture is already integrated
into the major open inference and serving stacks, including the vLLM stack
used for our own serving measurements (\Cref{subsec:efficiency}), with
mature, heavily optimized kernels for its Mamba-2, GQA, and MoE components, so
\ourmodel{} can be hosted efficiently by existing software from the day of
release, without bespoke integration work by downstream users. Second,
\emph{serving efficiency}: the predominantly Mamba-2 backbone makes the
architecture exceptionally fast in exactly the regime we target, prefill and
decode costs grow near-linearly in sequence length and the per-sequence cache
stays near-constant, which is what underpins the long-context throughput
results of \Cref{subsec:efficiency} and makes the 1M-token context extension
of \Cref{sec:longcontext} practical. Third, \emph{scientific control}: sharing
the backbone with Nemotron~3~Nano turns that model into an
architecture-identical baseline, so the effect of our German--English data
recipe can be measured in isolation (\Cref{subsec:eval-openweight}).

\begin{table}[ht]
\centering
\caption{\ourmodel{} 30B-A3B architecture, following the Nemotron~3~Nano
reference configuration~\cite{nvidia2025nemotron3}. The layer pattern
interleaves Mamba-2 and MoE layers, with 6 GQA layers distributed through the
network depth.}
\label{tab:architecture}
\begin{tabular}{lr}
\toprule
\textbf{Hyperparameter} & \textbf{Value} \\
\midrule
Total parameters              & 31.6B \\
Active parameters per token   & 3.2B (3.6B incl.\ embeddings) \\
Num layers                    & 52 (23 Mamba-2, 23 MoE, 6 GQA) \\
Model dimension               & 2688 \\
Attention Q-heads             & 32 \\
Attention KV-heads            & 2 \\
Attention head dimension      & 128 \\
Mamba-2 state dimension       & 128 \\
Mamba-2 groups                & 8 \\
Mamba-2 heads                 & 64 \\
Mamba-2 head dimension        & 64 \\
Expert dimension              & 1856 \\
Routed experts                & 128 \\
Activated experts per token   & 6 \\
Shared experts                & 2 \\
MoE activation                & squared ReLU \\
Router                        & learned MLP, sigmoid gating \\
Positional embeddings         & none \\
Normalization                 & RMSNorm \\
Embedding / projection tying  & untied \\
\bottomrule
\end{tabular}
\end{table}

\subsection{Optimization and Hyperparameters}
\label{sec:hyperparameters}
We train \ourmodel{} with the Megatron-Bridge
framework\footnote{For this training run, we adopted Megatron-Bridge (\url{https://github.com/NVIDIA/Megatron-Bridge}) rather than integrating the Nemotron-3 reference architecture into our open-source stack~\cite{lubbering2026modalities}, as Megatron-Bridge already provided mature, reliable support for the newly released architecture.} using
AdamW~\cite{loshchilov2019adamw} under a Warmup--Stable--Decay
(WSD)~\cite{hu2024minicpm,hagele2024scaling} learning-rate schedule whose decay
segment follows a $\texttt{minus\_sqrt}$ shape. With the exception of the
long-context phase (Section~\ref{sec:phase3-hparams}), all stages share the same
parallelism and batching configuration: tensor-model-parallel (TP) size $1$,
expert-parallel (EP) size 8, sequence parallelism
disabled, micro-batch size $2$, and global batch size $3072$ at a sequence
length of $8192$ tokens, with the remaining GPUs used for data parallelism and a
distributed (optimizer-state-sharded) optimizer. This corresponds to
$25{,}165{,}824$ tokens per optimizer step. All stages are trained in bf16 mixed precision. Since the granular MoE layers
route each token to experts held on different expert-parallel ranks, all-to-all
communication sits on the critical path, which motivates the node topology and
interconnect reported in Section~\ref{sec:compute-infrastructure}. Manual
garbage collection is triggered every $101$ iterations.
The schedule is realised across one warmup-plus-stable phase and three successive
annealing continuations, summarised in Table~\ref{tab:train-stages}; the data
mixture consumed by each stage is documented in
\Cref{sec:pretraining-data}:
\begin{itemize}
    \item \textbf{Base pretraining (stable).} $794{,}728$ iterations
          ($19{,}999{,}984{,}975{,}872$ tokens, ${\sim}20$T) on the 20T EN/DE
          mixture (see \Cref{sec:phase1}). After a $254$-iteration warmup to the peak learning rate of
          $1\mathrm{e}{-}3$, training proceeds on the WSD stable plateau.
    \item \textbf{Main annealing (decay).} A continuation of $198{,}682$
          iterations ($4{,}999{,}996{,}243{,}968$ tokens, ${\sim}5$T) on the 5T
          high-quality EN/DE mixture (see \Cref{sec:phase2}), applying the WSD $\texttt{minus\_sqrt}$ decay
          from $1\mathrm{e}{-}3$ to $1\mathrm{e}{-}5$. The configured total after
          this continuation is $993{,}410$ iterations.
    \item \textbf{Constant annealing.} Since the slope at the end of the decay
          segment remained relatively steep, we append a further $62{,}590$
          iterations ($1{,}575{,}128{,}924{,}160$ tokens (see \Cref{sec:phase2}), ${\sim}1.58$T) at a
          \emph{constant} learning rate of $1\mathrm{e}{-}5$ on the same 5T
          mixture, allowing additional training at the tail of the annealing
          curve.
    \item \textbf{Final annealing (discarded).} A final WSD $\texttt{minus\_sqrt}$
          decay of $11{,}920$ iterations (${\sim}0.30$T, see \Cref{sec:phase2})  from $1\mathrm{e}{-}5$ to $0$, beginning at iteration
          reference $1{,}056{,}000$ and targeting $1{,}067{,}920$. We report this
          stage for completeness but \emph{do not} use its checkpoints, as they
          showed no clear additional benchmark improvement over the constant
          annealing stage.
\end{itemize}
\begin{table}[ht]
\centering
\caption{Training stages and learning-rate schedule. ``Iters'' are the
iterations added in each stage; all $8$K-context stages use $25{,}165{,}824$
tokens per iteration. The final annealing stage was run but its checkpoints were
not used. The long-context stage uses a distinct configuration
(Section~\ref{sec:phase3-hparams}).}
\label{tab:train-stages}
\begin{tabular}{llrrl}
\toprule
\textbf{Stage} & \textbf{Role} & \textbf{Iters} & \textbf{Tokens} & \textbf{LR (peak\,$\to$\,min)} \\
\midrule
Base pretraining   & Stable (warmup $254$)        & 794{,}728 & ${\sim}20$T   & $1\mathrm{e}{-}3$ (plateau) \\
Main annealing     & WSD decay ($\texttt{minus\_sqrt}$) & 198{,}682 & ${\sim}5$T    & $1\mathrm{e}{-}3 \to 1\mathrm{e}{-}5$ \\
Constant annealing & Constant LR                  & 62{,}590  & ${\sim}1.58$T & $1\mathrm{e}{-}5$ \\
Final annealing$^{\dagger}$ & WSD decay ($\texttt{minus\_sqrt}$) & 11{,}920 & ${\sim}0.30$T & $1\mathrm{e}{-}5 \to 0$ \\
\midrule
Long context       & Constant (warmup $100$)      & 2{,}000   & ${\sim}0.10$T & $1\mathrm{e}{-}5 \to 1\mathrm{e}{-}7$ \\
\bottomrule
\multicolumn{5}{l}{\footnotesize $^{\dagger}$ Run for completeness; checkpoints not used (no clear benchmark gain).}\\
\end{tabular}
\end{table}
\subsection{Long-Context Extension}
\label{sec:phase3-hparams}
The long-context phase (Phase~3, Section~\ref{sec:longcontext}) is trained with a
distinct parallelism configuration to accommodate the $1$M-token sequences. We use
a context-parallel size of $16$, a micro-batch size of $1$, and a global batch
size of $48$ at a sequence length of $1{,}048{,}576$ tokens, giving
$50{,}331{,}648$ tokens per optimizer step. The stage runs for $2{,}000$
iterations, for an approximate total of $100.66$B tokens.
Optimization again uses AdamW. The learning rate follows a constant schedule with
a $100$-iteration warmup, holding at $1\mathrm{e}{-}5$ (minimum $1\mathrm{e}{-}7$).
Two settings distinguish this phase from the earlier stages: HybridEP is enabled
(it was not used previously), and we apply \emph{no} intra-document masking,
matching NVIDIA's long-context recipe~\cite{nvidia2025nemotron3} rather than the alternative adopted by some
other efforts.
\subsection{Training Dynamics}
Figure~\ref{fig:training-dynamics} summarizes the four central signals we logged over the
full ${\sim}27$T-token run, each plotted against the number of consumed tokens.
The pretraining-to-annealing transition at ${\sim}20$T tokens---the end of the
stable plateau and the start of the main WSD decay
(Table~\ref{tab:train-stages})---is the reference point for reading all four
panels. For the loss, gradient-norm, and throughput traces, the solid line is a
rolling median over $1{,}000$ steps and the faint trace is the raw per-step
signal; loss and gradient norm are clipped at $2.0$ for readability.
\paragraph{Training logs.}
In addition to the static training-dynamics plot in
Figure~\ref{fig:training-dynamics}, we provide the corresponding Weights \&
Biases dashboard for the full pretraining run\footnote{\url{https://api.wandb.ai/links/soofi-exchange/j11vi7rg}}.
The dashboard contains the raw and smoothed traces used to inspect optimization
stability, throughput, checkpointing effects, and the transitions between the
base-pretraining, annealing, and long-context phases.

The learning-rate schedule (Figure~\ref{fig:training-dynamics}a) follows the
Warmup--Stable--Decay (WSD) shape: a
short linear warmup to the peak of $1\times10^{-3}$, a constant plateau held
throughout base pretraining, and a $\texttt{minus\_sqrt}$ decay toward ${\sim}0$
once annealing begins. The language-modeling loss (Figure~\ref{fig:training-dynamics}b)
declines steadily across the stable phase and then drops sharply at the onset of
annealing due to the dataset switch, tracking the learning-rate decay as the model is concentrated on the
high-quality mixture. The gradient norm (Figure~\ref{fig:training-dynamics}c) remains
stable across the entire run, with only a mild increase during the annealing
phase and no divergence or sustained spikes. Training throughput
(Figure~\ref{fig:training-dynamics}d), measured in tokens per second, is steady for most
of the run; the downward excursions correspond to checkpointing, evaluation, and
restart iterations rather than to changes in the training dynamics themselves.

\subsection{Compute Infrastructure}
\label{sec:compute-infrastructure}
\ourmodel{} was trained on the Industrial AI
Cloud~\cite{telekom2026industrialaicloud,nvidia2025industrialaicloud} operated by
Deutsche Telekom in Munich built together with NVIDIA and brought into operation in February 2026.
Our run used up to 512 NVIDIA B200 GPUs, i.e.\ 64 DGX~B200 nodes
of 8 GPUs each. Within a node the 8 B200s are fully connected by
fifth-generation NVLink/NVSwitch, while nodes are interconnected by an eight-rail NVIDIA Quantum-2 NDR InfiniBand fabric, one 400\,Gb/s ConnectX-7 port per GPU, $3.2$\,Tb/s of scale-out bandwidth per node.
Keeping expert-parallel groups within a node lets the MoE all-to-all
(Section~\ref{sec:hyperparameters}) run over intra-node NVLink rather than the
slower inter-node fabric, which matters for MoE throughput at this scale. The
run took place from 24 March 2026 to 13 May 2026 and consumed approximately
253{,}000 B200 GPU-hours across the stable, annealing, and long-context stages.
Training on this infrastructure is part of the sovereignty of the model: it was
executed on German soil under European operational and data-protection
requirements rather than on extra-European hyperscale compute. \ourmodel{} was
one of the first flagship workloads on the Industrial AI
Cloud, whose infrastructure was procured for the sovereign open-source
foundation-model effort under which this model was
developed~\cite{telekom2026industrialaicloud}. The Munich facility is powered
entirely by renewable energy, designed for high energy efficiency, cooled with
water drawn from the nearby Eisbach canal, and integrated with a waste-heat-reuse
concept that feeds the surrounding Tucherpark district.

\section{Pretraining Data}
\label{sec:pretraining-data}
\begin{figure}[ht]
\centering
\includegraphics[width=\textwidth]{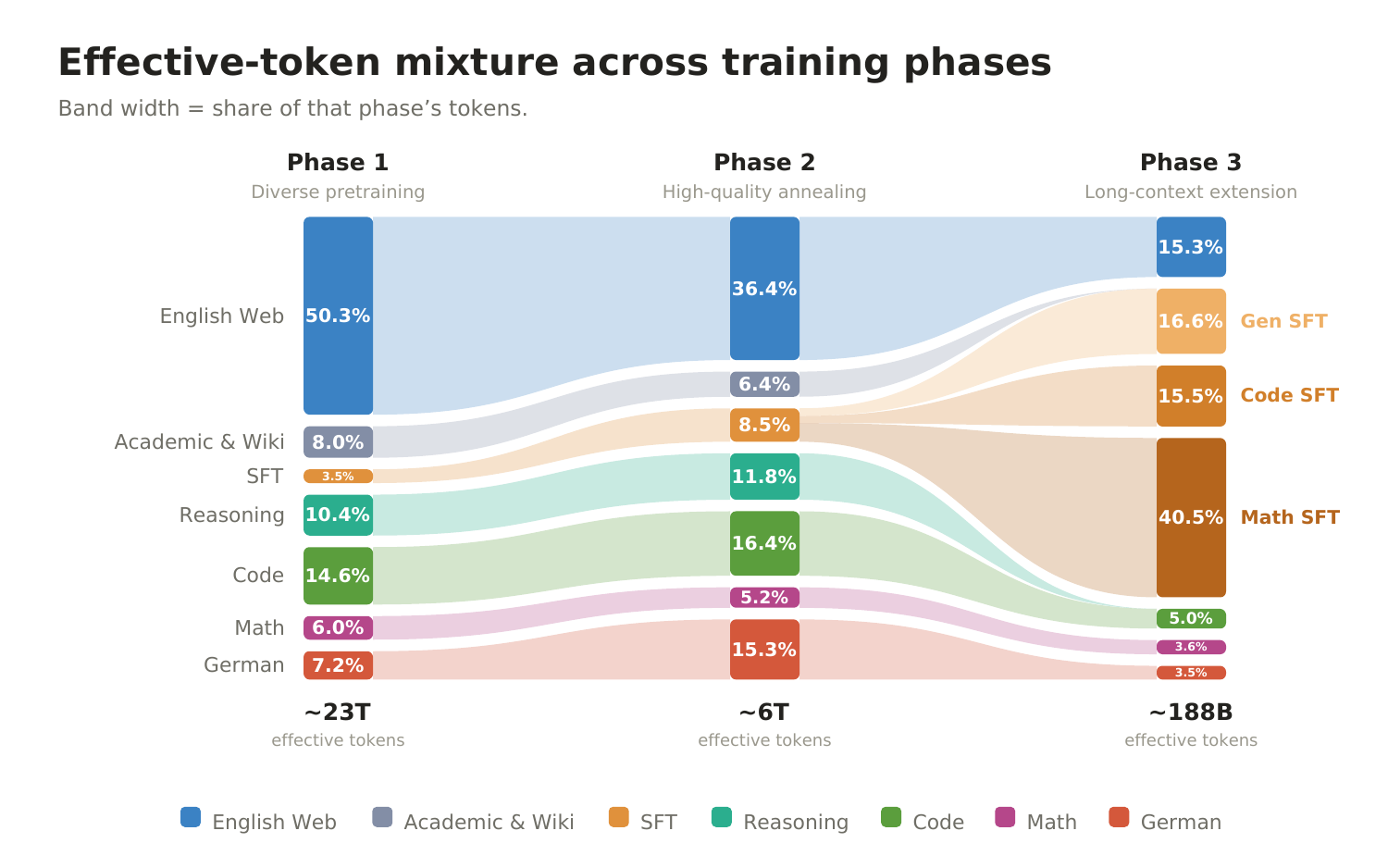}
\caption{\textbf{Effective-token mixture across the three training phases}. A single
flow diagram tracing seven data categories (English Web, Academic \& Wiki, SFT,
Reasoning, Code, Math, and German) from left to right across the phases. \textbf{Phase~1} (diverse pretraining),
$23{,}051.13$B effective tokens. \textbf{Phase~2} (high-quality annealing),
$6{,}303.0$B effective tokens, showing increased density of skill-oriented and
German data relative to Phase~1. \textbf{Phase~3} (long-context extension), $188$B
effective tokens, where the SFT band branches into its General, Code, and Math SFT
components.}
\label{fig:token-allocation}
\end{figure}

Consistent with our commitment to full reproducibility, we document the
pretraining corpus of \ourmodel{} at the granularity of individual source
datasets. For every constituent, we report its public identifier, its raw token
count, the number of epochs it was repeated, the resulting effective token
count, and its share of the phase. We deliberately also list sources that were
enumerated but \emph{excluded} from training (epoch count of zero), so that the
mixture can be audited and rebuilt end to end. This stands in contrast to the
common practice of disclosing only an aggregate token count for the training
data, and follows the openness ethos of recent fully open efforts~\cite{hernandezcano2025apertus,olmo3_2025}.

\ourmodel{} is trained on a three-phase curriculum, consistent with the
Warmup--Stable--Decay (WSD) learning-rate schedule
(\Cref{sec:hyperparameters}). Phase~1 (see \Cref{sec:phase1}) maximizes diversity over a large, quality-tiered mixture of web, synthetic, code, mathematics, and multilingual data. Phase~2 (\Cref{sec:phase2}) is an annealing (decay) phase that concentrates the highest-quality web data together with skill-focused code, mathematics, STEM,
reasoning, and instruction data, while further up-weighting German to
$15.3\%$ of the constructed annealing pool. Phase~3 (\Cref{sec:longcontext})
extends the usable context length up to 1M tokens via length-bucketed
up-sampling. Table~\ref{tab:phase-budget} summarizes the token budget of each
phase. Across all phases, the corpus comprises approximately 27 trillion tokens,
of which a deliberately elevated fraction is German, reflecting the design goal
of a German--English model rather than a broadly multilingual one. The full
per-source composition of every phase is documented in
Appendix~\ref{app:data-composition}; the figure in this section summarizes those tables as a mixture flow diagram.

\begin{table}[ht]
\centering
\caption{Three-phase pretraining curriculum and token budget. The
\emph{Pool} column is the effective-token mixture constructed for each phase
(documented per source in Appendix~\ref{app:data-composition}); for Phases~1--2
these are dataset-card counts and are approximate, whereas the Phase~3 pool is
tokenized with our own tokenizer. The \emph{Consumed} column is the
number of tokens actually trained on, counted exactly from the optimizer
schedule (iterations $\times$ tokens/iteration; Table~\ref{tab:train-stages}).
A phase may consume less than one epoch of its pool (Phase~1) or more than one
(Phase~2); the headline ${\sim}27$T figure refers to consumed tokens.}
\label{tab:phase-budget}
\begin{tabular}{llrr}
\toprule
\textbf{Phase} & \textbf{Role} & \textbf{Pool (eff.)} & \textbf{Consumed} \\
\midrule
Phase 1 & Diverse pretraining (stable)    & $\sim$23.05T & $\sim$20T \\
Phase 2 & High-quality annealing (decay)  & $\sim$6.30T  & $\sim$6.58T \\
Phase 3 & Long-context extension (4K--1M) & $\sim$0.19T  & $\sim$0.10T \\
\midrule
\textbf{Total} & & $\sim$29.5T & $\sim$26.68T \\
\bottomrule
\end{tabular}
\end{table}

\subsection{Quality Tiers, Synthetic Data, and Epoching}
\label{sec:quality-tiers}

Most of the web data is drawn from the openly released Nemotron-CC datasets~\cite{su2024nemotroncc,nvidia2025nemotron3}
(v1.0, v2.0, and v2.1), which provide documents pre-sorted into quality tiers
(High, Medium-High, Medium) and include several synthetically rephrased variants
(\textsc{-Synthetic}), diverse question--answer reformulations
(\textsc{Diverse QA} / \textsc{DQA}), and English translations of non-English
documents (\textsc{Translated-To-English}). We exploit these tiers directly: the highest-quality and synthetic tiers are repeated for multiple epochs to increase
their effective contribution, whereas the large Medium-Quality pools are listed
but set to zero epochs and thus excluded from the final mixture. An epoch count greater than one, therefore, denotes deliberate up-sampling of a high-value source, and an epoch count of zero denotes a source we evaluated but chose not to train
on. All effective-token figures and shares below are reported \emph{after}
applying these epoch multipliers.

\paragraph{Token accounting and its precision.}
For Phases~1 and~2, the raw and effective token counts below are taken from the
token statistics reported on each source's public (HuggingFace) dataset card.
Since different datasets are tokenized with different tokenizers, these counts
are not all expressed in our model's tokens; they should be read as close
approximations that document the \emph{relative composition} of the mixture
rather than an exact token ledger. By contrast, all per-iteration
training-token figures (Section~\ref{sec:hyperparameters},
Table~\ref{tab:train-stages}) and the entire Phase~3 long-context pool
(Section~\ref{sec:longcontext}, Tables~\ref{tab:lc-budget}--\ref{tab:lc-docs})
were obtained by tokenizing the data with the Nemotron-3 tokenizer and are therefore
exact. This distinction also explains why the constructed-pool totals reported
below (e.g.\ ${\sim}23.05$T effective for Phase~1) differ from the
exactly-counted tokens actually consumed during training (e.g.\ ${\sim}20$T;
Table~\ref{tab:phase-budget}): the former are dataset-card estimates of the
\emph{available} pool, the latter are exact counts of what the optimizer
\emph{saw}, and a phase may consume less than one epoch of its pool (Phase~1) or
slightly more than one (Phase~2).

\subsection{Phase 1: Diverse Pretraining}
\label{sec:phase1}

Phase~1 provides the bulk of the training signal. Its composition is given in
full in \Cref{tab:phase1-sources}, grouped by source family with per-family
subtotals. The phase totals $\sim$16.35T raw tokens, which after epoching yield $\sim$23.05T effective tokens in dataset-card terms; the run consumes $\sim$20T of this pool (Section~\ref{sec:hyperparameters}). The mixture is anchored on quality-filtered and
synthetic web text from the three Nemotron-CC releases (CC-v2.1, CC-v2.0, and
CC-v1.0 contribute $\sim$2.84T, $\sim$5.60T, and $\sim$3.15T effective tokens
respectively), supplemented by a large code component ($\sim$3.38T across the
Nemotron code datasets~\cite{nvidia2025nemotron3}), specialized STEM and scientific data
(Nemotron-Pretraining-Specialized-v1~\cite{nvidia2025nemotron3}, $\sim$1.35T), pretraining-stage
SFT mixtures ($\sim$1.57T of Math, Code, and General SFT), and dedicated
mathematics corpora ($\sim$1.06T from Nemotron-CC-Math v1/v2~\cite{karimimahabadi2025nemotronccmath} and a further
$352$B from UltraData-Math~\cite{ultradatamath2025}). PDF-derived text from
FinePDFs~\cite{finepdfs2025} and
Dolma3\_POOL~\cite{olmo3_2025} adds high-value
long-form document data. \Cref{fig:token-allocation} (Left) summarises the Phase~1
mixture by source family.

For the German--English objective, German is intentionally over-represented
relative to the base Nemotron recipe: German sources contribute $\sim$1.65T
effective tokens, or $7.2\%$ of Phase~1, against the $5\%$ multilingual share of
the reference Nemotron~3~Nano mixture~\cite{nvidia2025nemotron3}. The German component combines naturally
occurring web and document text (HPLT Monolingual Datasets 3.0~\cite{oepen2025hplt3},
German Commons~\cite{gienapp2025germancommons}, Genios~\cite{genios2025}, the German subset of FinePDFs~\cite{finepdfs2025} and
FineWiki~\cite{penedo2025finewiki}) with machine-translated (MT) and synthetic German
(MultiSynt/MT~\cite{idahl2026multisyntmttrilliontokenmultiparallelpretraining}, MT-Reasoning~\cite{glaiveai_2025, mt_reasoning_2025}, MT of Nemotron-Multilingual-Reasoning~\cite{gurgurov2025nemotronmultilingual}, PleIAs/Synth~\cite{pleias2025synth}).
Table~\ref{tab:phase1-categories} compares our full Phase 1 mixture against the published Nemotron 3 Nano mixture; relative to it we raise German and academic share, lean more heavily on high-quality web, and trim synthetic web and code-SFT.

\subsection{Phase 2: High-Quality Annealing}
\label{sec:phase2}
The annealing phase coincides with the learning-rate decay and is composed
exclusively of high-value data. The per-source breakdown is given in
Table~\ref{tab:phase2-sources}, organised by category with per-category
subtotals; the constructed pool totals ${\sim}6.30$T effective tokens
(dataset-card counts), of which the annealing schedule trains for ${\sim}6.58$T.
Relative to Phase~1, we drop the lower-tier web pools entirely, retain only the
High-Quality and High-Quality-Synthetic web tiers, and substantially increase
the density of skill-oriented data: code ($\sim$1.03T), mathematics
($\sim$330B), and a broad SFT mixture ($\sim$0.93T) spanning math, code,
agentic, competitive-programming, instruction-following, science, finance,
software-engineering, safety, and multilingual subsets. A dedicated reasoning
bucket---Nemotron-Pretraining-Specialized-v1 and related sources
($\sim$353B)---strengthens chain-of-thought ability ahead of
post-training~\cite{nvidia2025nemotron3}. English Web accounts for $36.4\%$
of the annealing mixture; including Academic \& Wiki, English web/document
data accounts for $42.8\%$.
Figure~\ref{fig:token-allocation} (Mid) summarizes the Phase~2 mixture by
category.

German is up-weighted again during the annealing phase. The German category alone
contributes $965.37$B effective tokens, drawn from a pre-release version of
HPLT-4\footnote{\url{https://hplt-project.org/datasets/v4.0}}, a German
translation of ClimbMix~\cite{diao2025climb} produced with the KletterMix
pipeline~\cite{kraus2026klettermixclimbinghighqualitygerman}, German
FinePDFs-Edu~\cite{finepdfs2025} and FineWiki~\cite{penedo2025finewiki}, and
synthetic/translated German reasoning sources; this brings the multilingual
share to $15.32\%$, more than triple the $5\%$ of the reference mixture.
\Cref{tab:phase2-categories} reports the category-level composition of the
annealing phase against Nemotron~3~Nano.

As part of the SFT mixture, we additionally include QA-base (${\sim}0.05\%$ of
the pool; the $1.43$B English tokens are counted under the SFT category and
the $1.87$B German tokens under the German category in
Table~\ref{tab:phase2-sources}), paraphrased training splits of 25 standard
NLP benchmarks in English and German, analogous to the paraphrase-augmented
benchmark training data in Olmo~3's mid-training mix (e.g.\ TinyMATH, Dolmino
Flan)~\cite{olmo3_2025} and the benchmark-seeded synthetic data in
Nemotron~3~\cite{nvidia2025nemotron3}. A contamination incident affecting four QA-base constituents is disclosed in \Cref{subsec:contamination}.

\subsection{Phase 3: Long-Context Extension}
\label{sec:longcontext}

To extend the usable context window up to 1M tokens, we assemble a long-context
data pool of approximately \textbf{188.5B tokens} drawn from $\sim$21 million 
documents, partitioned into nine sequence-length buckets (4K, 8K, 16K, 32K, 64K,
128K, 256K, 512K, and 1M tokens). The schema targets balanced exposure across
context lengths by allocating comparable token mass to each bucket; in practice the
realized mass per bucket varies with source availability, and several buckets at
the extremes are sparsely populated or empty (Table~\ref{tab:lc-docs}). Seven
domains contribute to the pool: general web (28.78B effective tokens), code
(9.51B), mathematics (6.73B), German (6.68B), and three supervised-fine-tuning
streams, general SFT (31.25B), code SFT (29.24B), and mathematics SFT
(76.31B), which together account for roughly $73\%$ of the pool.
Table~\ref{tab:lc-budget} lists the per-domain token budgets, document counts, and
source priorities; Table~\ref{tab:lc-docs} gives the per-bucket document counts. Unlike Phases~1--2, every Phase~3 token count was produced by tokenizing the
data with the Nemotron-3 tokenizer, so the figures in this section and in
Tables~\ref{tab:lc-budget}--\ref{tab:lc-docs} are exact rather than dataset-card
estimates.
 
The long-context stage trained for 2{,}000 optimizer steps at a 1M-token sequence length, i.e. approximately 100.66B tokens (Section~\ref{sec:phase3-hparams})---about $53\%$ of
the pool, or roughly half an epoch. We stopped at this point because no further
loss improvement was observed from additional long-context training, mirroring
the rationale for the discarded final-annealing stage (Section~\ref{sec:hyperparameters}).
 
When a bucket draws from several sources, we fill it according to a fixed priority
order: for web, ClimbMix~\cite{diao2025climb} is preferred over
OlmoOCR~\cite{poznanski2025olmocr}, which is preferred over
FinePDFs~\cite{finepdfs2025}; for code,
SwallowCode~\cite{fujii2025swallowcode} is preferred over
Nemotron-Pretraining-Code-v1/v2~\cite{nvidia2025nemotron3}; and German is composed of 40\% HPLTv4 and 60\%
German translation of
ClimbMix~\cite{diao2025climb} with the pipeline of KletterMix~\cite{kraus2026klettermixclimbinghighqualitygerman}. 
 
We note the population irregularities in the interest of full disclosure.
Our mathematics sources do not provide documents beyond the 64K bucket, so the longer mathematics buckets are effectively unpopulated (2 documents at 128K, none beyond). The web and German domains have no populated 256K bucket, and the code domain's 4K and 8K buckets are effectively
empty (whole-file repository code is scarce at these lengths after packing); these
short sequence lengths are instead carried by the web, mathematics, and SFT
streams, the latter being densest at 8K--64K. The SFT mathematics buckets nominally
extend to 256K but are negligible above 64K (19 and 2 documents at 128K and 256K,
respectively).

\subsection{Data Provenance and Release}
\label{sec:data-provenance}

The overwhelming majority of our pretraining data is openly available. The web,
code, mathematics, specialized, and SFT components build on NVIDIA's publicly
released Nemotron pretraining datasets~\cite{su2024nemotroncc,karimimahabadi2025nemotronccmath,nvidia2025nemotron3} (Nemotron-CC v1.0/v2.0/v2.1,
Nemotron-CC-Math, Nemotron-CC-Code, Nemotron-Pretraining-Code/Specialized/SFT,
and the Nemotron post-training SFT collections), complemented by open corpora
including the Dolma~3 pools~\cite{olmo3_2025}, the Fine* family~\cite{penedo2024fineweb,finepdfs2025} (FinePDFs,
FinePDFs-edu, FineWiki), ClimbMix~\cite{diao2025climb},
SwallowCode~\cite{fujii2025swallowcode}, UltraData-Math~\cite{ultradatamath2025}, and AceReason~\cite{liu2025acereasonnemotron11advancingmath}. German coverage is provided
by open resources such as HPLT (v3 and v4)~\cite{burchell2025hplt2,oepen2025hplt3}, German-Commons~\cite{gienapp2025germancommons}, KletterMix~\cite{kraus2026klettermixclimbinghighqualitygerman}, 
and the Fine* German subsets, augmented with machine-translated and synthetic
German data (the MultiSynt/MT-* sources, Soofi-Think-SFT~\cite{reasonxl2026}), plus the commercially licensed Genios~\cite{genios2025} corpus.
Consistent with the openness goals of this work, we release the complete mixture specification; every source,
its raw token count, epoch multiplier, and effective contribution for all three
phases, so that the corpus can be independently reconstructed where source licenses
permit; for commercially licensed Genios, we release aggregate statistics and
exact mixture accounting rather than redistributing raw text.

\paragraph{Openness classification.}
The definition of open-source AI remains contested: the OSI's Open Source
AI Definition 1.0~\cite{osi2024osaid} permits documented but unsharable
training data, whereas stricter proposals for a European definition
require every training token to be redistributable~\cite{osai_definitions_2026}.
\ourmodel{} satisfies OSAID~1.0: we will release weights, intermediate checkpoints,
training and evaluation code, and exact per-source data accounting under permissive licenses. Under the stricter open-data
standard, \ourmodel{} falls short in exactly one documented component:
the commercially licensed Genios~\cite{genios2025} corpus ($1.3\%$ of Phase~1 effective tokens, reported in aggregate in Appendix~\ref{app:genios}). Every other
source is publicly obtainable, so ${\sim}99\%$ of the mixture can be
independently reconstructed. We state this boundary explicitly rather
than claim a stronger openness status than the release supports.

\subsubsection{Code Web Data}
\label{sec:code-web-data}

Our code-from-web component is Nemotron-CC-Code-v1~\cite{nvidia2025nemotron3}, a corpus of
code and code-adjacent documents recovered directly from Common Crawl rather
than from repository hosting platforms. Standard web-extraction pipelines tend
to mangle source code embedded in HTML, collapsing the whitespace and
indentation that is syntactically meaningful in languages such as Python, so
this corpus is built with a code-aware extraction pipeline that preserves the
formatting of code blocks, retains the surrounding natural-language context
(tutorials, documentation, Q\&A threads, blog posts), and applies dedicated
cleaning, language identification, and deduplication stages. The result is code
in its \emph{instructional habitat}: implementations interleaved with the prose
that explains them, which is a complementary signal to raw repository files.
We use the \textsc{Actual} subset at 3 epochs ($427.9$B raw, $1{,}283.7$B
effective tokens; Table~\ref{tab:phase1-sources}), making it the single largest
code source in Phase~1.

\subsubsection{Curated Code Data}
\label{sec:curated-code-data}

Repository-sourced code is drawn from
Nemotron-Pretraining-Code-v1~\cite{nvidia2025nvidianemotronnano2} and -v2~\cite{nvidia2025nemotron3}, which curate
permissively licensed source code from public repositories with
quality filtering, per-language balancing, and aggressive deduplication. Both
releases pair the curated \textsc{Actual} code with synthetic derivatives
generated from it, and we train on five such variants from v2:
\emph{code review} (critiques and improvement suggestions for real code),
\emph{question answering} (Q\&A pairs grounded in repository code),
\emph{rewriting} (refactorings and alternative implementations of the same
functionality), \emph{student--teacher} (dialogues that explain code
step-by-step), and \emph{transpilation} (translations of programs between
programming languages). These variants convert passive code exposure into
bidirectional code--language supervision during pretraining itself. In Phase~1,
the curated-code family contributes $2{,}091.14$B effective tokens (the
\textsc{Actual} subsets at 3 epochs, synthetic subsets at 2), and the full
family is retained at 1 epoch during annealing
(Table~\ref{tab:phase2-sources}). During annealing, we additionally include
Swallow-Code-v2~\cite{fujii2025swallowcode} (stage~5 subset for 2 epochs) and the Dolma~3
Dolmino code pool~\cite{olmo3_2025} as high-quality curated complements.

\subsubsection{German and English Web Data}
\label{sec:web-data}

General web text is the backbone of the corpus. The English side builds on the
three Nemotron-CC releases~\cite{su2024nemotroncc,nvidia2025nemotron3} (v1.0, v2.0, v2.1), which classify Common Crawl
documents into quality tiers using ensembles of model-based quality
classifiers, and augment the high-value tiers with synthetic transformations:
rephrased variants of high-quality pages (\textsc{-Synthetic}), diverse
question--answer reformulations (\textsc{Diverse QA}), and English translations
of high-quality non-English crawl data (\textsc{Translated-To-English}). As
described in Section~\ref{sec:quality-tiers}, we up-sample the High-Quality and
synthetic tiers and exclude the Medium tier entirely; the three releases
together contribute $11{,}601.4$B effective tokens to Phase~1. Long-form
English document data comes from PDF-derived corpora
(FinePDFs~\cite{finepdfs2025} and the Dolma~3 PDF pool \cite{olmo3_2025}), which supply the
book-, report-, and paper-style text that is underrepresented in HTML crawls.

The German web component is assembled from sources of complementary character.
Naturally occurring German is provided by quality-filtered crawl data---the
HPLT corpora~\cite{burchell2025hplt2,oepen2025hplt3} restricted to the top decile of their educational-quality score HPLT-3-Top10\% in Phase~1 (edu-scores based on JQL~\cite{ali2025judgingqualitylanguagesmultilingual}, HPLT-4-Top10\% in
Phase~2 (edu-scores based on Propella~\cite{idahl2026propella1multipropertydocumentannotation})---together with German-Commons~\cite{gienapp2025germancommons} (openly licensed German text),
Genios~\cite{genios2025}, a commercially licensed corpus of 916 German newspaper and trade-press archives comprising 193M articles (57.6B words, 2010--2025; see Appendix~\ref{app:genios}), the German
FinePDFs/FinePDFs-Edu subsets, German FineWiki, and
the curated mixture used during annealing. 

Since the supply of high-quality native German text is far smaller than for English, we
extend it with translated and synthetic German: KletterMix~\cite{kraus2026klettermixclimbinghighqualitygerman} (machine translations of ClimbMix~\cite{diao2025climb}), MultiSynt/MT~\cite{idahl2026multisyntmttrilliontokenmultiparallelpretraining} (machine translations of high-quality Nemotron-CC~\cite{su2024nemotroncc} English documents into German), MT-Reasoning~\cite{mt_reasoning_2025}, and Nemotron-Multilingual-Reasoning~\cite{gurgurov2025nemotronmultilingual} (translated reasoning traces), and the German subset of PleIAs/Synth~\cite{pleias2025synth}. 

Phase~1 German up-sampling leans on
the scarce high-quality crawl (the HPLT~v3 top-decile pool runs $8.4$ epochs),
while annealing switches to the fresher HPLT~v4 pool and a German translation
of ClimbMix produced with the KletterMix pipeline~\cite{kraus2026klettermixclimbinghighqualitygerman}.
In total, German receives $7.2\%$ of Phase~1 and $15.3\%$ of Phase~2
effective tokens, well above the ${\sim}5\%$ \emph{total} multilingual share
of the reference recipe, concentrated in a single language.

\subsubsection{Specialized Synthetic Data}
\label{sec:specialized-synthetic}

Beyond web and code, we train on three families of specialized, largely
synthetic data. First, nvidia/Nemotron-CC-Math~\cite{karimimahabadi2025nemotronccmath} (v1 and v2) recovers
mathematical content from Common Crawl with a rendering-based pipeline that
preserves equations and converts them to a uniform LaTeX representation,
sorted into quality bands; we train on all bands for 4 epochs in Phase~1 and
keep the top v1 bands during annealing, complemented by
UltraData-Math~\cite{ultradatamath2025}.
Second, Nemotron-Pretraining-Specialized-v1~\cite{nvidia2025nemotron3} provides targeted synthetic corpora for STEM and reasoning, including synthetic math, Wikipedia-style
rewrites of encyclopedic content, and competition-style problems, which we
up-sample aggressively (5 epochs, $1{,}353.5$B effective tokens, $5.9\%$ of
Phase~1) and retain at 1 epoch during annealing alongside the Dolma~3
\textsc{Thinking} pool~\cite{olmo3_2025}. Third, we include SFT-formatted data already at the pretraining stage: Nemotron-Pretraining-SFT-v1 (Math, Code, and
General splits) in Phase~1, broadened during annealing by the Nemotron
post-training collections\footnote{\url{https://huggingface.co/collections/nvidia/nemotron-post-training-v3}} (agentic, competitive programming, instruction
following, math proofs, science, finance, software engineering, safety, and
multilingual subsets; ~\Cref{tab:phase2-sources}). Exposing the model to
instruction- and reasoning-formatted text before post-training shortens the
distribution shift at the SFT stage and measurably strengthens
chain-of-thought behaviour of the base model.

\subsection{Data Mixture and Ordering}
\label{sec:data-mixture-ordering}

We organise all sources into seven unified categories, English Web,
Academic \& Wiki, Code, Mathematics, SFT, Reasoning, and German, and steer the mixture at this category level (the per-source realisation is given in
Appendix~\ref{app:data-composition}). The appendix tables retain the finer Nemotron-style accounting categories; Figure~\ref{fig:token-allocation} folds them into the seven unified categories
used in the main text. In Phase~2, the main-text Reasoning category corresponds
to the appendix row ``Reasoning / STEM-SFT'', while the narrower longtable
``Reasoning subtotal'' reports only the non-SFT reasoning sources. The guiding principle of the ordering is a quality- and skill-based curriculum aligned with the WSD learning-rate
schedule: breadth while the learning rate is high, concentration while it
decays.

During the stable phase (Phase~1), the mixture is dominated by diverse web text
($50.3\%$ English Web plus $8.0\%$ academic and wiki text), with code at
$14.6\%$, reasoning at $10.4\%$, German at $7.2\%$, mathematics at $6.0\%$,
and SFT at $3.5\%$ (Figure~\ref{fig:token-allocation}).
Up-sampling via epoch multipliers, rather than the inclusion of lower-quality
pools, is the primary lever for hitting these targets: high-quality and
synthetic tiers run $2$--$6$ epochs, the medium tiers run zero. At the onset of learning-rate decay (Phase~2), the mixture shifts decisively
toward skill density and German depth: English Web drops to $36.4\%$,
Academic \& Wiki accounts for $6.4\%$, SFT contributes $8.5\%$, Reasoning
accounts for $11.8\%$, Code accounts for $16.4\%$, Mathematics accounts for
$5.2\%$, and German rises to $15.3\%$
(Figure~\ref{fig:token-allocation}).

This places the highest-value tokens in the regime where the decaying learning rate
consolidates them most effectively. Sources seen for multiple epochs in
Phase~1 are re-weighted downward in Phase~2 (typically to a single epoch, or
fresh replacements are substituted, e.g.\ HPLT~v3\,$\to$\,v4) so that the
annealing phase adds new signal rather than repeating saturated data. 

Phase~3 builds on top of the highest-quality documents from Phase~2 and spans seven
document-level domains (web, code, math, German, Math SFT, General-SFT, and Code-SFT). It
is organized by \emph{length}, using the length-bucketed
scheme of Section~\ref{sec:longcontext}.

Relative to the reference Nemotron~3~Nano mixture, our category targets differ
in one deliberate respect: German replaces the broad multilingual bucket and is
raised from $5\%$ (Phase~1 \& Phase~2) to $7.2\%$ (our Phase~1) and $15.32\%$ (our Phase~2), funded by
reductions in synthetic web and code-SFT share
(Tables~\ref{tab:phase1-categories} and~\ref{tab:phase2-categories}). Within
each phase, category proportions are held stationary: data of all categories is
shuffled and interleaved uniformly at the batch level, so the curriculum acts
between phases, not within them.

\section{Evaluations}
\label{sec:base-evals}
We evaluate \ourmodel{} 30B-A3B Base against \numbaselines{} open-source and open-weight base models using a common \texttt{lm-evaluation-harness}~\cite{gao2024lmeval} pipeline with identical prompts, few-shot configurations, and task settings.\footnote{The comparison includes Nemotron 3 Nano 30B-A3B; Qwen3.5 35B-A3B and 9B; Gemma 3 27B; Ministral 3 3B, 8B, and 14B; Olmo 3 32B; Alia 40B; Apertus 8B and 70B; EuroLLM 9B and 22B; Teuken 7B; and Salamandra 7B.} The benchmark suite closely follows the Olmo 3 evaluation setup~\cite{olmo3_2025} and covers code, mathematics, knowledge, reasoning, science, reading comprehension, and German-language proficiency.

The main text focuses on the strongest baselines in two comparison groups. The open-source group comprises Alia~40B~\cite{gonzalezagirre2025salamandra}, EuroLLM~22B~\cite{martins2024eurollm,martins2025eurollm9btechnicalreport}, Apertus~70B~\cite{hernandezcano2025apertus}, and Olmo~3~32B~\cite{olmo3_2025}. The open-weight group comprises the architectural reference Nemotron~3~Nano~30B-A3B~\cite{nvidia2025nemotron3} and models with comparable or larger active parameter counts: Qwen3.5~35B-A3B~\cite{qwen3_2025,qwen35_2026}, Ministral~3~14B~\cite{mistral2025mistral3}, and Gemma~3~27B~\cite{gemma3_2025}.

\Cref{fig:serving-efficiency} summarizes aggregate evaluation performance and 40K serving throughput. Tables~\ref{tab:base-evals-opensource} and~\ref{tab:base-evals-openweight} report detailed results for the two main comparison groups, while complete per-task results are released alongside this report.

\ourmodel{} is evaluated at the selected base checkpoint
\texttt{iter\_1056000}, the final checkpoint of the constant-annealing stage
(Section~\ref{sec:hyperparameters}, Appendix~\ref{app:checkpoint-merging});
for other checkpointed models we report the highest available training step. The
complete per-task results files are released alongside the model; headline
numbers are collected in Tables~\ref{tab:base-evals-opensource} and~\ref{tab:base-evals-openweight} and
Figures~\ref{fig:eval-overview-opensource} and~\ref{fig:eval-overview-openweight}.

\subsection{\ourmodel{} vs Open-Source Models}
\label{subsec:eval-opensource}

\begin{figure}[ht]
\centering
\includegraphics[width=\textwidth]{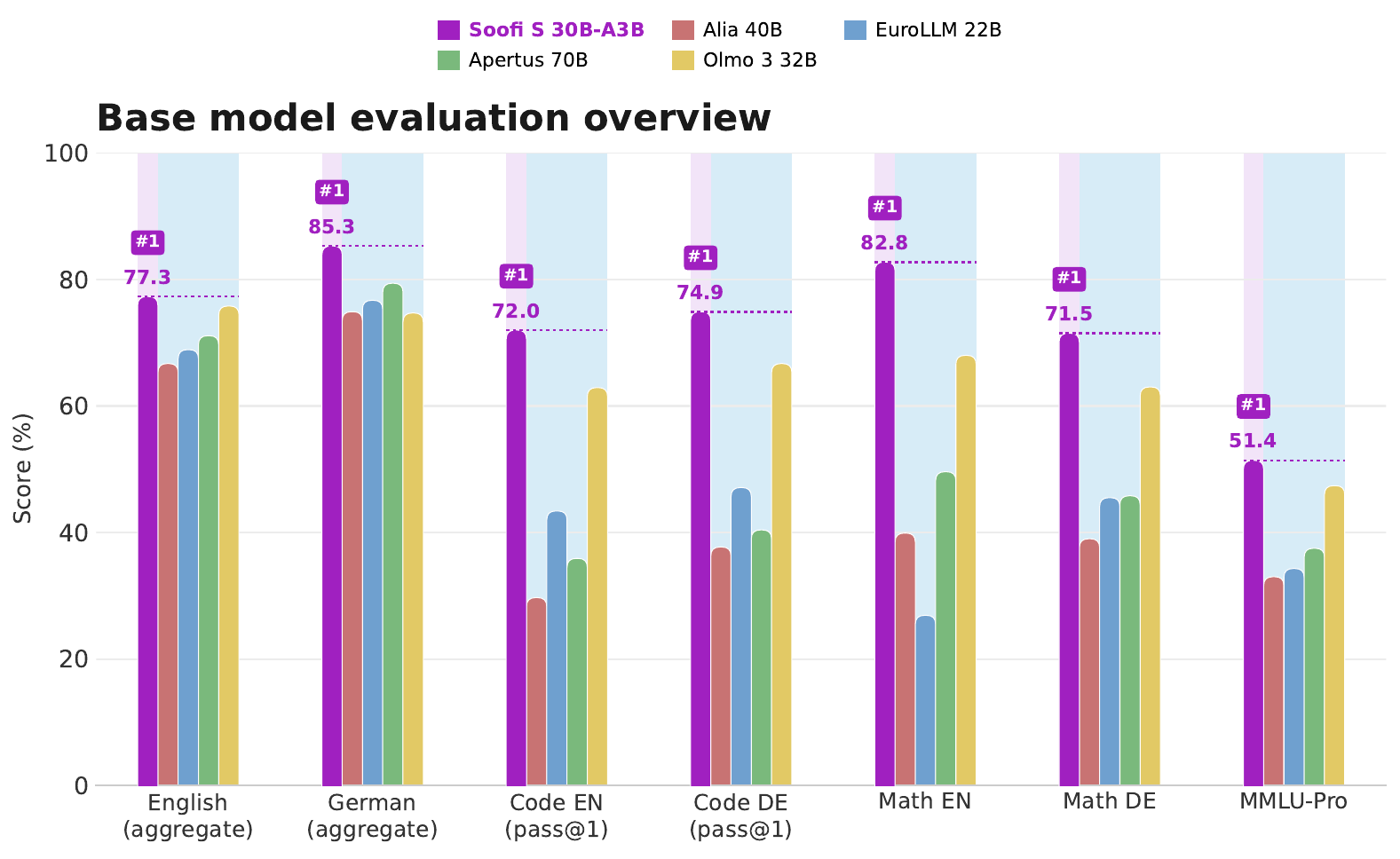}
\caption{Evaluation overview for the open-source comparison. \colorbox{soofipurple}{\,\ourmodel{}\,} is compared against large \colorbox{osrcblue}{\,open-source models\,} (Alia, EuroLLM, Apertus, and Olmo~3). Aggregates are the harness-level English and German suite means. Code EN averages HumanEval and MBPP, Code DE averages HumanEval-DE and MBPP-DE, and LBPP is reported separately.}

\label{fig:eval-overview-opensource}
\end{figure}

The open-source comparison reports results against Alia~40B, EuroLLM~22B,
Apertus~70B, and Olmo~3~32B. We separate these models from the larger
open-weight baselines in Section~\ref{subsec:eval-openweight} so that each
table compares models with similar release categories. Within this set, \ourmodel{} is the strongest model overall: it obtains the highest English aggregate (+1.5 over Olmo~3~32B) and German aggregate (+5.9 over Apertus~70B) (Figure~\ref{fig:eval-overview-opensource}).

Table~\ref{tab:base-evals-opensource} gives the per-task results. The largest
margins over the next-best open-source baseline occur on German and technical
benchmarks: German aggregate ($+5.9$), HumanEval ($+10.8$), MBPP-DE ($+13.4$),
GSM8K-Platinum-DE ($+9.7$), INCLUDE-DE ($+10.1$), and GLP-DE ($+7.6$).
Olmo~3~32B is the strongest non-Soofi model overall and leads this subset on
LBPP, SocialIQA, SQuAD, DROP, and NaturalQuestions. These results identify the
main residual gaps for \ourmodel{} in this comparison as contamination-aware
code evaluation, open-domain factual recall, and extractive reading
comprehension.

\begin{table}
    \centering
    \setlength{\textwidth}{\dimexpr\paperheight-2in\relax}%
    \setlength{\linewidth}{\textwidth}%
    \setlength{\LTcapwidth}{\textwidth}%
    \setlength{\tabcolsep}{2.5pt}
    \setlength{\LTleft}{0pt}\setlength{\LTright}{\fill}
    \caption{Base model evaluation results (\%) against large \emph{open-source}
    models. Best result per row in \textbf{bold}, second best
    \underline{underlined}. All models evaluated with identical harness,
    prompts, and few-shot settings; ``-DE'' denotes the German variant of a benchmark. Column shading:
    \colorbox{soofipurple}{\,\ourmodel{}\,} (ours),
    \colorbox{osrcblue}{\,open-source\,} (Alia, EuroLLM, Apertus, Olmo 3). Aggregates are the harness-level English and German suite
    means (recomputed with GPQA and held-out group excluded for all models; see
    Section~\ref{subsec:contamination}).}
    \label{tab:base-evals-opensource}
    \resizebox{0.45\linewidth}{!}{
    \begin{tabular}{l
      >{\columncolor{soofipurple}}c
      >{\columncolor{osrcblue}}c
      >{\columncolor{osrcblue}}c
      >{\columncolor{osrcblue}}c
      >{\columncolor{osrcblue}}c}
    \toprule
    \textbf{Benchmark} & \textbf{\ourmodel{}} & \textbf{Alia} & \textbf{EuroLLM} & \textbf{Apertus} & \textbf{Olmo 3} \\
     & \textbf{30B-A3B} & \textbf{40B} & \textbf{22B} & \textbf{70B} & \textbf{32B} \\
    \midrule
    \multicolumn{6}{l}{\textbf{Aggregates}}\\
    English aggregate & \textbf{77.3} & 66.7 & 68.9 & 71.1 & \underline{75.8} \\
    German aggregate & \textbf{85.3} & 74.9 & 76.7 & \underline{79.4} & 74.7 \\
    \midrule
    \multicolumn{6}{l}{\textbf{Code} (pass@1)}\\
    HumanEval & \textbf{73.8} & 23.8 & 39.3 & 30.2 & \underline{63.0} \\
    MBPP & \textbf{70.2} & 35.5 & 47.6 & 41.5 & \underline{62.8} \\
    LBPP & \underline{31.0} & 8.6 & 10.7 & 6.4 & \textbf{32.1} \\
    HumanEval-DE & \textbf{65.5} & 29.9 & 34.7 & 29.9 & \underline{62.5} \\
    MBPP-DE & \textbf{84.2} & 45.6 & 59.4 & 50.9 & \underline{70.8} \\
    \midrule
    \multicolumn{6}{l}{\textbf{Mathematics}}\\
    GSM8K & \textbf{86.1} & 65.4 & 25.1 & 65.4 & \underline{80.7} \\
    GSM8K-Platinum-DE & \textbf{87.1} & 65.0 & 62.6 & 62.6 & \underline{77.4} \\
    Minerva 500 & \textbf{79.4} & - & 28.7 & 33.7 & \underline{55.2} \\
    Minerva Math-EN & \textbf{81.0} & - & 26.4 & 34.5 & \underline{54.0} \\
    Minerva MATH-DE & \textbf{56.0} & - & 28.4 & 29.0 & \underline{48.5} \\
    \midrule
    \multicolumn{6}{l}{\textbf{Knowledge}}\\
    MMLU stem & \textbf{75.9} & 54.0 & 58.2 & 58.4 & \underline{70.9} \\
    MMLU-Pro & \textbf{51.4} & 33.0 & 34.3 & 37.5 & \underline{47.4} \\
    MMLU-Pro-DE & \textbf{49.4} & 29.4 & 33.4 & 34.0 & \underline{38.0} \\
    INCLUDE-DE & \textbf{61.2} & 43.9 & \underline{51.1} & 50.4 & 48.2 \\
    NaturalQuestions (acc) & \underline{79.0} & 69.0 & 70.1 & 73.0 & \textbf{79.1} \\
    \midrule
    \multicolumn{6}{l}{\textbf{Commonsense and reading comprehension}}\\
    HellaSwag & \underline{64.1} & 62.9 & 60.4 & \textbf{64.5} & 63.3 \\
    HellaSwag-DE & \textbf{74.0} & 65.8 & 57.9 & \underline{68.6} & 38.5 \\
    PIQA & \textbf{85.7} & 83.7 & 82.4 & \underline{85.3} & 83.5 \\
    PIQA-DE & \textbf{91.5} & 85.7 & 80.5 & \underline{86.4} & 67.4 \\
    SocialIQA & \underline{60.5} & 55.2 & 57.4 & 59.3 & \textbf{62.4} \\
    SocialIQA-DE & \textbf{87.6} & \underline{86.6} & 85.0 & \textbf{87.6} & 75.9 \\
    SQuAD (EM) & \underline{87.5} & 78.7 & 81.2 & 83.2 & \textbf{88.1} \\
    DROP (EM) & \underline{66.5} & 41.5 & 43.0 & 50.0 & \textbf{72.2} \\
    \midrule
    \multicolumn{6}{l}{\textbf{Reasoning and science}}\\
    BBH (CoT) & \textbf{78.8} & 53.9 & 57.8 & 60.5 & \underline{77.4} \\
    AGIEval & \textbf{66.9} & 48.1 & 52.2 & 55.7 & \underline{63.2} \\
    ARC-Challenge & \textbf{90.6} & 79.8 & 81.1 & 85.8 & \underline{89.9} \\
    \midrule
    \multicolumn{6}{l}{\textbf{German language proficiency}}\\
    GLP-DE & \textbf{88.8} & 65.4 & 78.2 & \underline{81.2} & 73.0 \\
    ARC-Challenge-DE & \textbf{92.3} & 81.5 & 83.6 & \underline{85.3} & 83.6 \\
    \bottomrule
    \end{tabular}}
\end{table}

\begin{figure}[ht]
\centering
\includegraphics[width=\textwidth]{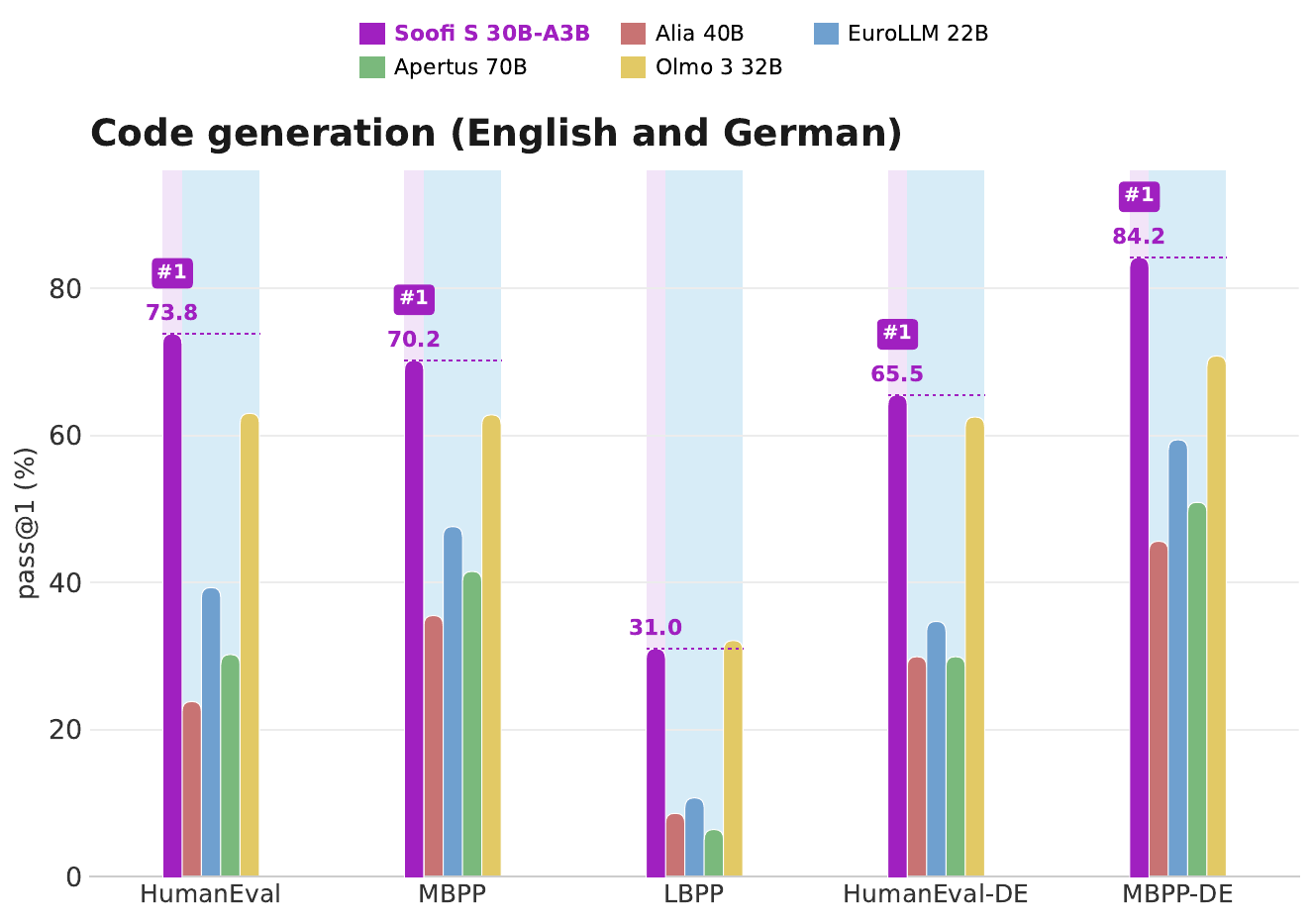}
\caption{Code generation results (pass@1) against large open-source models on English and German benchmarks. \ourmodel{} leads this comparison on HumanEval, MBPP, HumanEval-DE, and MBPP-DE; Olmo~3~32B is strongest on LBPP.}
\label{fig:eval-code-opensource}
\end{figure}

\paragraph{Code performance.}
\ourmodel{} ranks first on four of the five code benchmarks in the open-source
comparison (Figure~\ref{fig:eval-code-opensource}). It exceeds the next-best
open-source baseline by $10.8$ points on HumanEval, $7.4$ points on MBPP,
$3.0$ points on HumanEval-DE, and $13.4$ points on MBPP-DE. LBPP is the only
code benchmark in this subset where \ourmodel{} is not first; Olmo~3~32B scores
$32.1$ compared with $31.0$ for \ourmodel{}.

\begin{figure}[ht]
\centering
\includegraphics[width=\textwidth]{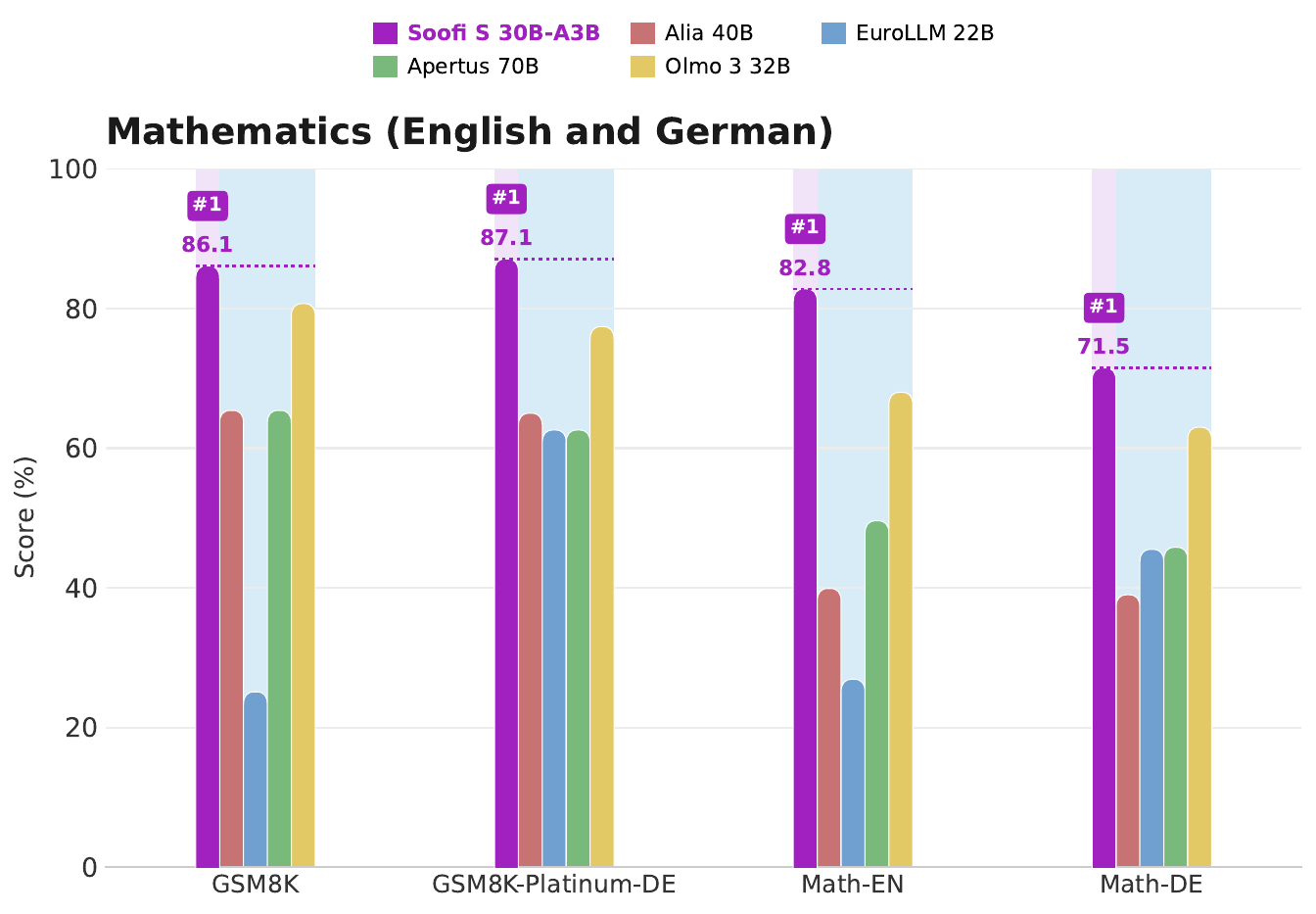}
\caption{Mathematics results against large open-source models on English and German benchmarks. \ourmodel{} leads the open-source comparison on GSM8K, GSM8K-Platinum-DE, Math-EN, and Math-DE.}
\label{fig:eval-math-opensource}
\end{figure}

\begin{figure}[ht]
\centering
\includegraphics[width=\textwidth]{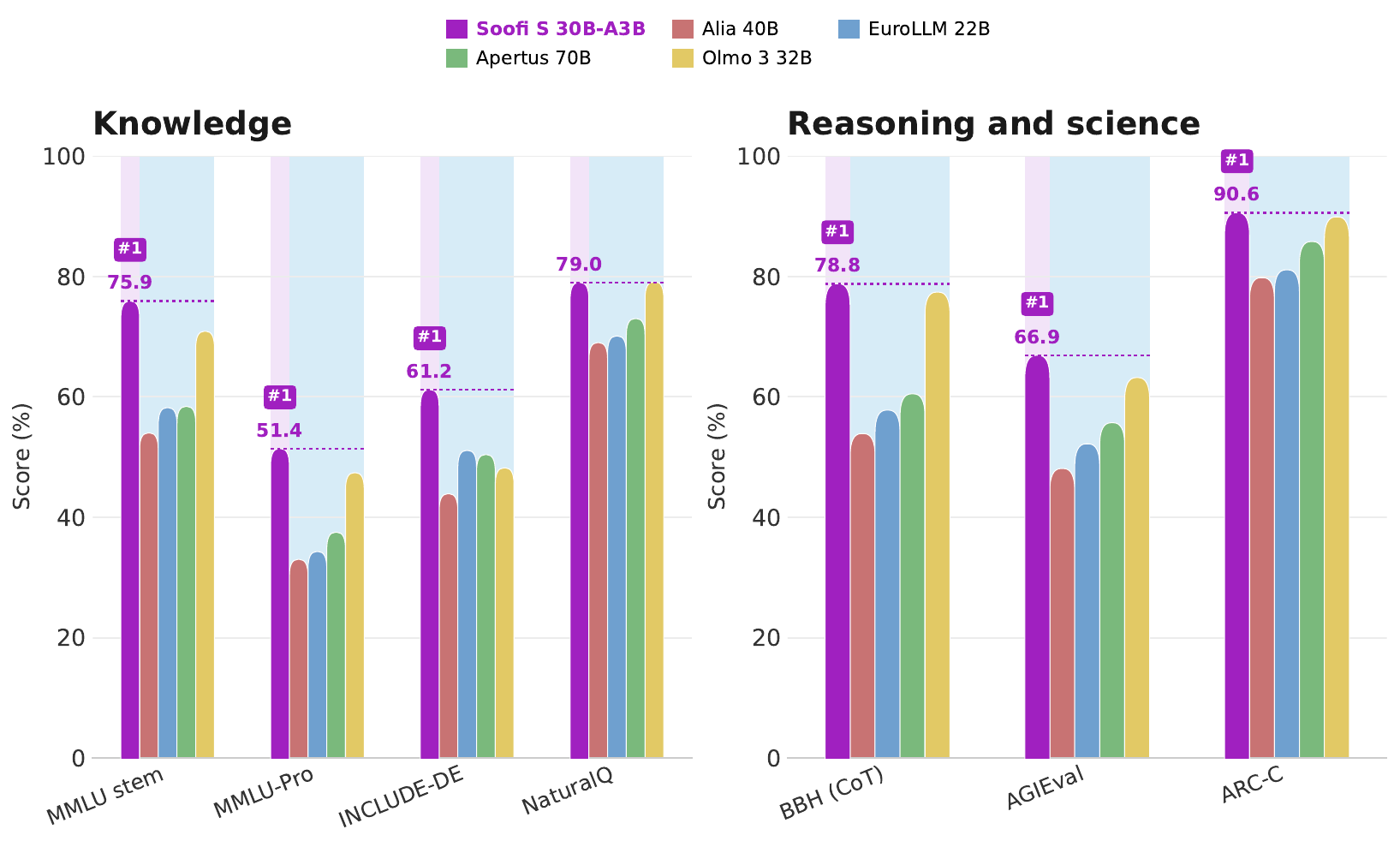}
\caption{Knowledge (left) and reasoning/science (right) benchmarks against large open-source models. \ourmodel{} leads the open-source comparison on MMLU-STEM, MMLU-Pro, INCLUDE-DE, BBH, AGIEval and ARC-Challenge.}
\label{fig:eval-knowledge-reasoning-opensource}
\end{figure}

\paragraph{Mathematics, knowledge, and reasoning.}
\ourmodel{} also ranks first within the open-source subset on all mathematics
benchmarks shown in Table~\ref{tab:base-evals-opensource} and
Figure~\ref{fig:eval-math-opensource}. The largest mathematics margins are on
GSM8K-Platinum-DE ($+9.7$ over Olmo~3~32B), Minerva-500 ($+24.2$ over
Olmo~3~32B), Minerva Math-EN ($+27.0$ over Olmo~3~32B), and Minerva MATH-DE
($+7.5$ over Olmo~3~32B). In knowledge, reasoning, and science
(Figure~\ref{fig:eval-knowledge-reasoning-opensource}), \ourmodel{} ranks
first on MMLU-STEM, MMLU-Pro, MMLU-Pro-DE, INCLUDE-DE, BBH, AGIEval, and ARC-Challenge.

\begin{figure}[ht]
\centering
\includegraphics[width=\textwidth]{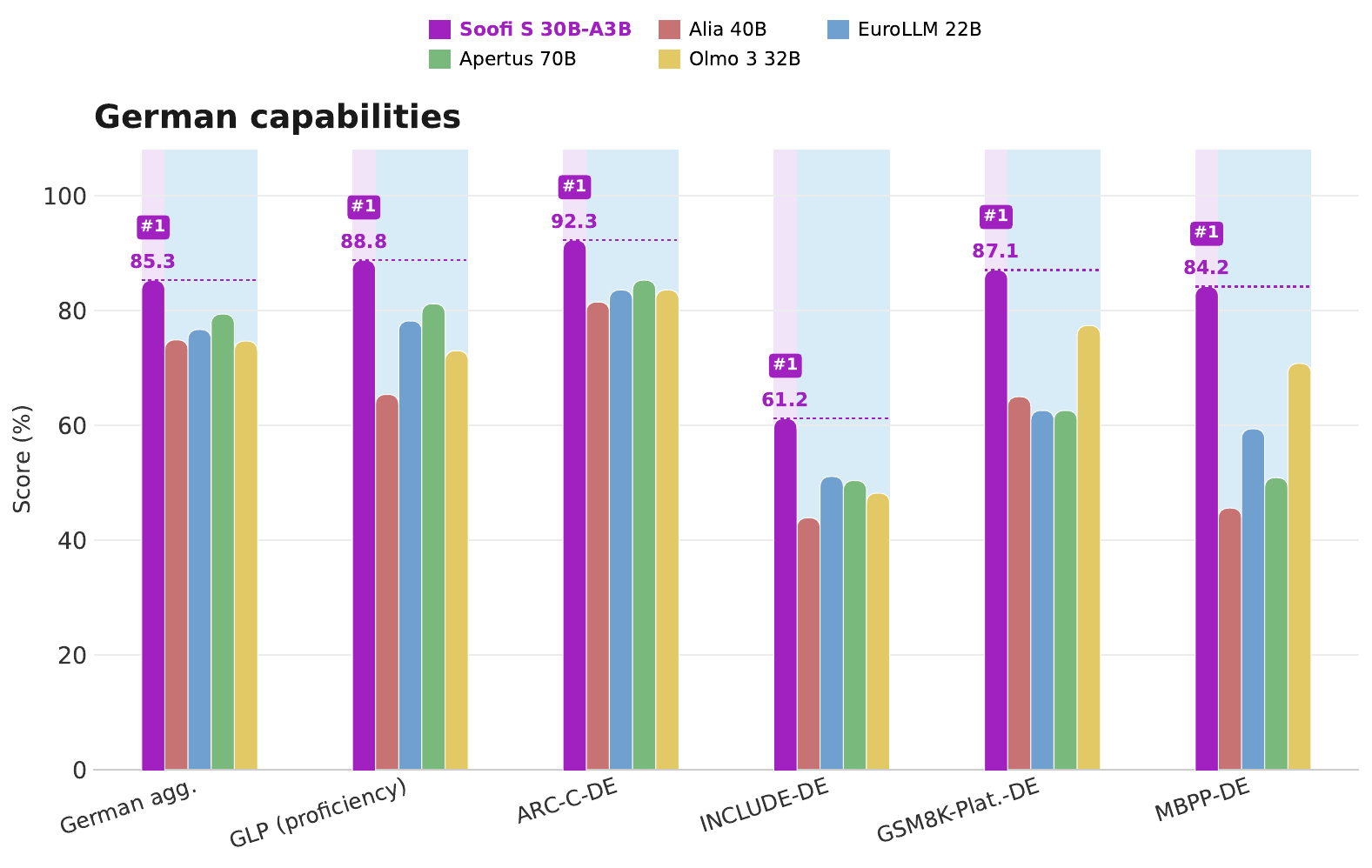}
\caption{German benchmark results against large open-source models. \ourmodel{} ranks first on the German aggregate, GLP-DE, ARC-Challenge-DE, INCLUDE-DE, GSM8K-Platinum-DE, and MBPP-DE in this comparison.}
\label{fig:eval-german-opensource}
\end{figure}

\paragraph{German capabilities.}
Figure~\ref{fig:eval-german-opensource} isolates the German benchmarks for the
open-source comparison. \ourmodel{} ranks first on every German row in this
subset, with margins of $+5.9$ on the German aggregate, $+7.6$ on GLP-DE,
$+7.0$ on ARC-Challenge-DE, $+10.1$ on INCLUDE-DE, $+9.7$ on
GSM8K-Platinum-DE, and $+13.4$ on MBPP-DE
relative to the strongest open-source baseline for each metric.

\FloatBarrier
\subsection{\ourmodel{} vs Open-Weight Models}
\label{subsec:eval-openweight}

The open-weight comparison contains two types of baselines: the
architecture-identical Nemotron~3~Nano reference, and larger public-weight
models from Qwen, Ministral, and Gemma. This split separates the
effect of the German--English data recipe from comparisons against larger
models with different architectures and training corpora. In this subset,
\ourmodel{} is not the top model on aggregate scores, but it is competitive
with larger dense baselines and improves over Nemotron~3~Nano on all aggregate metrics.

\begin{figure}[ht]
\centering
\includegraphics[width=\textwidth]{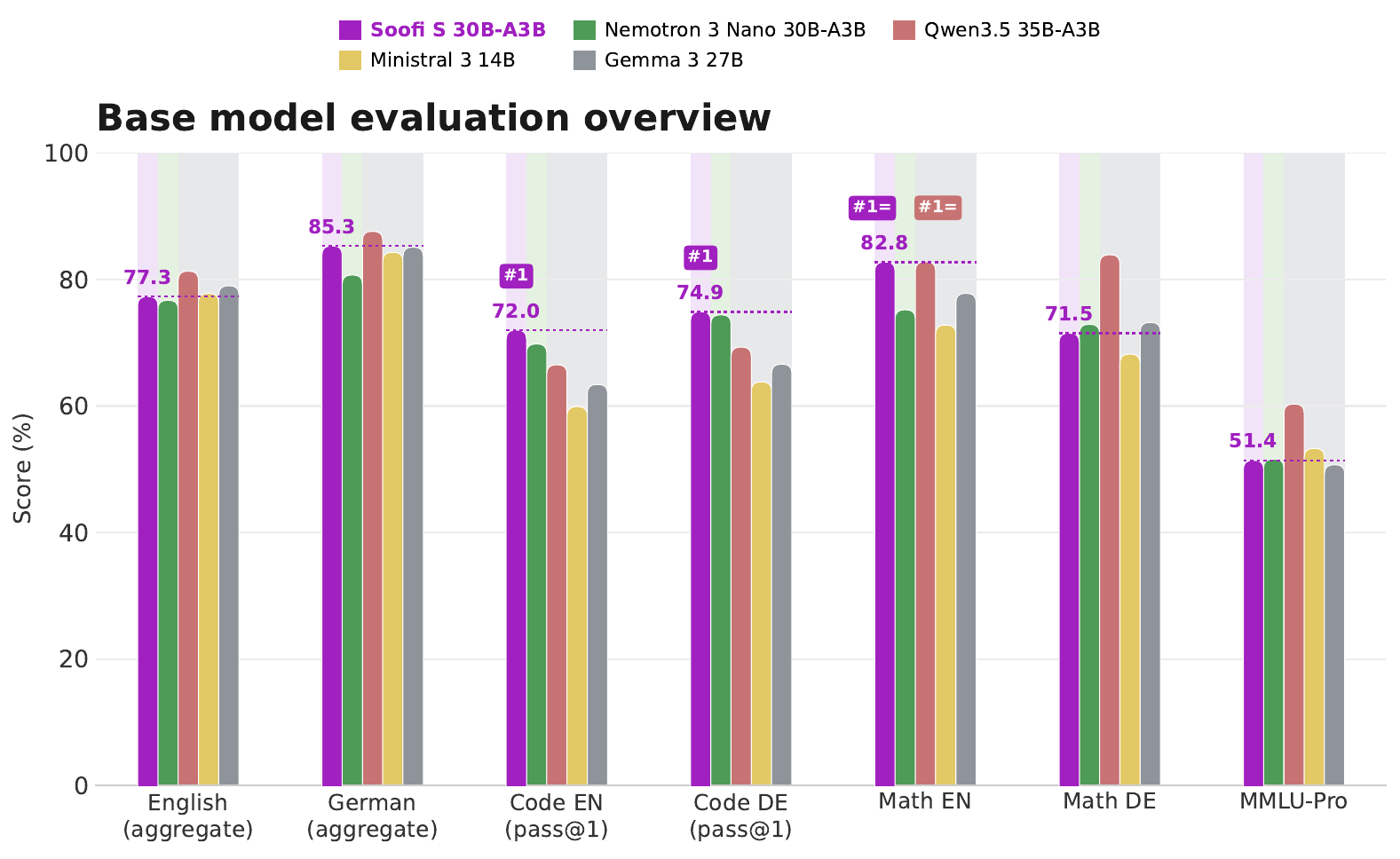}
\caption{Base model evaluation overview for \colorbox{soofipurple}{\,\ourmodel{}\,} against \colorbox{archgreen}{\,Nemotron\,} (same architecture) and large \colorbox{openwt}{\,open-weight models\,} (Qwen, Ministral, and Gemma). Aggregates are the harness-level English and German suite means. Code EN averages HumanEval and MBPP, Code DE averages HumanEval-DE and MBPP-DE, and LBPP is reported separately.}
\label{fig:eval-overview-openweight}
\end{figure}

At the aggregate level, Qwen3.5~35B-A3B has the highest English and German aggregates in this subset (Table~\ref{tab:base-evals-openweight}).
\ourmodel{} scores $77.3$ on the English aggregate, closely behind Gemma~3~27B ($79.0$) and Ministral~3~14B ($77.8$). On the
German aggregate, \ourmodel{} scores $85.3$, compared with $84.3$ for
Ministral~3~14B and $85.1$ for Gemma~3~27B.
Relative to Nemotron~3~Nano, \ourmodel{} improves the English aggregate by
$+0.6$ and the German aggregate by $+4.6$.

\begin{table}
    \centering
    \setlength{\textwidth}{\dimexpr\paperheight-2in\relax}%
    \setlength{\linewidth}{\textwidth}%
    \setlength{\LTcapwidth}{\textwidth}%
    \setlength{\tabcolsep}{2.5pt}
    \setlength{\LTleft}{0pt}\setlength{\LTright}{\fill}
    \caption{Base model evaluation results (\%) against Nemotron 3 Nano and
    large \emph{open-weight} models. Best result per row in \textbf{bold},
    second best \underline{underlined}. All models evaluated with identical
    harness, prompts, and few-shot settings; ``-DE'' denotes the German variant
    of a benchmark. Aggregates are the harness-level English and German suite
    means (recomputed with GPQA and held-out set excluded for all models; see
    Section~\ref{subsec:contamination}). Nemotron 3 Nano shares the same
    30B-A3B Mixture-of-Experts architecture as \ourmodel{}. Column shading:
    \colorbox{soofipurple}{\,\ourmodel{}\,} (ours),
    \colorbox{archgreen}{\,Nemotron\,} (same architecture),
    \colorbox{openwt}{\,open-weight\,} (Qwen, Ministral, Gemma).}
    \label{tab:base-evals-openweight}
    \resizebox{0.62\linewidth}{!}{
    \begin{tabular}{l
      >{\columncolor{soofipurple}}c
      >{\columncolor{archgreen}}c
      >{\columncolor{openwt}}c
      >{\columncolor{openwt}}c
      >{\columncolor{openwt}}c}
    \toprule
    \textbf{Benchmark} & \textbf{\ourmodel{}} & \textbf{Nemotron 3} & \textbf{Qwen3.5} & \textbf{Ministral 3} & \textbf{Gemma 3} \\
     & \textbf{30B-A3B} & \textbf{Nano 30B-A3B} & \textbf{35B-A3B} & \textbf{14B} & \textbf{27B} \\
    \midrule
    \multicolumn{6}{l}{\textbf{Aggregates}}\\
    English aggregate & 77.3 & 76.7 & \textbf{81.3} & 77.8 & \underline{79.0} \\
    German aggregate & \underline{85.3} & 80.7 & \textbf{87.6} & 84.3 & 85.1 \\
    \midrule
    \multicolumn{6}{l}{\textbf{Code} (pass@1)}\\
    HumanEval & \textbf{73.8} & \underline{72.2} & 67.1 & 60.9 & 60.2 \\
    MBPP & \textbf{70.2} & \underline{67.5} & 65.8 & 58.8 & 66.7 \\
    LBPP & 31.0 & \textbf{38.1} & \underline{32.4} & 27.0 & 22.4 \\
    HumanEval-DE & \underline{65.5} & \textbf{68.8} & 59.5 & 55.5 & 57.7 \\
    MBPP-DE & \textbf{84.2} & \underline{79.9} & 79.0 & 72.0 & 75.6 \\
    \midrule
    \multicolumn{6}{l}{\textbf{Mathematics}}\\
    GSM8K & \underline{86.1} & \textbf{86.5} & 82.6 & 83.7 & 79.9 \\
    GSM8K-Platinum-DE & 87.1 & \underline{87.7} & \textbf{91.2} & 81.3 & 80.7 \\
    Minerva 500 & \underline{79.4} & 64.0 & \textbf{83.0} & 61.8 & 75.7 \\
    Minerva Math-EN & \underline{81.0} & 64.2 & \textbf{82.4} & 61.3 & 73.7 \\
    Minerva MATH-DE & 56.0 & 58.1 & \textbf{76.5} & 55.1 & \underline{65.6} \\
    \midrule
    \multicolumn{6}{l}{\textbf{Knowledge}}\\
    MMLU stem & 75.9 & 74.8 & \textbf{81.9} & \underline{76.4} & 74.7 \\
    MMLU-Pro & 51.4 & 51.6 & \textbf{60.3} & \underline{53.3} & 50.7 \\
    MMLU-Pro-DE & 49.4 & 47.1 & \textbf{56.6} & \underline{50.2} & 47.2 \\
    INCLUDE-DE & \textbf{61.2} & 59.7 & \textbf{61.2} & \underline{60.4} & 57.6 \\
    NaturalQuestions (acc) & 79.0 & 80.3 & \underline{81.0} & 77.1 & \textbf{83.5} \\
    \midrule
    \multicolumn{6}{l}{\textbf{Commonsense and reading comprehension}}\\
    HellaSwag & 64.1 & \underline{65.4} & 65.0 & 61.3 & \textbf{65.9} \\
    HellaSwag-DE & \textbf{74.0} & 66.6 & \underline{71.2} & 60.9 & 69.0 \\
    PIQA & \textbf{85.7} & 84.8 & \underline{84.9} & 82.9 & \underline{84.9} \\
    PIQA-DE & \textbf{91.5} & 85.7 & \underline{87.6} & 81.3 & 86.7 \\
    SocialIQA & \textbf{60.5} & 57.6 & \underline{58.0} & 53.9 & 56.2 \\
    SocialIQA-DE & \textbf{87.6} & 82.7 & \underline{87.2} & 83.0 & 86.7 \\
    SQuAD (EM) & \textbf{87.5} & \underline{87.3} & 80.4 & 82.7 & 85.1 \\
    DROP (EM) & \textbf{66.5} & 64.2 & 62.6 & 62.9 & \underline{65.0} \\
    \midrule
    \multicolumn{6}{l}{\textbf{Reasoning and science}}\\
    BBH (CoT) & 78.8 & 77.8 & \textbf{84.2} & \underline{79.6} & 77.8 \\
    AGIEval & 66.9 & 65.1 & \textbf{71.7} & 67.0 & \underline{68.3} \\
    ARC-Challenge & 90.6 & 89.4 & \textbf{93.1} & \underline{91.6} & \underline{91.6} \\
    \midrule
    \multicolumn{6}{l}{\textbf{German language proficiency}}\\
    GLP-DE & 88.8 & 73.7 & \textbf{94.0} & \underline{89.5} & 88.3 \\
    ARC-Challenge-DE & 92.3 & 91.4 & \textbf{96.6} & \underline{93.5} & 93.3 \\
    \bottomrule
    \end{tabular}}
\end{table}

\begin{figure}[ht]
\centering
\includegraphics[width=\textwidth]{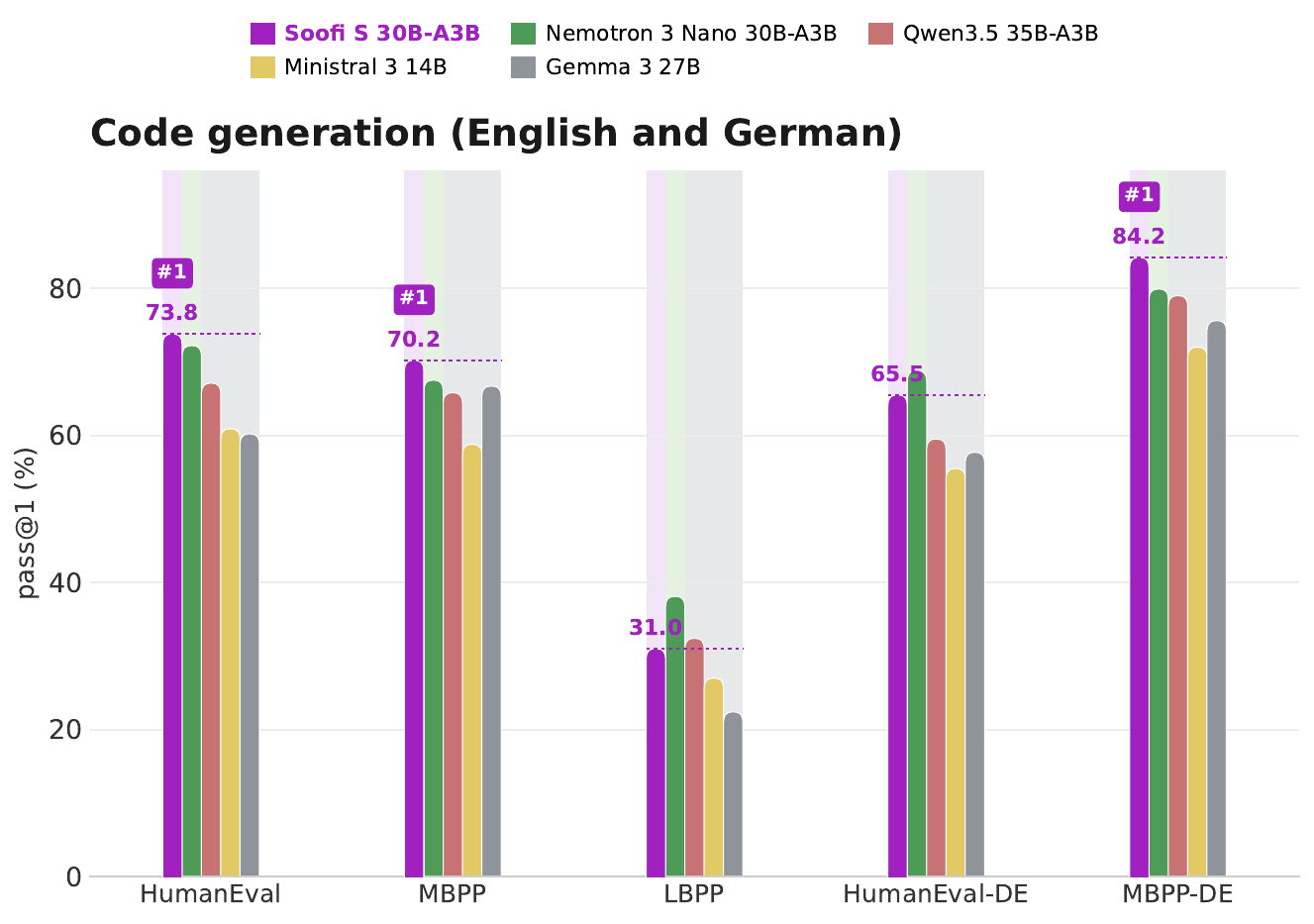}
\caption{Code generation results (pass@1) against Nemotron 3 Nano and large
open-weight models on English and German benchmarks. \ourmodel{} achieves the
best HumanEval, MBPP, and MBPP-DE scores in this comparison. The code
aggregates used in Figure~\ref{fig:eval-overview-openweight} average
HumanEval with MBPP and HumanEval-DE with MBPP-DE; LBPP is reported separately.}
\label{fig:eval-code-openweight}
\end{figure}

\paragraph{Code performance.}
Figure~\ref{fig:eval-code-openweight} reports the code-generation results for
the open-weight comparison. \ourmodel{} sets the best score in this subset on HumanEval~\cite{chen2021humaneval} ($73.8$),
MBPP~\cite{austin2021mbpp} ($70.2$), and MBPP-DE ($84.2$), and is second only to its architectural
reference on HumanEval-DE ($65.5$ vs.\ $68.8$). The German code benchmarks
(problem statements and docstrings in German) show the largest positive margins
for \ourmodel{} in this subset: on MBPP-DE, \ourmodel{} leads the best dense
open-weight model by $8.6$ points. The contamination-aware LBPP benchmark~\cite{matton2024lbpp}
($31.0$) is the one code benchmark where \ourmodel{} trails, behind Nemotron
($38.1$) and Qwen3.5~35B-A3B ($32.4$), while staying ahead of Ministral and
Gemma.

\begin{figure}[ht]
\centering
\includegraphics[width=\textwidth]{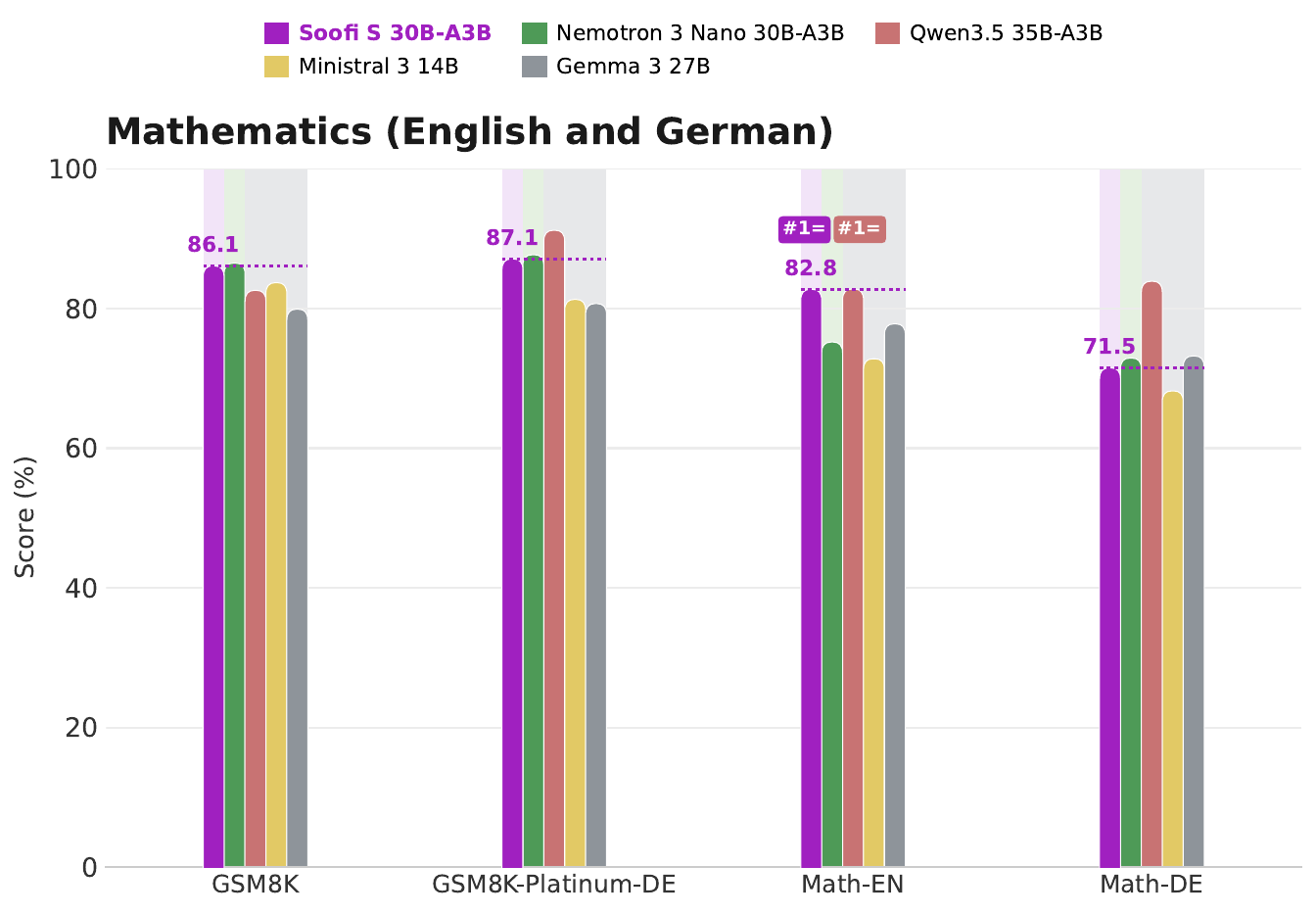}
\caption{Mathematics results against Nemotron 3 Nano and large open-weight
models on English and German benchmarks. \ourmodel{} is second-best on
GSM8K~\cite{cobbe2021training}; on the German side it is essentially level with
its architectural reference on GSM8K-Platinum-DE, while competition-style
Math-DE trails Qwen3.5~35B-A3B.}
\label{fig:eval-math-openweight}
\end{figure}

\paragraph{Mathematics.}
On grade-school reasoning, \ourmodel{} scores $86.1$ on GSM8K~\cite{cobbe2021training} and $87.1$ on
GSM8K-Platinum-DE~\cite{vendrow2025largelanguagemodelbenchmarks}---within
$0.4$ and $4.1$ points of the best open-weight result on the respective tasks
(Figure~\ref{fig:eval-math-openweight}). On competition-style
mathematics~\cite{hendrycks2021math,lewkowycz2022minerva}, \ourmodel{} reaches
$79.4$ on Minerva-500 and $81.0$ on Minerva Math-EN---the second-best results
in this comparison, behind only Qwen3.5~35B-A3B---lifting its Math-EN score to $82.8$,
essentially tied with Qwen3.5~35B-A3B under this suite definition. German
competition mathematics is weaker: on Minerva MATH-DE ($56.0$) \ourmodel{}
trails Qwen3.5~35B-A3B ($76.5$), Gemma~3~27B ($65.6$), and its architectural
reference ($58.1$), leaving Math-DE at $71.5$.

\begin{figure}[ht]
\centering
\includegraphics[width=\textwidth]{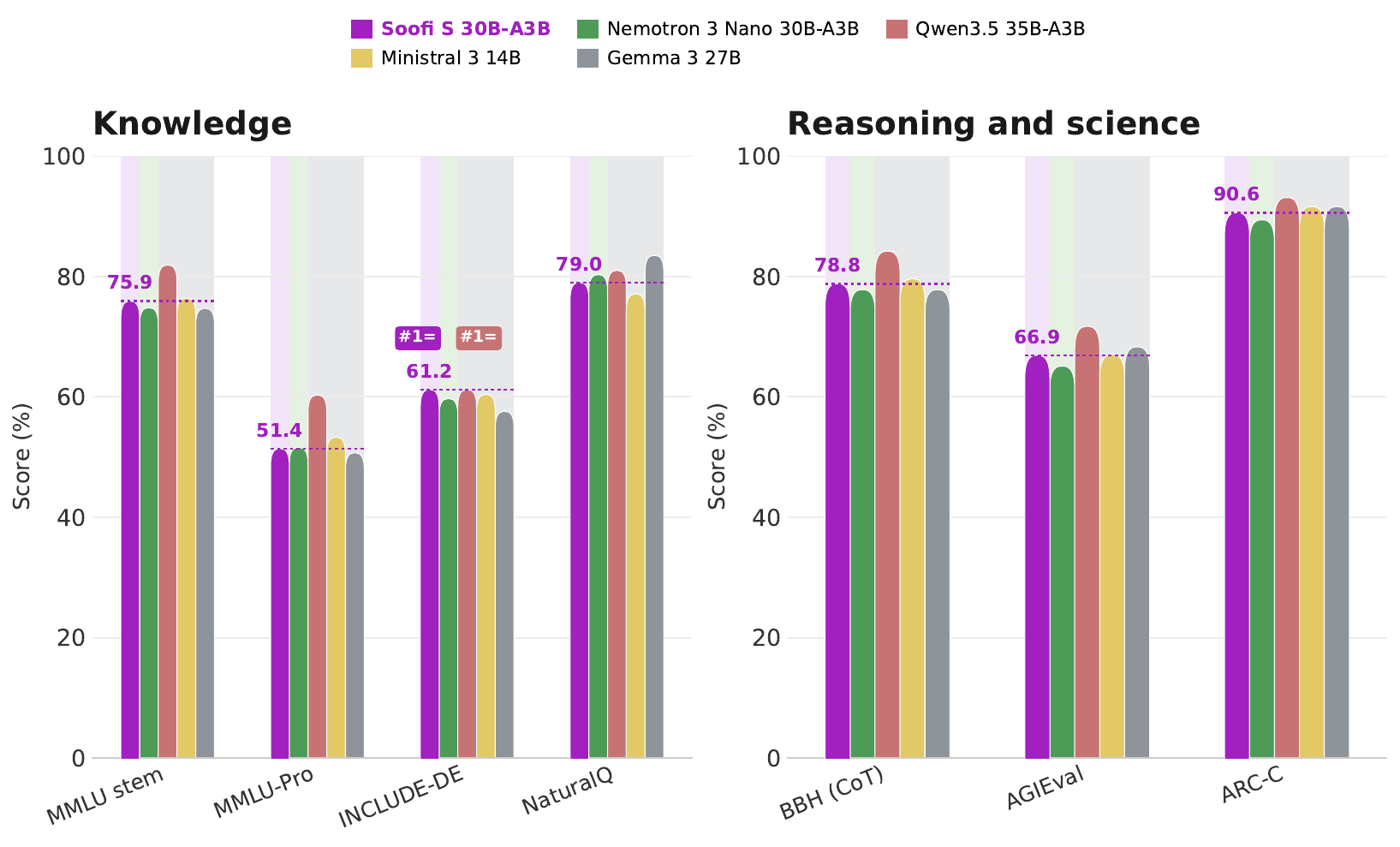}
\caption{Knowledge (left) and reasoning/science (right) benchmarks against
Nemotron 3 Nano and large open-weight models.
\ourmodel{} ties Qwen3.5~35B-A3B on INCLUDE-DE while matching dense 14--27B models on MMLU-STEM, MMLU-Pro, BBH, and AGIEval.}

\label{fig:eval-knowledge-reasoning-openweight}
\end{figure}

\paragraph{Knowledge.}
On broad academic knowledge, \ourmodel{} is competitive with dense models
several times its active size: MMLU-STEM~\cite{hendrycks2021mmlu} $75.9$ and MMLU-Pro~\cite{wang2024mmlupro} $51.4$ sit between
Gemma~3~27B and Ministral~3~14B, and the German MMLU-Pro-DE ($49.4$) shows the
same pattern (Figure~\ref{fig:eval-knowledge-reasoning-openweight}, left). On
INCLUDE-DE~\cite{romanou2024include}---regional, Germany-specific knowledge spanning driving-licence,
social-science, and STEM exams---\ourmodel{} ties Qwen3.5~35B-A3B for the best
score in the table ($61.2$), consistent with the up-weighted native German
data described in Sections~\ref{sec:phase1} and~\ref{sec:phase2}. Open-domain factual
recall as measured by NaturalQuestions~\cite{kwiatkowski2019nq} ($79.0$) narrowly trails the largest dense
models, consistent with storing world knowledge in $3$B active
parameters; we return to this in the Limitations paragraph.

\paragraph{Reasoning and science.}
\ourmodel{} achieves strong general-reasoning performance, with scores of
$78.8$ on BBH~\cite{suzgun2022bbh}, $66.9$ on AGIEval~\cite{zhong2023agieval},
and $90.6$ on ARC-Challenge~\cite{clark2018arc}. These results place it
consistently within a point of Ministral~3~14B
(Figure~\ref{fig:eval-knowledge-reasoning-openweight}, right).

\begin{figure}[ht]
\centering
\includegraphics[width=\textwidth]{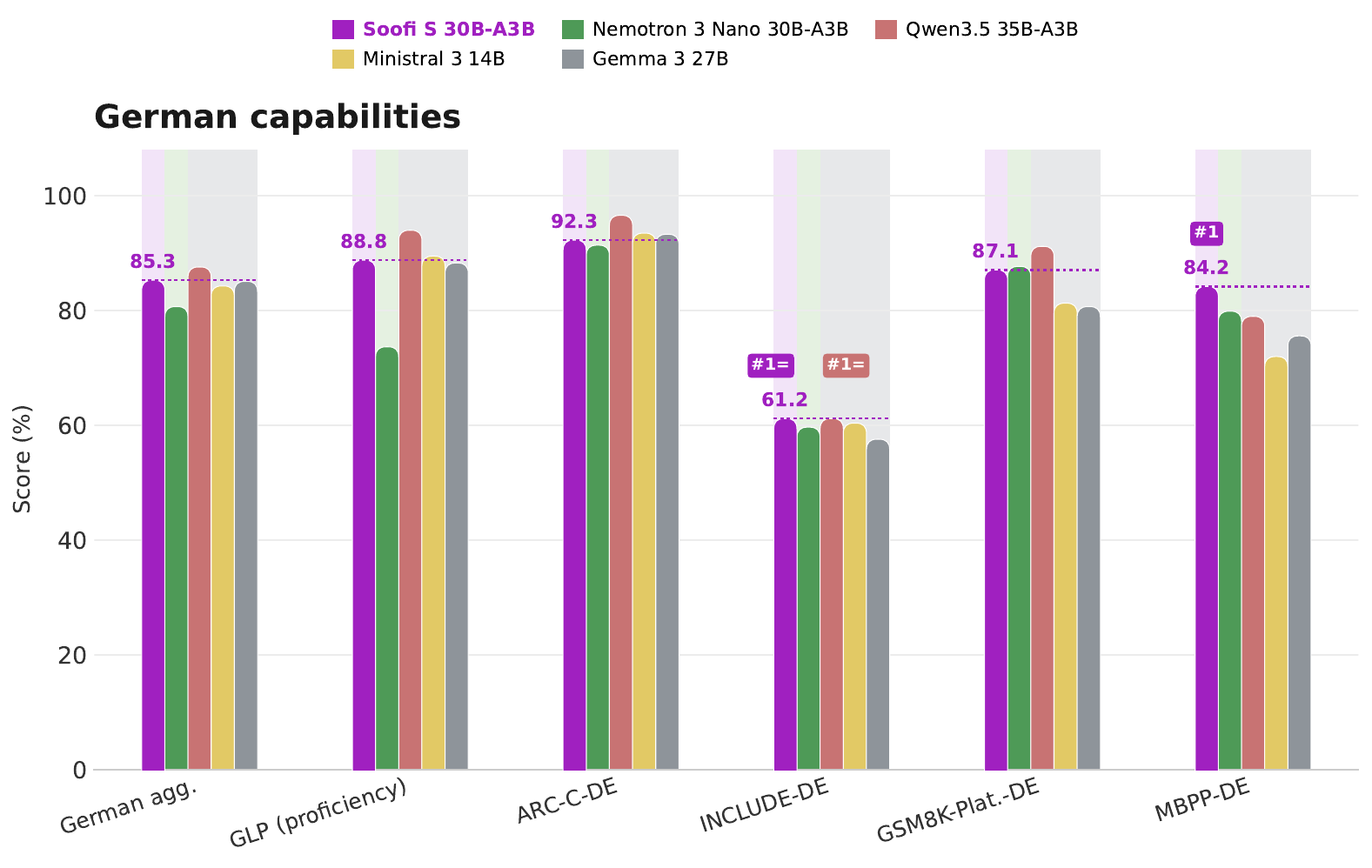}
\caption{German benchmark results against Nemotron 3 Nano and large open-weight models. \ourmodel{} ranks first on MBPP-DE, ties Qwen3.5~35B-A3B on INCLUDE-DE, and improves the German aggregate over the architecture-identical Nemotron baseline by $+4.6$ points.}
\label{fig:eval-german-openweight}
\end{figure}

\paragraph{German capabilities.}
Figure~\ref{fig:eval-german-openweight} reports the German benchmarks for the
open-weight comparison. \ourmodel{} ranks first on MBPP-DE ($84.2$) and ties
Qwen3.5~35B-A3B on INCLUDE-DE ($61.2$). It is close to the larger dense models
on the German aggregate ($85.3$), behind Qwen3.5~35B-A3B ($87.6$) and ahead of
Gemma~3~27B ($85.1$), Ministral~3~14B ($84.3$), and Nemotron~3~Nano ($80.7$).
Relative to Nemotron, \ourmodel{}
improves GLP-DE by $+15.1$, and INCLUDE-DE by $+1.5$ while remaining slightly below Nemotron on GSM8K-Platinum-DE and HumanEval-DE.

\begin{figure}[ht]
\centering
\includegraphics[width=0.82\textwidth]{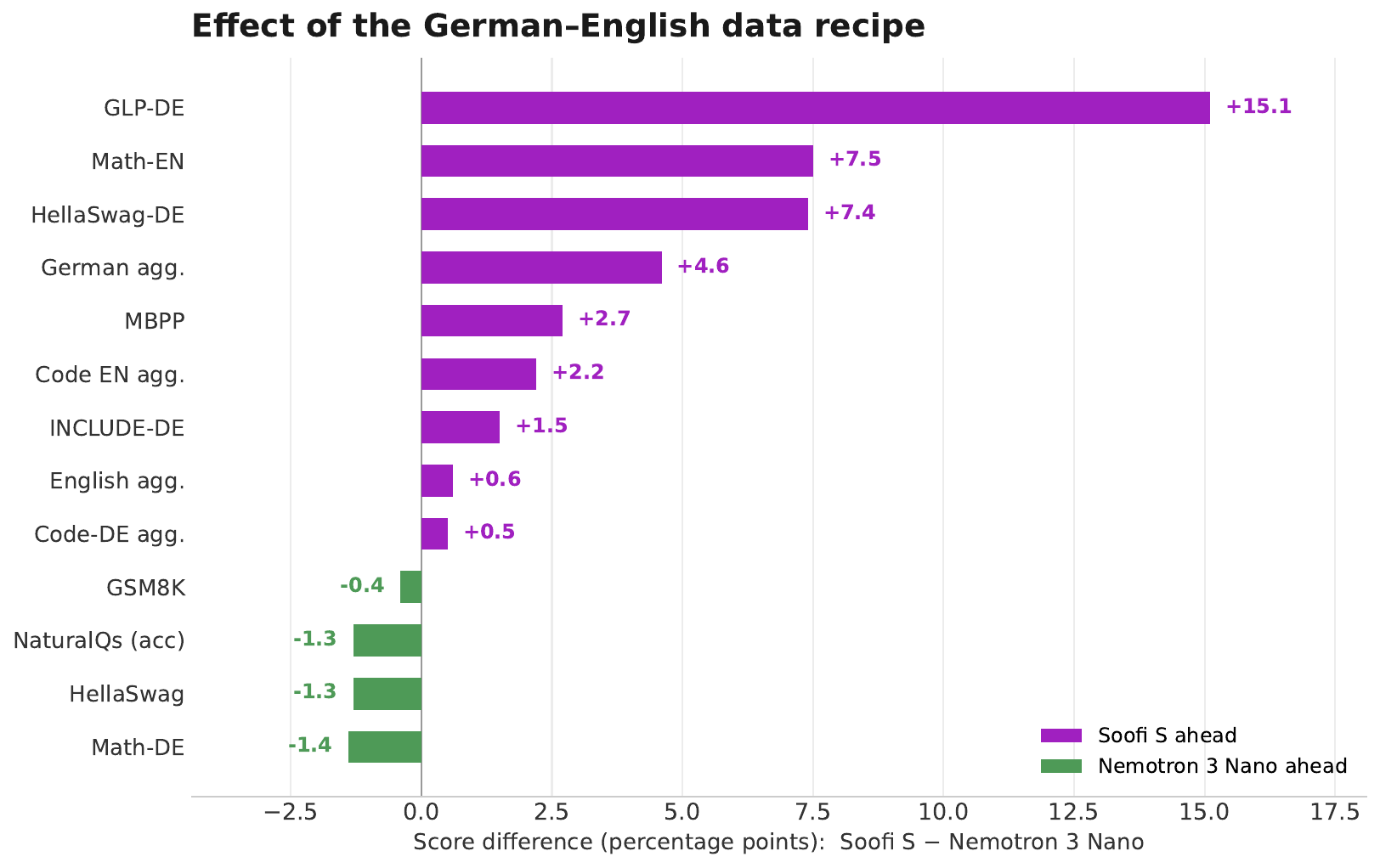}
\caption{Per-benchmark score difference between \ourmodel{} and the
architecture-identical Nemotron~3~Nano~30B-A3B, isolating the effect of the
German--English data recipe. The recipe yields large German gains (GLP-DE
$+15.1$, HellaSwag-DE $+7.4$, German aggregate $+4.6$) without sacrificing
overall English performance (EN agg.\ $+0.6$, Math-EN $+7.5$).}

\label{fig:eval-delta-nemotron}
\end{figure}

\paragraph{Effect of the German--English recipe.}
Since \ourmodel{} shares its architecture with Nemotron~3 Nano~30B-A3B, the
pairwise comparison in Figure~\ref{fig:eval-delta-nemotron} isolates the
effect of our data recipe from architecture. The German interventions deliver
large, targeted gains---GLP-DE $+15.1$, German aggregate $+4.6$, INCLUDE-DE
$+1.5$, German code $+0.5$---without the regression on English that monolingual specialisation usually incurs: the English aggregate
\emph{improves} by $+0.6$, code by $+2.2$, and the Math-EN aggregate by $+7.5$. The main regressions are German competition mathematics (Math-DE aggregate $-1.4$), open-domain factual recall (NaturalQuestions $-1.3$), $-0.4$ on GSM8K, and $-1.3$ on HellaSwag, consistent with
German tokens displacing a portion of English web knowledge.
Overall, the Nemotron comparison indicates that the German-focused data recipe
substantially improves German capability while preserving or improving the
English aggregate, code and reasoning scores reported here.

\paragraph{Limitations.}

We report four caveats in the interest of full transparency. First, the Minerva
mathematics~\cite{hendrycks2021math,lewkowycz2022minerva} scores required a
corrected evaluation protocol. Our initial harness configuration used a
\texttt{\textbackslash n\textbackslash n} stop sequence and a generation limit
of $1{,}024$ tokens, which truncated long chain-of-thought solutions before a
final answer was produced and collapsed the Minerva scores of the strongest
models (\ourmodel{} measured $9.8$ on Minerva-500 under this configuration,
and the architecture-identical Nemotron~3~Nano $5.6$). Removing the stop
condition and raising the generation limit to $4{,}096$ tokens yields the
results reported in Tables~\ref{tab:base-evals-opensource} and~\ref{tab:base-evals-openweight} and the figures ($79.4$ and
$81.0$ for \ourmodel{} on Minerva-500 and Minerva Math-EN); all models were
re-evaluated under the same corrected configuration. Second, competition-style
mathematics in German remains the clearest capability gap to the frontier: on
Minerva MATH-DE ($56.0$) \ourmodel{} trails Qwen3.5~35B-A3B ($76.5$) and
Gemma~3~27B ($65.6$), even though its German grade-school mathematics
(GSM8K-Platinum-DE, $87.1$) is essentially level with the architectural
reference.
Third, open-domain factual recall remains capacity-limited: NaturalQuestions
($79.0$) trails the largest dense baselines (Gemma~3~27B $83.5$), consistent
with storing world knowledge in 3B active parameters; we expect
retrieval-augmented deployment to close this gap in practice.
Finally, the GPQA contamination incident and its remediation are documented in \Cref{subsec:contamination}.

\subsection{GPQA Contamination Disclosure}
\label{subsec:contamination}
We regret to disclose a benchmark-contamination incident discovered after the
initial version of this report. The QA-base dataset (\Cref{sec:phase2}) was
designed to contain paraphrased \emph{training} splits of 25 standard NLP
benchmarks in English and German. Shortly after release, community members
examining the published dataset correctly identified that it additionally
contained paraphrased items from the
GPQA~\cite{rein2023gpqagraduatelevelgoogleproofqa} \emph{evaluation} set,
including the GPQA-Diamond subset, in both English and machine-translated
German. We subsequently verified this finding against our construction
pipeline and confirmed it.

\paragraph{Root cause.}
GPQA is distributed without a training split: its entire evaluation set is
published as a single split that carries the default Hugging Face label
\texttt{train}. The split-selection step of the QA-base construction
pipeline keyed on split \emph{names} rather than split \emph{semantics}, and
therefore ingested the evaluation set under the label \texttt{train} instead
of excluding the benchmark. We state this as an explanation of the failure
mode, not as an excuse: verifying that a split labeled \texttt{train} is in
fact a training split was our responsibility.

\begin{figure}[ht]
\centering
\includegraphics[width=0.82\textwidth]{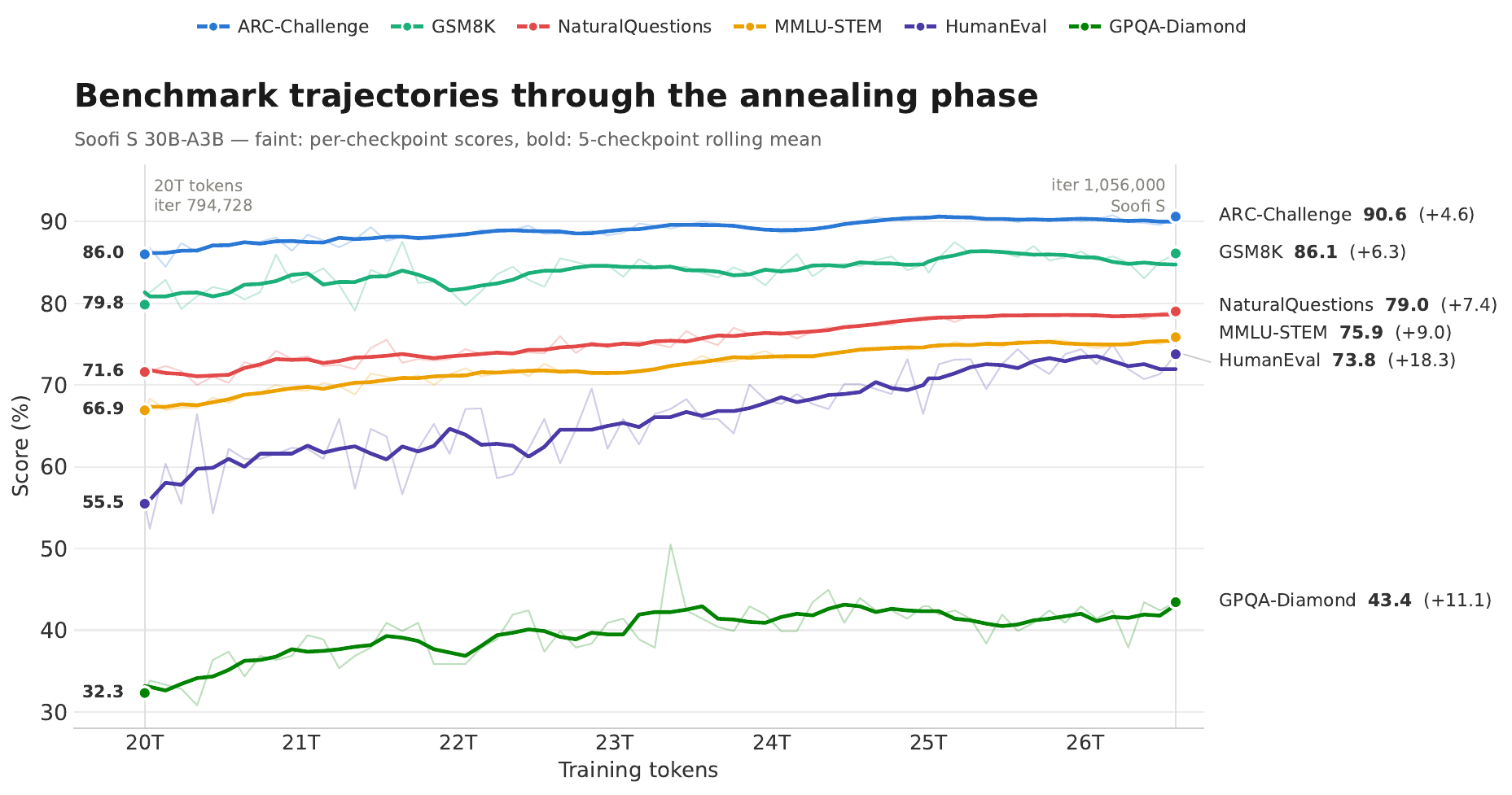}
\caption{Benchmark trajectories across annealing checkpoints. GPQA-Diamond
rises gradually ($32.3 \rightarrow 43.4$), without a discontinuity at the
introduction of the contaminated data and within the range of gains on other
benchmarks under the same protocol (e.g., HumanEval, MMLU-STEM). We include
this to document how the contamination remained invisible in routine
training monitoring: paraphrased, heavily diluted derivatives produce slow
inflation indistinguishable at the trajectory level from genuine capability
gains.}
\label{fig:annealing-trajectories}
\end{figure}

\paragraph{Training dynamics of the contaminated benchmark.}
For transparency, we also report how the contamination appeared, or rather
did not appear, in the training signals we observed
(\Cref{fig:annealing-trajectories}). Throughout the annealing phase we
tracked benchmark trajectories across checkpoints as part of routine
training monitoring. A sudden, disproportionate jump on a single benchmark
would have stood out and prompted investigation; no such signal occurred.
GPQA-Diamond improved gradually ($32.3 \rightarrow 43.4$), without a
discontinuity at the introduction of the contaminated data and within the
range of gains observed on other benchmarks under the same protocol (e.g.,
HumanEval $+18.3$, MMLU-STEM $+9.0$). In retrospect, this is what
paraphrased, machine-generated derivatives diluted to a small fraction of
the pool look like: slow inflation that is indistinguishable, at the
trajectory level, from genuine capability gains. We emphasize that we do not
consider trajectory monitoring a contamination defense, nor did we treat it
as one. We report it to document why the issue remained invisible to us
until the community's inspection of the data itself surfaced it.

\paragraph{Full audit scope.}
Prompted by this discovery, we re-audited all QA-base constituents at the
level of their source loaders and data files. The same failure mode,
evaluation-only benchmarks whose sole published split carries the label
\texttt{train} or \texttt{validation}, affected four constituents in
total: GPQA, TruthfulQA, BLiMP, and the Inverse Scaling tasks. Of these, only
GPQA appears in our evaluation suite; we do not evaluate or report
TruthfulQA, BLiMP, or Inverse Scaling anywhere in this report. We nevertheless
disclose all four so that independently obtained scores on these benchmarks
for \ourmodel{} are not misinterpreted as capability.

\paragraph{Remediation.}
This revision applies the following corrections. First, GPQA-Diamond and
GPQA-Diamond-DE are removed from all tables and figures. Second, the English and German harness suite means in \Cref{fig:serving-efficiency} are recomputed without GPQA and held-out group for \emph{all} models symmetrically, so cross-model comparisons remain fair; the relative rankings reported elsewhere in \Cref{sec:base-evals} are unaffected by this
change. The English and German suite aggregates in Tables \ref{tab:base-evals-opensource}--\ref{tab:base-evals-openweight} are likewise recomputed without GPQA, and the held-out suites of the initial version, which contained GPQA, are withdrawn. Third, a corrected
QA-base\footnote{\url{https://huggingface.co/datasets/AIML-TUDA/QA-base}}
release with the items derived from all four affected benchmarks removed
accompanies this revision, and the dataset card and model card document the
incident. Fourth, split selection in our data pipelines is no longer
name-based: every constituent is now validated against a per-dataset
allowlist derived from its original publication, and final training mixtures
are screened against our full evaluation suite via $n$-gram overlap before
training.

\subsection{Serving Efficiency}
\label{subsec:efficiency}
Active parameter count is only a proxy for inference cost. In deployment, the relevant quantity is sustained throughput at realistic context lengths and request concurrency, where decoding is often limited by memory bandwidth: each generated token must stream the active model weights and read the attention cache for every active sequence. This is the regime targeted by the hybrid Mamba--MoE architecture. In \ourmodel{}, only 6 of 52 layers maintain a KV cache, with 2 KV heads each, while the 23 Mamba-2 layers carry a fixed-size recurrent state. Consequently, the incremental attention-cache footprint is only about 6 KB per token per sequence, which is $11$--$53{\times}$ lower than for the dense models in our comparison. As context length grows, only this small attention component scales with sequence length; the Mamba recurrent state remains constant-size.

\begin{figure}[ht]
  \centering
  \includegraphics[width=0.85\linewidth]{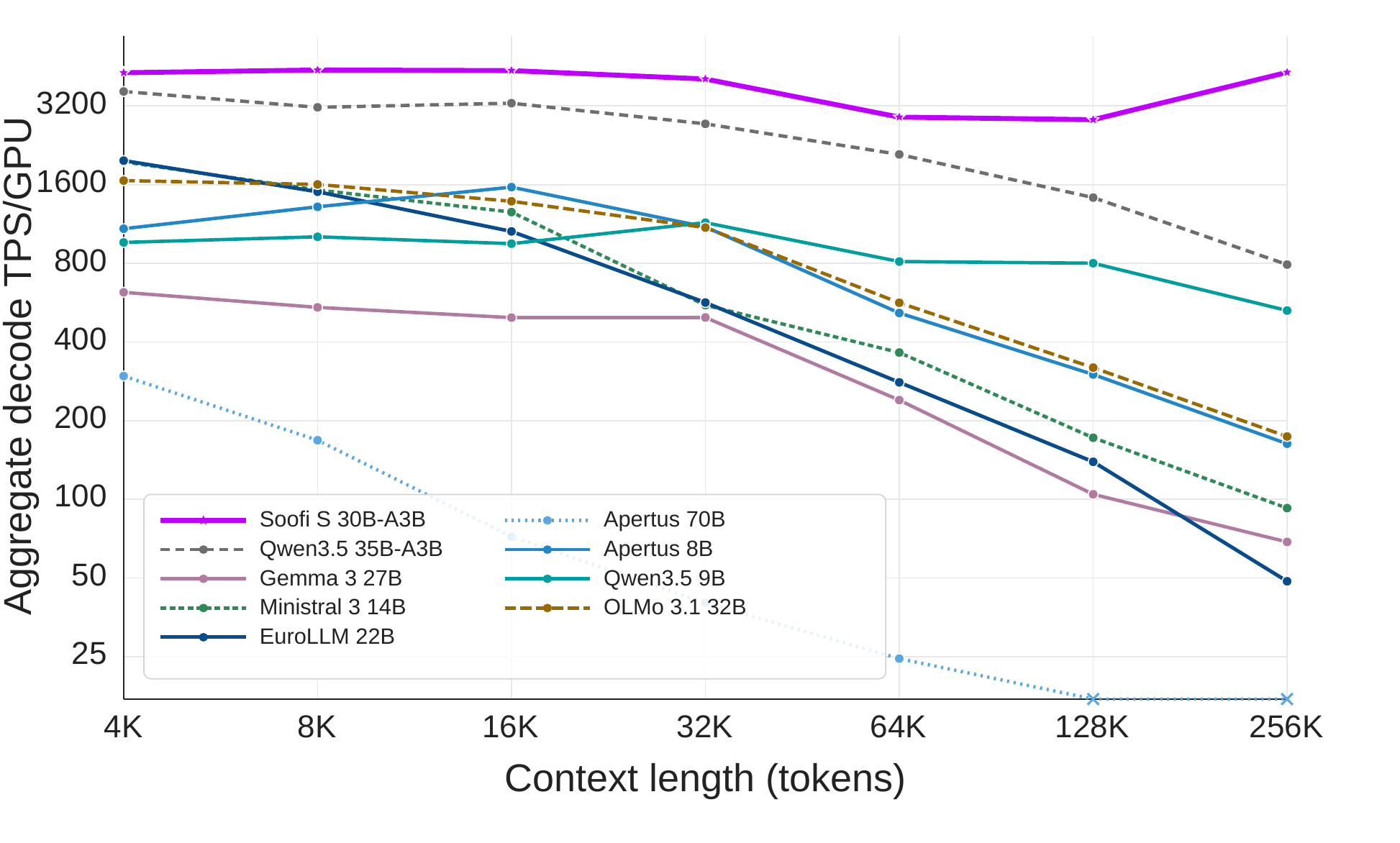}
  \caption{\textbf{Long-context decode-throughput scaling.} Measured aggregate
  decode TPS/GPU as a function of input context length under the TP=1,
  single-B200, batch-32 latency-subtraction protocol.}
  \label{fig:serving-throughput-scaling}
\end{figure}

Figures~\ref{fig:serving-efficiency} and~\ref{fig:serving-throughput-scaling}
quantify this effect using measured aggregate decode tokens per second (TPS) per
GPU. We use a TP=1, single-B200 vLLM latency-subtraction protocol. For each context length, we measure fixed-batch latency as a function of output length $O$. Let $t(O)$ denote this latency. We estimate aggregate decode TPS per GPU by subtracting the $O_{\mathrm{short}}=1$ run from the $O_{\mathrm{long}}=1024$ run:
\begin{equation}
\label{eq:decode-tps}
\mathrm{TPS}_{\mathrm{decode}}^{\mathrm{agg/GPU}}
=
\frac{
  B\left(O_{\mathrm{long}} - O_{\mathrm{short}}\right)
}{
  N_{\mathrm{GPU}}
  \left[
    t\left(O_{\mathrm{long}}\right)
    -
    t\left(O_{\mathrm{short}}\right)
  \right]
}
\end{equation}
Here $O_{\mathrm{short}}=1$, $O_{\mathrm{long}}=1024$,
$N_{\mathrm{GPU}}=1$, and $B=32$ for all plotted measurements. This
subtraction removes most prompt-prefill cost and isolates the per-GPU
cache-bandwidth pressure that tensor parallelism can otherwise mask. At a 40K context and batch size 32, \ourmodel{} sustains a measured aggregate decode rate of 4.82k TPS/GPU, which is $9.2{\times}$ higher than Ministral~3~14B, while fitting the weights and all 32 sequence states on a single GPU. In contrast, Apertus~70B does not fit on a single GPU under the current fixed-256K, TP=1 setup.

The same measurements also expose the prefill/TTFT side of the hybrid
architecture: $t(O_{\mathrm{short}})=t(1)$ is the measured batch-32 latency to process the prompt and produce the first output token, i.e., the TTFT-like measurement in this protocol. At 40K context, \ourmodel{} reaches $t(1)=22.7$ s, compared with $71.9$ s for Ministral~3~14B, $92.9$ s for
Gemma~3~27B, $101.7$ s for OLMo~3~32B, and $213.4$ s for the
Qwen3~32B dense control. At 256K context, \ourmodel{} remains the fastest
complete sweep in our batch-32 protocol ($372.7$ s), ahead of
Qwen3.5~35B-A3B ($579.2$ s), Gemma~3~27B ($999.4$ s), OLMo~3~32B
($1{,}487.3$ s), Ministral~3~14B ($2{,}058.9$ s), and Qwen3~32B dense
($6{,}428.6$ s). This is the prefill analogue of the decode-cache advantage: most sequence mixing in \ourmodel{} is carried by Mamba layers rather than full attention, so long-context TTFT grows much more favorably than in full-attention dense baselines.

Figure~\ref{fig:serving-throughput-scaling} shows that the decode gap widens with context length: dense-model throughput decreases as KV-cache reads dominate decoding, whereas \ourmodel{} remains nearly flat from 4K to 256K in the current measurement snapshot, with endpoint aggregate decode rate changing from 4.29k to 4.30k TPS/GPU and no point more than ${\sim}34\%$ below the 4K value. Among the comparison models, only Qwen3.5, a Gated-DeltaNet hybrid~\cite{qwen35_2026}, exhibits similar scaling behavior. In its published 9B configuration, 8 of 32 layers remain full-attention layers, with 4 KV heads of dimension 256, corresponding to a per-sequence cache of 32 KB per token, or
$5.3{\times}$ that of \ourmodel{}. For the 35B-A3B variant, we measure 2.60k aggregate decode TPS/GPU at 40K under the same protocol, which is
$1.9{\times}$ lower than \ourmodel{}. This qualitative separation is consistent with NVIDIA's Nemotron-H measurements, where a related hybrid engine achieved $3.3{\times}$ higher throughput than Qwen3-30B-A3B on production inference engines~\cite{nvidia2025nemotron3}.

\section{Related Work}
\label{sec:related}
We situate \ourmodel{} with respect to five strands of prior work: open
language-model pretraining, data acquisition and filtering, training curricula,
efficient sparse and hybrid architectures, and multilingual European foundation
models. We focus on base-model pretraining; instruction tuning, preference
optimization, and reasoning-specific reinforcement learning are complementary
and outside the scope of this report.

\paragraph{Open language-model pretraining.}
The modern causal language-modeling paradigm was established by GPT-style
pretraining: GPT-2 showed that a decoder-only Transformer trained with
next-token prediction on large web text could perform a wide range of tasks in a
zero-shot setting, and GPT-3 demonstrated the emergence of strong in-context
few-shot behavior at a substantially larger scale
\cite{radford2019language,brown2020language}. Subsequent scaling-law work made model size, data size, and compute budget explicit design variables, while the
Chinchilla analysis shifted attention from parameter count alone toward
data--compute balance and the importance of training sufficiently long on enough
tokens \cite{kaplan2020scaling,hoffmann2022training}. These results motivate the central design principle of recent pretraining recipes: capability is a
joint function of architecture, token budget, data quality, curriculum, and
serving cost, not merely of total parameters.

Open releases have followed a separate but related trajectory. OPT released a
family of decoder-only models together with code and a detailed account of
training infrastructure, GPT-NeoX-20B provided a permissively released large
autoregressive model trained on the Pile, and Pythia made training dynamics
inspectable by releasing many intermediate checkpoints and a fixed data order
\cite{zhang2022opt,black2022gptneox,biderman2023pythia,gao2020pile}. BLOOM
extended this community-scale approach to multilingual pretraining, combining a
large open-access model with the ROOTS multilingual corpus and a collaborative
development process \cite{bigscience2022bloom}. These efforts established the
scientific value of releasing more than a final checkpoint: intermediate states,
training code, data documentation, and evaluation recipes make it possible to
study memorization, bias, scaling behavior, and data effects.

The broader open-weight ecosystem has since produced increasingly strong
general-purpose baselines, including LLaMA and Llama~2, Falcon, Mistral~7B,
Mixtral, Qwen, Gemma, and recent Mistral/Ministral releases
\cite{touvron2023llama,touvron2023llama2,almazrouei2023falcon,jiang2023mistral7b,jiang2024mixtral,qwen3_2025,qwen35_2026,gemma3_2025,mistral2025mistral3,mistral2025small31}.
However, many such models are open-weight rather than fully reproducible: the
weights and inference code are available, but the exact data mixture, filtering
pipeline, training order, logs, intermediate checkpoints, and rejected data
sources are often missing. OLMo, OLMoE, OLMo~3, and Apertus mark a stronger
notion of openness by releasing training data or data recipes, code, model
artifacts, and broader documentation for scientific audit and reuse
\cite{groeneveld2024olmo,muennighoff2024olmoe,olmo3_2025,hernandezcano2025apertus}.
\ourmodel{} follows this fully open line, but differs in scope and
architecture: it is a German--English base model trained with exact per-source
token accounting on a sparse hybrid Mamba--Transformer MoE backbone, rather
than a dense general-purpose or broadly multilingual Transformer.

\paragraph{Pretraining data acquisition and filtering.}
Pretraining data pipelines have evolved from relatively simple web-crawl
selection toward multi-stage corpus construction. Early influential corpora such as WebText, C4, and the Pile used web-scale collection, heuristic filtering,
deduplication, and source balancing to turn noisy internet data into training
data suitable for language modeling
\cite{radford2019language,raffel2020exploring,gao2020pile}. BLOOM's ROOTS
corpus added a multilingual community-curated data effort, combining web data
with manually selected sources across many languages \cite{bigscience2022bloom}.
More recent corpora make the filtering pipeline itself a central contribution:
FineWeb and Dolma document large-scale web cleaning, deduplication, and mixture
construction; Nemotron-CC provides quality-tiered Common Crawl data with
synthetic and translated variants; and JQL shows that language-model-based
quality judgments can be distilled into multilingual data filters that transfer
across languages more robustly than English-centric heuristics
\cite{penedo2024fineweb,soldaini2024dolma,su2024nemotroncc,ali2025judgingqualitylanguagesmultilingual}.

Domain-specific acquisition has become equally important. PDF- and OCR-derived
corpora such as FinePDFs and OlmoOCR add long-form reports, books, papers, and
educational material that are underrepresented in HTML-only crawls
\cite{finepdfs2025,poznanski2025olmocr}. Code corpora such as Nemotron-Pretraining-Code-v1/v2 and SwallowCode complement web-extracted code with repository-derived or rewritten
programming data \cite{nvidia2025nemotron3, nvidia2025nvidianemotronnano2,fujii2025swallowcode}. Mathematical
data pipelines such as Nemotron-CC-Math and UltraData-Math target equation-rich
documents and problem-solving text, where generic web extraction often destroys
structure \cite{karimimahabadi2025nemotronccmath,ultradatamath2025}. For German,
large multilingual crawls and national resources such as HPLT, German Commons,
and KletterMix are especially relevant because high-quality native German tokens
are scarcer than English web tokens
\cite{burchell2025hplt2,oepen2025hplt3,gienapp2025germancommons,kraus2026klettermixclimbinghighqualitygerman}.
\ourmodel{} builds on these trends but reports the corpus at finer granularity: for every phase, we disclose source identifiers, raw tokens, epoch multipliers,
effective token counts, and sources considered but excluded. This makes the
data mixture auditable in a way that aggregate token counts cannot provide.

\paragraph{Training curricula and optimization recipes.}
Large-scale pretraining recipes increasingly separate the problem of collecting
tokens from the problem of ordering and weighting them. Compute-optimal scaling
results motivate training smaller or sparse models for more tokens when
inference efficiency matters, while modern open reports show that curriculum
phase boundaries, learning-rate schedules, and annealing mixtures can have large
effects on downstream behavior
\cite{hoffmann2022training,hu2024minicpm,hagele2024scaling,nvidia2025nemotron3}.
A common pattern is to allocate early training to broad coverage and later
training to higher-quality or skill-focused data, often under a
Warmup--Stable--Decay schedule. OLMo-style releases emphasize logging,
intermediate checkpoints, and reproducible recipes; Nemotron-style recipes
emphasize quality-tiered web data, synthetic skill data, and high-quality
annealing \cite{groeneveld2024olmo,olmo3_2025,nvidia2025nemotron3}.
\ourmodel{} follows the same high-level philosophy but adapts it to a
German--English target: Phase~1 maximizes diversity, Phase~2 concentrates
high-quality web, code, mathematics, reasoning, SFT-formatted, and German data,
and Phase~3 extends context length with document-length buckets. The difference is the explicit bilingual reallocation of token budget and the disclosure of the exact realized mixtures.

\paragraph{Efficient sparse and hybrid architectures.}
Most open LLMs remain dense Transformers, but dense attention becomes costly at
long context and high concurrency because decoding must repeatedly read both
model weights and the per-sequence KV cache. Sparse Mixture-of-Experts models
address the weight side of this problem by increasing total capacity while
activating only a small expert subset per token. Foundational MoE work,
Switch Transformers, DeepSeekMoE, and fine-grained MoE studies show that sparse
routing can improve the capability--compute trade-off when routing and load
balancing are stable
\cite{shazeer2017moe,fedus2022switch,dai2024deepseekmoe,krajewski2024granularity}.
OLMoE demonstrates that this sparse route can also be made fully open at
smaller active-parameter scales \cite{muennighoff2024olmoe}.

A complementary line of work reduces the sequence-state cost of attention.
Mamba-2 and related state-space models replace much of the quadratic attention
machinery with recurrent state updates, yielding linear-time sequence mixing and
a near-constant state during decoding \cite{dao2024mamba2}. Nemotron-H and
Nemotron~3 combine Mamba-style sequence mixing, sparse attention, and MoE layers
to obtain strong long-context serving efficiency, while Qwen3.5 explores a
different hybrid sequence-modeling path with Gated DeltaNet layers
\cite{nvidia2025nemotronh,nvidia2025nemotron3,qwen35_2026}. \ourmodel{} adopts
the Nemotron-style 30B-A3B hybrid Mamba--Transformer MoE design, but evaluates
it under a distinct sovereign bilingual pretraining recipe. This makes the
architectural comparison unusually clean: relative to Nemotron~3~Nano, gains and
trade-offs can largely be attributed to data mixture, German up-weighting,
annealing, and long-context continuation rather than to a different backbone.

\paragraph{Multilingual and European language models.}
Multilingual pretraining aims to reduce the English bias of large language
models, but it requires careful allocation of model capacity and high-quality
tokens across languages. BLOOM was an early large-scale open-access
demonstration of multilingual decoder-only pretraining
\cite{bigscience2022bloom}. More recent European efforts add requirements
around sovereignty, transparency, language coverage, and regulatory
compatibility. Teuken-7B targets the official EU languages with a European
multilingual tokenizer and a large non-English data share
\cite{ali2024teuken}. EuroLLM develops a family of European multilingual
models across several scales, with multilingual data filtering, tokenizer
design, and evaluation as central components
\cite{martins2024eurollm,martins2025eurollm9btechnicalreport,ramos2026eurollm22btechnicalreport}.
Salamandra and the subsequent ALIA family focus on European and Iberian language modeling~\cite{gonzalezagirre2025salamandra}. Apertus
emphasizes fully open and compliant multilingual foundation models, and
OpenEuroLLM extends this direction as a coordinated European initiative
\cite{hernandezcano2025apertus,openeurollm2025}.

\ourmodel{} is complementary to these broad-coverage efforts. Rather than optimizing for many languages at once, it studies a narrower German--English setting in which the data mixture, annealing phase, and evaluation suite are designed around bilingual depth. Broad European models address coverage across languages, while \ourmodel{} tests how much capability and efficiency can be gained when one bilingual deployment setting is given a dedicated, fully documented pretraining recipe.

\paragraph{Positioning of \ourmodel{}.}
The closest architectural reference for \ourmodel{} is Nemotron~3~Nano, because
both use a 30B-A3B hybrid Mamba--Transformer MoE architecture
\cite{nvidia2025nemotron3}. The closest openness references are OLMo, OLMoE,
OLMo~3, and Apertus, because they emphasize releases that enable audit and
reconstruction rather than merely inference
\cite{groeneveld2024olmo,muennighoff2024olmoe,olmo3_2025,hernandezcano2025apertus}.
The closest European language references are EuroLLM, Teuken, Salamandra,
Apertus, and OpenEuroLLM
\cite{martins2024eurollm,martins2025eurollm9btechnicalreport,ramos2026eurollm22btechnicalreport,ali2024teuken,gonzalezagirre2025salamandra,hernandezcano2025apertus,openeurollm2025}.
\ourmodel{} combines these lines in a configuration not covered by prior work:
a fully documented European pretraining run, a German--English data curriculum
with exact per-source accounting, and a sparse hybrid architecture designed for
long-context, high-concurrency serving. The resulting model fills a gap between
broadly multilingual European sovereignty efforts and efficient international
open-weight baselines: it asks whether a sovereign model can be simultaneously
open, bilingual-depth-oriented, and competitive in capability per active
parameter.

\section{Conclusion}
\label{sec:conclusion}

We presented \ourmodel{} 30B-A3B, a sovereign, open-source MoE
hybrid Mamba--Transformer foundation model for German and English. Built on the
Nemotron~3~Nano architecture---52 layers combining Mamba-2, Grouped-Query
Attention, and granular MoE layers that activate roughly 3B of ${\sim}30$B
parameters per token---\ourmodel{} was pretrained on approximately 27 trillion
tokens under a three-phase Warmup--Stable--Decay curriculum: 20T tokens of
diverse, quality-tiered pretraining, ${\sim}6.6$T tokens of high-quality
annealing, and a length-bucketed long-context phase extending the usable
context to 1M tokens. Throughout, German was deliberately up-weighted---to
$7.2\%$ of the stable phase and $15.32\%$ of the annealing mixture, more than
triple the multilingual share of the reference recipe---realizing the design
goal of a German--English champion rather than a thinly spread multilingual
model.

The result is a model that reaches the capability frontier at a fraction of the
inference cost of dense alternatives. On a unified evaluation of \nummodels{}  open base
models (\Cref{sec:base-evals}), \ourmodel{} achieves the best English and German code aggregates among the measured models (HumanEval/MBPP averages, with LBPP reported separately), a Math-EN score essentially tied with the strongest model, second-best scores on
GSM8K~\cite{cobbe2021training} and the English Minerva benchmarks
(Minerva-500 and Minerva Math-EN), a tied-best INCLUDE-DE score, and English and German
aggregates that match dense 14--27B models---all while activating only 3B
parameters per token. It outperforms every European sovereign baseline in our comparison---including those an order of magnitude larger in active
parameters---matching or outperforming them on every German benchmark in the suite, often by 10--30 points (\Cref{fig:serving-efficiency}, \Cref{fig:eval-german-opensource}, and~\Cref{fig:eval-german-openweight}). To our
knowledge, this makes \ourmodel{} the first European sovereign model to sit on
the same capability-per-active-parameter frontier as the strongest
international open-weight releases, the strongest open German base model
in its inference-cost class, and the strongest fully open base model in our evaluation on both English and German aggregates.

Equally central to this work is \emph{how} the model is released. Trained end-to-end on the German Industrial AI Cloud by a consortium of German
research institutions, \ourmodel{} ships not only weights but
the complete set of artifacts needed to audit and rebuild it: the full
per-source token accounting of all three pretraining phases (including sources
we evaluated and excluded), every hyperparameter and learning-rate stage---
including a discarded final annealing stage, reported for completeness---and
the training and evaluation code under permissive licenses, with licensed data sources documented through aggregate statistics and exact mixture accounting rather than redistributed raw text.
We hope this level of transparency moves the open ecosystem from open-weight
toward genuinely open-source, and provides a reproducible template for other
language communities seeking capable, efficient, sovereign foundation models.

Future work will extend \ourmodel{} along three axes: open post-training (SFT
and large-scale RL) toward instruct and reasoning variants (including modular reasoning in the spirit of FlexOlmo and Bar~\cite{shi2025flexolmo,morrison2026bar}),
broader and deeper
German evaluation suites, and continued scaling of the high-quality German data
pipeline that this release identified as the principal bottleneck for further
gains.

\section*{Acknowledgments}
This work was supported by the German Federal Ministry for Economic Affairs and Energy (BMWE) in the context of IPCEI-CIS and 8ra through ``Soofi: Souveräne KI für Europa'' (grant number 13IPC040A-J). Parts of it have benefited from the hessian.AI Service Center (funded by the Federal Ministry of Research, Technology and Space, BMFTR, grant no. 16IS22091) and the hessian.AI
Innovation Lab (funded by the Hessian Ministry for Digital Strategy and Innovation, grant no. SDIW04/0013/003). We are also grateful to all the many people who have supported and enabled this project, including the Telekom Industrial AI Cloud and NVIDIA teams. In particular, we would like to thank Pramod Kumbhar and Oleg Sudakov, whose in-depth expertise and tremendous dedication have been invaluable to our work. We further thank Miroslav Shaltev and Oleh Astappiev from the L3S Research Center for operating the CPU cluster and its Slurm scheduling. We would also like to thank Christian Kotulek, Marek Soha and all their colleagues for their hard work in ensuring that the cluster runs round the clock at full capacity. Finally, we thank Lara Lawniczak, Nora Malke and Synje Jungbehr for their project coordination, organizing project meetings, and keeping the project running smoothly.

\bibliographystyle{abbrv}
\bibliography{main}

\newpage
\appendix

\section{Author Contributions}
\label{app:contributions}

\subsection{Training}
\begin{enumerate}
    \item Pretraining stack development and evaluation: Max Lübbering, Richard Rutmann, Timm Ruland, David Fitzek, Mehdi Ali
    \item Model architecture, training methodology and framework-correctness validation: Timm Ruland, David Fitzek, Max Lübbering, Richard Rutmann
    \item Compute infrastructure, cluster benchmarking and interconnect tuning: David Fitzek, Timm Ruland, Richard Rutmann, Max Lübbering
    \item Distributed-training scaling and memory/throughput optimization: David Fitzek, Timm Ruland, Max Lübbering, Richard Rutmann
    \item Training execution, stability analysis, emergency debugging, framework bug fixes and experiment tracking: Timm Ruland, David Fitzek, Max Lübbering, Richard Rutmann
\end{enumerate}

\subsection{Data}
\begin{enumerate}
    \item Base model data acquisition: Michael Fromm, Alex Jude, Abbas Khan, Ruben Härle, Maurice Kraus, Jan Pfister, Daniil Gurgurov
    \item Pretraining data mixture: Michael Fromm
    \item Data curation infrastructure and experimentation: Michael Fromm, Alex Jude, Abbas Khan, Ruben Härle, Maurice Kraus, Richard Rutmann, Mehdi Ali, Max Lübbering, Maximilian Idahl
    \item Data preprocessing / tokenization pipeline: Richard Rutmann, Max Lübbering, Alex Jude
    \item Mid- and long-context data curation and experimentation: Michael Fromm, Alex Jude, Abbas Khan, Ruben Härle, Maurice Kraus, Sebastian Sztwiertnia, Tom Röhr, Sebastian von Rohrscheidt
\end{enumerate}

\subsection{Evaluation}
\begin{enumerate}
    \item Evaluation methodology and infrastructure: Maximilian Idahl, Benedikt Droste, Alex Jude, Abbas Khan
\end{enumerate}

\subsection{Other}
\begin{enumerate}
    \item Mentorship, advising, program management, and broader strategy: Nicolas Flores-Herr, Simon Gottschalk, Jörg Bienert, Kristian Kersting, Andreas Hotho, Alexander Löser, Wolfgang Nejdl, Simon Ostermann, Jan Plogsties, Björn Plüster, Patrick Putzky
    \item Technical leadership and cross-workstream contributions: Mehdi Ali, Michael Fromm, Max Lübbering, Sebastian Sztwiertnia, Tom Röhr
\end{enumerate}

\section{Detailed Pretraining Data Composition}
\label{app:data-composition}

This appendix gives the full per-source token accounting underlying the mixture flow diagram in Section~\ref{sec:pretraining-data} (\Cref{fig:token-allocation}). Tables~\ref{tab:phase1-sources} and~\ref{tab:phase1-categories} cover Phase~1, Tables~\ref{tab:phase2-sources} and~\ref{tab:phase2-categories} cover Phase~2 (annealing), and Tables~\ref{tab:lc-budget} and~\ref{tab:lc-docs} cover the Phase~3 long-context extension. Rows with zero epochs are sources we enumerated but excluded from training.

{\scriptsize\tabcolsep=0.5pt\begin{longtable}{p{0.425\linewidth}p{0.28\linewidth}rrrr}
\caption{Phase~1 (diverse pretraining) data composition. ``Raw'' is the source
token count in billions; ``Ep.'' is the number of epochs; ``Eff.'' is the
effective token count after epoching; ``Share'' is the percentage of Phase~1
effective tokens. Rows with zero epochs are enumerated but excluded from
training. Raw and effective counts are taken from the source dataset cards (HuggingFace)
and may reflect different tokenizers; they are approximate. Exact tokenizer
counts of consumed tokens appear in Table~\ref{tab:train-stages}.}
\label{tab:phase1-sources}\\
\toprule
\textbf{Source} & \textbf{Subset / Quality} & \textbf{Raw} & \textbf{\phantom{0}Ep.} & \textbf{Eff.} & \textbf{Share} \\
\midrule
\endfirsthead
\multicolumn{6}{c}{{\normalsize\tablename~\thetable{} -- continued}}\\
\toprule
\textbf{Source} & \textbf{Subset / Quality} & \textbf{Raw} & \textbf{\phantom{0}Ep.} & \textbf{Eff.} & \textbf{Share} \\
\midrule
\endhead
\midrule \multicolumn{6}{r}{\textit{continued on next page}}\\
\endfoot
\bottomrule
\endlastfoot

\hf{nvidia/Nemotron-CC-v2.1}{nvidia/Nemotron-CC-v2.1} & High-Quality                         & 26.0   & 3 & 78.0    & 0.3\% \\
\hf{nvidia/Nemotron-CC-v2.1}{nvidia/Nemotron-CC-v2.1} & Medium-High-Quality                  & 16.9   & 1 & 16.9    & 0.1\% \\
\hf{nvidia/Nemotron-CC-v2.1}{nvidia/Nemotron-CC-v2.1} & Medium-Quality                       & 53.5   & 0 & 0.0     & 0.0\% \\
\hf{nvidia/Nemotron-CC-v2.1}{nvidia/Nemotron-CC-v2.1} & High-Quality-Synthetic               & 93.5   & 2 & 187.0   & 0.8\% \\
\hf{nvidia/Nemotron-CC-v2.1}{nvidia/Nemotron-CC-v2.1} & Medium-High-Quality-Synthetic        & 2122.8 & 1 & 2122.8  & 9.2\% \\
\hf{nvidia/Nemotron-CC-v2.1}{nvidia/Nemotron-CC-v2.1} & HQ-Translated-To-English             & 39.6   & 2 & 79.2    & 0.3\% \\
\hf{nvidia/Nemotron-CC-v2.1}{nvidia/Nemotron-CC-v2.1} & MHQ-Translated-To-English            & 26.8   & 1 & 26.8    & 0.1\% \\
\hf{nvidia/Nemotron-CC-v2.1}{nvidia/Nemotron-CC-v2.1} & HQ-Translated-To-English-Synthetic   & 157.8  & 2 & 315.6   & 1.4\% \\
\hf{nvidia/Nemotron-CC-v2.1}{nvidia/Nemotron-CC-v2.1} & High-Quality-DQA                     & 8.0    & 2 & 16.0    & 0.1\% \\
\addlinespace
\multicolumn{4}{l}{\textit{Nemotron-CC-v2.1 subtotal}} & \textit{2842.3} & \textit{12.3\%} \\
\midrule
\hf{nvidia/Nemotron-CC-v2}{nvidia/Nemotron-CC-v2.0} & High-Quality                         & 613.7  & 3 & 1841.1  & 8.0\% \\
\hf{nvidia/Nemotron-CC-v2}{nvidia/Nemotron-CC-v2.0} & Medium-High-Quality                  & 545.6  & 1 & 545.6   & 2.4\% \\
\hf{nvidia/Nemotron-CC-v2}{nvidia/Nemotron-CC-v2.0} & Medium-Quality                       & 2200.8 & 0 & 0.0     & 0.0\% \\
\hf{nvidia/Nemotron-CC-v2}{nvidia/Nemotron-CC-v2.0} & High-Quality-Synthetic               & 1257.0 & 2 & 2514.0  & 10.9\% \\
\hf{nvidia/Nemotron-CC-v2}{nvidia/Nemotron-CC-v2.0} & Diverse QA                           & 692.4  & 1 & 692.4   & 3.0\% \\
\hf{nvidia/Nemotron-CC-v2}{nvidia/Nemotron-CC-v2.0} & Translated-Diverse QA (DE)           & 2.0    & 2 & 4.0     & 0.0\% \\
\addlinespace
\multicolumn{4}{l}{\textit{Nemotron-CC-v2.0 subtotal}} & \textit{5597.1} & \textit{24.3\%} \\
\midrule
\webdataset{https://data.commoncrawl.org/contrib/Nemotron/Nemotron-CC/index.html}{nvidia/Nemotron-CC-v1.0} & High-Quality                         & 553.0  & 3 & 1659.0  & 7.2\% \\
\webdataset{https://data.commoncrawl.org/contrib/Nemotron/Nemotron-CC/index.html}{nvidia/Nemotron-CC-v1.0} & Medium-High-Quality                  & 504.0  & 1 & 504.0   & 2.2\% \\
\webdataset{https://data.commoncrawl.org/contrib/Nemotron/Nemotron-CC/index.html}{nvidia/Nemotron-CC-v1.0} & Medium-Quality                       & 2023.0 & 0 & 0.0     & 0.0\% \\
\webdataset{https://data.commoncrawl.org/contrib/Nemotron/Nemotron-CC/index.html}{nvidia/Nemotron-CC-v1.0} & High-Synthetic-Diverse QA Pairs      & 499.5  & 2 & 999.0   & 4.3\% \\
\addlinespace
\multicolumn{4}{l}{\textit{Nemotron-CC-v1.0 subtotal}} & \textit{3162.0} & \textit{13.7\%} \\
\midrule
\hf{nvidia/Nemotron-CC-Code-v1}{nvidia/Nemotron-CC-Code-v1}         & Actual                   & 427.9  & 3 & 1283.7  & 5.6\% \\
\hf{nvidia/Nemotron-Pretraining-Code-v1}{nvidia/Nemotron-Pretraining-Code-v1}    & Synthetic                & 174.9  & 2 & 349.8   & 1.5\% \\
\hf{nvidia/Nemotron-Pretraining-Code-v1}{nvidia/Nemotron-Pretraining-Code-v1}    & Actual                   & 125.0  & 3 & 375.0   & 1.6\% \\
\hf{nvidia/Nemotron-Pretraining-Code-v2}{nvidia/Nemotron-Pretraining-Code-v2}    & synthetic-code-review    & 71.38  & 2 & 142.76  & 0.6\% \\
\hf{nvidia/Nemotron-Pretraining-Code-v2}{nvidia/Nemotron-Pretraining-Code-v2}  & synthetic-question-answering & 212.52 & 2 & 425.04 & 1.8\% \\
\hf{nvidia/Nemotron-Pretraining-Code-v2}{nvidia/Nemotron-Pretraining-Code-v2}  & synthetic-rewriting      & 76.84  & 2 & 153.68  & 0.7\% \\
\hf{nvidia/Nemotron-Pretraining-Code-v2}{nvidia/Nemotron-Pretraining-Code-v2}  & synthetic-student-teacher& 28.34  & 2 & 56.68   & 0.2\% \\
\hf{nvidia/Nemotron-Pretraining-Code-v2}{nvidia/Nemotron-Pretraining-Code-v2}  & synthetic-transpilation  & 24.09  & 2 & 48.18   & 0.2\% \\
\hf{nvidia/Nemotron-Pretraining-Code-v2}{nvidia/Nemotron-Pretraining-Code-v2}  & Actual                   & 180.0  & 3 & 540.0   & 2.3\% \\
\addlinespace
\multicolumn{4}{l}{\textit{Code subtotal}} & \textit{3374.84} & \textit{14.6\%} \\
\midrule
\hf{nvidia/Nemotron-Pretraining-Specialized-v1}{nvidia/Nemotron-Pretraining-Specialized-v1}& ---              & 270.7  & 5 & 1353.5  & 5.9\% \\
\hf{nvidia/Nemotron-Pretraining-SFT-v1}{nvidia/Nemotron-Pretraining-SFT-v1} & Math SFT                  & 190.6  & 5 & 953.0   & 4.1\% \\
\hf{nvidia/Nemotron-Pretraining-SFT-v1}{nvidia/Nemotron-Pretraining-SFT-v1} & Code SFT                  & 58.5   & 6 & 351.0   & 1.5\% \\
\hf{nvidia/Nemotron-Pretraining-SFT-v1}{nvidia/Nemotron-Pretraining-SFT-v1} & General SFT               & 87.5   & 3 & 262.5   & 1.1\% \\
\addlinespace
\multicolumn{4}{l}{\textit{Specialized + SFT subtotal}} & \textit{2920.0} & \textit{12.7\%} \\
\midrule
\hf{nvidia/Nemotron-CC-Math-v1}{nvidia/Nemotron-CC-Math-v1} & 3plus                            & 133.0  & 4 & 532.0   & 2.3\% \\
\hf{nvidia/Nemotron-CC-Math-v1}{nvidia/Nemotron-CC-Math-v1} & 4plus                            & 52.0   & 4 & 208.0   & 0.9\% \\
\hf{nvidia/Nemotron-CC-Math-v1}{nvidia/Nemotron-CC-Math-v1} & v1                               & 73.0   & 4 & 292.0   & 1.3\% \\
\hf{nvidia/Nemotron-Math-v2}{nvidia/Nemotron-Math-v2} & high                             & 2.7    & 4 & 10.8    & 0.0\% \\
\hf{nvidia/Nemotron-Math-v2}{nvidia/Nemotron-Math-v2} & medium                           & 2.0    & 4 & 8.0     & 0.0\% \\
\hf{nvidia/Nemotron-Math-v2}{nvidia/Nemotron-Math-v2} & low                              & 1.2    & 4 & 4.8     & 0.0\% \\
\hf{openbmb/UltraData-Math}{openbmb/UltraData-Math}     & en                               & 88.0   & 4 & 352.0   & 1.5\% \\
\addlinespace
\multicolumn{4}{l}{\textit{Mathematics subtotal}} & \textit{1407.6} & \textit{6.1\%} \\
\midrule
\hf{MultiSynt/MT-Reasoning}{MultiSynt/MT-Reasoning}     & en                               & 35.8   & 2 & 71.6    & 0.3\% \\
\hf{nvidia/AceReason-1.1-SFT}{nvidia/AceReason-1.1-SFT}   & en                               & 30.0   & 2 & 60.0    & 0.3\% \\
\hf{HuggingFaceFW/finewiki}{HuggingFaceFW/finewiki}    & en                               & 10.0   & 5 & 50.0    & 0.2\% \\
\hf{allenai/dolma3\_pool}{allenai/dolma3\_pool}       & pdfs                             & 600.0  & 1 & 600.0   & 2.6\% \\
\hf{PleIAs/Synth}{PleIAs/Synth}               & en                               & 60.0   & 2 & 120.0   & 0.5\% \\
\hf{HuggingFaceFW/finepdfs}{HuggingFaceFW/finepdfs}   & en                               & 1190.65& 1 & 1190.65 & 5.2\% \\
\addlinespace
\multicolumn{4}{l}{\textit{English (other) subtotal}} & \textit{2092.25} & \textit{9.1\%} \\
\midrule
\hf{HuggingFaceFW/finepdfs}{HuggingFaceFW/finepdfs}     & de                               & 177.56 & 2 & 355.12  & 1.5\% \\
\webdataset{https://hplt-project.org/datasets/v3.0}{HPLT-3-Top10\%}       & de                               & 60.9   & 8.4 & 511.56 & 2.2\% \\
\hf{coral-nlp/german-commons}{coral-nlp/german-commons}            & de                               & 154.56 & 1 & 154.56  & 0.7\% \\
\hf{MultiSynt/MT-Nemotron-CC}{MultiSynt/MT-Nemotron-CC}   & de                               & 117.0  & 2 & 234.0   & 1.0\% \\
\webdataset{https://www.genios.de/browse/Alle}{Genios}                      & de                               & 150.0  & 2 & 300.0   & 1.3\% \\
\hf{HuggingFaceFW/finewiki}{HuggingFaceFW/finewiki}    & de                               & 3.5    & 2 & 7.0     & 0.0\% \\
\hf{MultiSynt/MT-Reasoning}{MultiSynt/MT-Reasoning}     & de                               & 42.0   & 2 & 84.0    & 0.4\% \\
\hf{DGurgurov/Nemotron-Multilingual-Reasoning}{DGurgurov/Nemotron-Multilingual-Reasoning} & de                & 2.0    & 2 & 4.0     & 0.0\% \\
\hf{PleIAs/Synth}{PleIAs/Synth}                  & de                               & 2.4    & 2 & 4.8     & 0.0\% \\
\addlinespace
\multicolumn{4}{l}{\textit{German subtotal}} & \textit{1655.04} & \textit{7.2\%} \\
\midrule
\multicolumn{2}{l}{\textbf{Total}} & \textbf{16{,}350.04} & & \textbf{23{,}051.13} & \textbf{100.0\%} \\
\end{longtable}}

\begin{table}[ht]
\centering
\caption{Phase~1 composition by category, with Nemotron~3~Nano's reported shares
for comparison. These categories cover the full Phase~1 mixture of
$23{,}051.13$B effective tokens ($100\%$).}
\label{tab:phase1-categories}
\begin{tabular}{lrrr}
\toprule
\textbf{Category} & \textbf{Soofi (B)} & \textbf{Soofi} & \textbf{Nemotron 3 Nano} \\
\midrule
Web Medium           & 0.0     & 0.00\%  & 6.8\%  \\
Academic             & 1790.65 & 7.77\%  & 4.1\%  \\
German/Multiling     & 1655.04 & 7.18\%  & 5.0\%  \\
Web High-Q-Synthetic & 4110.8  & 17.83\% & 20.4\% \\
Mathematics          & 1407.6  & 6.11\%  & 6.4\%  \\
Crawl++              & 0.0     & 0.00\%  & 2.9\%  \\
Wiki                 & 50.0    & 0.22\%  & 0.6\%  \\
Web Medium-High      & 1066.5  & 4.63\%  & 5.7\%  \\
Web Med-H-Synthetic  & 2846.0  & 12.35\% & 11.7\% \\
Code-SFT             & 351.0   & 1.52\%  & 3.3\%  \\
Web High             & 3578.1  & 15.52\% & 6.5\%  \\
Code                 & 3374.84 & 14.64\% & 15.3\% \\
General SFT          & 454.1   & 1.97\%  & 0.2\%  \\
STEM SFT             & 2366.5  & 10.27\% & 11.1\% \\
\midrule
\textbf{Total} & \textbf{23{,}051.13} & \textbf{100.0\%} & \textbf{100.0\%} \\
\bottomrule
\end{tabular}
\end{table}

{\footnotesize\tabcolsep=1pt\begin{longtable}{p{0.49\linewidth}p{0.32\linewidth}rrr}
\caption{Phase~2 (high-quality annealing) data composition. Columns as in
Table~\ref{tab:phase1-sources}. Rows with zero epochs are enumerated but excluded
from training. Raw and effective counts are taken from the source dataset cards (HuggingFace)
and may reflect different tokenizers; they are approximate. Exact tokenizer
counts of consumed tokens appear in Table~\ref{tab:train-stages}.}
\label{tab:phase2-sources}\\
\toprule
\textbf{Source} & \textbf{Subset} & \textbf{Raw} & \textbf{\phantom{0}Ep.} & \textbf{Eff.} \\
\midrule
\endfirsthead
\multicolumn{5}{c}{{\normalsize\tablename~\thetable{} -- continued}}\\
\toprule
\textbf{Source} & \textbf{Subset} & \textbf{Raw} & \textbf{\phantom{0}Ep.} & \textbf{Eff.} \\
\midrule
\endhead
\midrule \multicolumn{5}{r}{\textit{continued on next page}}\\
\endfoot
\bottomrule
\endlastfoot

\multicolumn{5}{l}{\textbf{Web}}\\
\hf{nvidia/Nemotron-CC-v2.1}{nvidia/Nemotron-CC-v2.1}  & High-Quality                       & 26.0   & 1   & 26.0   \\
\hf{nvidia/Nemotron-CC-v2.1}{nvidia/Nemotron-CC-v2.1}  & High-Quality-Synthetic             & 93.5   & 1   & 93.5   \\
\hf{nvidia/Nemotron-CC-v2.1}{nvidia/Nemotron-CC-v2.1}  & HQ-Translated-To-English           & 39.6   & 1   & 39.6   \\
\hf{nvidia/Nemotron-CC-v2.1}{nvidia/Nemotron-CC-v2.1}  & HQ-Translated-To-English-Synthetic & 157.8  & 0   & 0.0    \\
\hf{nvidia/Nemotron-CC-v2.1}{nvidia/Nemotron-CC-v2.1}  & High-Quality-DQA                   & 8.0    & 1   & 8.0    \\
\hf{nvidia/Nemotron-CC-v2}{nvidia/Nemotron-CC-v2.0}  & High-Quality                       & 613.7  & 0   & 0.0    \\
\hf{nvidia/Nemotron-CC-v2}{nvidia/Nemotron-CC-v2.0}  & High-Quality-Synthetic             & 1257.0 & 0.5 & 628.5  \\
\hf{nvidia/Nemotron-CC-v2}{nvidia/Nemotron-CC-v2.0}  & Diverse QA                         & 692.4  & 1   & 692.4  \\
\hf{nvidia/Nemotron-CC-v2}{nvidia/Nemotron-CC-v2.0}  & Translated-Diverse QA (DE)         & 2.0    & 0   & 0.0    \\
\webdataset{https://data.commoncrawl.org/contrib/Nemotron/Nemotron-CC/index.html}{nvidia/Nemotron-CC-v1.0}  & High-Quality                       & 553.0  & 0   & 0.0    \\
\webdataset{https://data.commoncrawl.org/contrib/Nemotron/Nemotron-CC/index.html}{nvidia/Nemotron-CC-v1.0}  & High-Synthetic-Diverse QA Pairs    & 499.5  & 0   & 0.0    \\
\hf{allenai/dolma3\_dolmino\_pool}{allenai/dolma3\_dolmino\_pool}        & Web                          & 5.21   & 1   & 5.21   \\
\hf{allenai/dolma3\_dolmino\_pool}{allenai/dolma3\_dolmino\_pool}        & pdfs                         & 240.0  & 1   & 240.0  \\
\hf{karpathy/climbmix-400b-shuffle}{karpathy/climbmix-400b-shuffle}  & en                         & 400.0  & 2   & 800.0  \\
\hf{HuggingFaceFW/finepdfs-edu}{HuggingFaceFW/finepdfs-edu}      & en                             & 142.0  & 1   & 142.0  \\
\addlinespace
\multicolumn{4}{l}{\textit{Web subtotal}} & \textit{2675.21} \\
\midrule
\multicolumn{5}{l}{\textbf{Code}}\\
\hf{nvidia/Nemotron-Pretraining-Code-v1}{nvidia/Nemotron-Pretraining-Code-v1}     & Synthetic              & 174.9  & 1 & 174.9  \\
\hf{nvidia/Nemotron-Pretraining-Code-v1}{nvidia/Nemotron-Pretraining-Code-v1}     & Actual                 & 125.0  & 1 & 125.0  \\
\hf{nvidia/Nemotron-Pretraining-Code-v2}{nvidia/Nemotron-Pretraining-Code-v2}     & synthetic-code-review  & 71.38  & 1 & 71.38  \\
\hf{nvidia/Nemotron-Pretraining-Code-v2}{nvidia/Nemotron-Pretraining-Code-v2}     & synthetic-question-answering & 212.52 & 1 & 212.52 \\
\hf{nvidia/Nemotron-Pretraining-Code-v2}{nvidia/Nemotron-Pretraining-Code-v2}     & synthetic-rewriting    & 76.84  & 1 & 76.84  \\
\hf{nvidia/Nemotron-Pretraining-Code-v2}{nvidia/Nemotron-Pretraining-Code-v2}     & synthetic-student-teacher & 28.34 & 1 & 28.34 \\
\hf{nvidia/Nemotron-Pretraining-Code-v2}{nvidia/Nemotron-Pretraining-Code-v2}     & synthetic-transpilation& 24.09  & 1 & 24.09  \\
\hf{nvidia/Nemotron-Pretraining-Code-v2}{nvidia/Nemotron-Pretraining-Code-v2}     & Actual                 & 180.0  & 1 & 180.0  \\
\hf{tokyotech-llm/swallow-code-v2}{tokyotech-llm/swallow-code-v2}  & stage5                       & 49.8   & 2 & 99.6   \\
\hf{allenai/dolma3\_dolmino\_pool}{allenai/dolma3\_dolmino\_pool} & Code                         & 40.0   & 1 & 40.0   \\
\addlinespace
\multicolumn{4}{l}{\textit{Code subtotal}} & \textit{1032.67} \\
\midrule
\multicolumn{5}{l}{\textbf{Mathematics}}\\
\hf{nvidia/Nemotron-CC-Math-v1}{nvidia/Nemotron-CC-Math-v1}  & 3plus                           & 133.0  & 1   & 133.0  \\
\hf{nvidia/Nemotron-CC-Math-v1}{nvidia/Nemotron-CC-Math-v1}  & 4plus                           & 52.0   & 1   & 52.0   \\
\hf{allenai/dolma3\_dolmino\_pool}{allenai/dolma3\_dolmino\_pool} & Math                         & 21.34  & 1   & 21.34  \\
\hf{openbmb/UltraData-Math}{openbmb/UltraData-Math}      & en                                  & 88.0   & 1.4 & 123.2  \\
\addlinespace
\multicolumn{4}{l}{\textit{Mathematics subtotal}} & \textit{329.54} \\
\midrule
\multicolumn{5}{l}{\textbf{SFT}}\\
\hf{nvidia/Nemotron-Pretraining-SFT-v1}{nvidia/Nemotron-Pretraining-SFT-v1}  & Math SFT                & 190.6  & 2 & 381.2  \\
\hf{nvidia/Nemotron-Pretraining-SFT-v1}{nvidia/Nemotron-Pretraining-SFT-v1}  & Code SFT                & 58.5   & 2 & 117.0  \\
\hf{nvidia/Nemotron-Pretraining-SFT-v1}{nvidia/Nemotron-Pretraining-SFT-v1}  & General SFT             & 87.5   & 1 & 87.5   \\
\hf{nvidia/Nemotron-Agentic-v1}{nvidia/Nemotron-Agentic-v1} & agentic                         & 1.0    & 2 & 2.0    \\
\hf{nvidia/Nemotron-Competitive-Programming-v1}{nvidia/Nemotron-Competitive-Programming-v1} & code-sft        & 50.0   & 2 & 100.0  \\
\hf{nvidia/Nemotron-Instruction-Following-Chat-v1}{nvidia/Nemotron-Instruction-Following-Chat-v1} & General SFT  & 1.5    & 2 & 3.0    \\
\hf{nvidia/Nemotron-Math-Proofs-v1}{nvidia/Nemotron-Math-Proofs-v1} & stem-sft                    & 6.0    & 2 & 12.0   \\
\hf{nvidia/Nemotron-Math-v2}{nvidia/Nemotron-Math-v2} & stem-sft                           & 30.0   & 2 & 60.0   \\
\hf{nvidia/Nemotron-RLHF-GenRM-v1}{nvidia/Nemotron-RLHF-GenRM-v1} & ---                          & 0.5    & 2 & 1.0    \\
\hf{nvidia/Nemotron-Science-v1}{nvidia/Nemotron-Science-v1} & stem-sft                        & 0.5    & 2 & 1.0    \\
\hf{nvidia/Nemotron-SFT-Agentic-v2}{nvidia/Nemotron-SFT-Agentic-v2} & agentic                     & 1.5    & 2 & 3.0    \\
\hf{nvidia/Nemotron-SFT-Competitive-Programming-v2}{nvidia/Nemotron-SFT-Competitive-Programming-v2} & code-sft    & 20.0   & 2 & 40.0   \\
\hf{nvidia/Nemotron-SFT-Instruction-Following-Chat-v2}{nvidia/Nemotron-SFT-Instruction-Following-Chat-v2} & General SFT & 3.0  & 2 & 6.0    \\
\hf{nvidia/Nemotron-SFT-Math-v3}{nvidia/Nemotron-SFT-Math-v3} & stem-sft                       & 1.0    & 2 & 2.0    \\
\hf{nvidia/Nemotron-SFT-Multilingual-v1}{nvidia/Nemotron-SFT-Multilingual-v1} & multilingual-sft       & 3.5    & 2 & 7.0    \\
\hf{nvidia/Nemotron-SFT-OpenCode-v1}{nvidia/Nemotron-SFT-OpenCode-v1} & code-sft                   & 7.0    & 2 & 14.0   \\
\hf{nvidia/Nemotron-SFT-Safety-v1}{nvidia/Nemotron-SFT-Safety-v1} & safety-sft                   & 0.03   & 2 & 0.06   \\
\hf{nvidia/Nemotron-SpecializedDomains-Finance-v1}{nvidia/Nemotron-SpecializedDomains-Finance-v1} & stem-sft     & 4.0    & 2 & 8.0    \\
\hf{nvidia/Nemotron-SWE-v1}{nvidia/Nemotron-SWE-v1} & code-sft                            & 0.7    & 2 & 1.4    \\
\hf{nvidia/Nemotron-SFT-SWE-v2}{nvidia/Nemotron-SFT-SWE-v2} & code-sft                        & 2.5    & 2 & 5.0    \\
\hf{nvidia/AceReason-1.1-SFT}{nvidia/AceReason-1.1-SFT} & en                                & 30.0   & 1 & 30.0   \\
\hf{allenai/dolma3\_dolmino\_pool}{allenai/dolma3\_dolmino\_pool} & QA                           & 25.8   & 1 & 25.8   \\
\hf{allenai/dolma3\_dolmino\_pool}{allenai/dolma3\_dolmino\_pool} & Instruction-Data             & 18.41  & 1 & 18.41  \\
\hf{AIML-TUDA/QA-base}{AIML-TUDA/QA-base} & en                              & 0.143  & 10 & 1.43   \\
\addlinespace
\multicolumn{4}{l}{\textit{SFT subtotal}} & \textit{926.80} \\
\midrule
\multicolumn{5}{l}{\textbf{Reasoning}}\\
\hf{nvidia/Nemotron-Pretraining-Specialized-v1}{nvidia/Nemotron-Pretraining-Specialized-v1} & ---            & 270.7  & 1 & 270.7  \\
\hf{allenai/dolma3\_dolmino\_pool}{allenai/dolma3\_dolmino\_pool} & Thinking                    & 37.6 & 1 & 37.6 \\
\hf{nvidia/Nemotron-Pretraining-Specialized-v1.1}{nvidia/Nemotron-Pretraining-Specialized-v1.1} & en          & 9.3    & 1 & 9.3    \\
\hf{MultiSynt/MT-Reasoning}{MultiSynt/MT-Reasoning} & en                                 & 35.8   & 1 & 35.8   \\
\addlinespace
\multicolumn{4}{l}{\textit{Reasoning subtotal}} & \textit{353.40} \\
\midrule
\multicolumn{5}{l}{\textbf{Wiki}}\\
\hf{HuggingFaceFW/finewiki}{HuggingFaceFW/finewiki} & en                                 & 10.0   & 2  & 20.0   \\
\addlinespace
\multicolumn{4}{l}{\textit{Wiki subtotal}} & \textit{20.00} \\
\midrule
\multicolumn{5}{l}{\textbf{German}}\\
\hf{HuggingFaceFW/finepdfs-edu}{HuggingFaceFW/finepdfs-edu} & de                             & 20.0   & 2  & 40.0   \\
\hf{AIML-TUDA/QA-base}{AIML-TUDA/QA-base} & de                              & 0.187  & 10 & 1.87   \\
\webdataset{https://hplt-project.org/datasets/v4.0}{HPLT-4-Top10\%}  & de                                     & 291.0  & 1  & 291.0  \\
\hf{coral-nlp/german-commons}{coral-nlp/german-commons}             & de                                         & 154.56 & 0  & 0.0    \\
\hf{karpathy/climbmix-400b-shuffle}{German Translation of ClimbMix}            & de                                        & 571.0  & 1  & 571.0  \\
\webdataset{https://www.genios.de/browse/Alle}{Genios}                       & de                                                 & 150.0  & 0  & 0.0    \\
\hf{HuggingFaceFW/finewiki}{HuggingFaceFW/finewiki}     & de                                 & 3.5    & 1  & 3.5    \\
\hf{MultiSynt/MT-Reasoning}{MultiSynt/MT-Reasoning}      & de                                 & 42.0   & 1  & 42.0   \\
\hf{DGurgurov/Nemotron-Multilingual-Reasoning}{DGurgurov/Nemotron-Multilingual-Reasoning}  & de              & 2.0    & 1  & 2.0    \\
\hf{toroe/Soofi-Think-SFT-10B-multilingual}{toroe/Soofi-Think-SFT-10B-multilingual}& de                       & 7.0    & 2  & 14.0   \\
\addlinespace
\multicolumn{4}{l}{\textit{German subtotal}} & \textit{965.37} \\
\midrule
\multicolumn{4}{l}{\textbf{Total}} & \textbf{6{,}303.0} \\
\end{longtable}}

\begin{table}[ht]
\centering
\caption{Phase~2 (annealing) composition by category, with Nemotron~3~Nano's
reported shares for comparison. Percentages are shares of the $6{,}303$\,B
annealing pool. For the seven-category main-text view in
Figure~\ref{fig:token-allocation}, the Reasoning/STEM-SFT bucket is folded into
Reasoning.}\label{tab:phase2-categories}
\begin{tabular}{lrrr}
\toprule
\textbf{Category} & \textbf{Soofi (B)} & \textbf{Soofi} & \textbf{Nemotron 3 Nano} \\
\midrule
Web High              & 831.21  & 13.19\% & 6.50\%  \\
Web Medium-High       & 0.0     & 0.00\%  & 4.00\%  \\
Web High-Q-Synthetic  & 1462.0  & 23.20\% & 20.40\% \\
Web Med-H-Synthetic   & 0.0     & 0.00\%  & 5.00\%  \\
Code                  & 1032.67 & 16.38\% & 14.00\% \\
Code-SFT              & 282.4   & 4.48\%  & 6.70\%  \\
Reasoning / STEM-SFT   & 744.20  & 11.81\% & 22.30\% \\
General-SFT           & 253.60  & 4.02\%  & 0.40\%  \\
Academic / PDFs (High)& 382.0   & 6.06\%  & 2.00\%  \\
Wiki                  & 20.0    & 0.32\%  & 1.30\%  \\
Multilingual          & 965.37  & 15.32\% & 5.00\%  \\
Math                  & 329.54  & 5.23\%  & 12.50\% \\
\midrule
\textbf{Total} & \textbf{6303.0} & \textbf{100.0\%} & \textbf{100.0\%}\\
\bottomrule
\end{tabular}
\end{table}

\begin{table}[t]
\centering
\caption{Long-context phase: per-domain token budgets, document counts, and source
priorities for the released data pool. The run consumed
$\sim$100.66B of the 188.5B pool (\Cref{sec:longcontext}).}
\label{tab:lc-budget}
\begin{tabular}{llrl}
\toprule
Domain & Tokens (B) & Documents & Sources (priority order) \\
\midrule
Web          & 28.78 & 3{,}051{,}984 & \hf{karpathy/climbmix-400b-shuffle}{ClimbMix} $>$ \hf{allenai/dolma3\_dolmino\_pool}{OlmoOCR} $>$ \hf{HuggingFaceFW/finepdfs}{FinePDFs}      \\
Code         &  9.51 &   344{,}625   & \hf{tokyotech-llm/swallow-code-v2}{Swallow-Code-v2} $>$ \hf{nvidia/Nemotron-Pretraining-Code-v1}{Nemotron-Pretraining-Code-v1}, \hf{nvidia/Nemotron-Pretraining-Code-v2}{v2}        \\
Mathematics  &  6.73 & 2{,}526{,}637 & \hf{openbmb/UltraData-Math}{openbmb/UltraData-Math-L3}; no data $>$64K \\
German       &  6.68 &   714{,}989   & 40\% \webdataset{https://hplt-project.org/datasets/v4.0}{HPLT-4-Top10\%} $+$ 60\% \hf{karpathy/climbmix-400b-shuffle}{German Translation of ClimbMix} \\
\midrule
General SFT  & 31.25 & 4{,}386{,}847 & \hf{nvidia/Nemotron-Pretraining-SFT-v1}{nvidia/Nemotron-Pretraining-SFT-v1} \\
Code SFT     & 29.24 & 2{,}541{,}896 & \hf{nvidia/Nemotron-Pretraining-SFT-v1}{nvidia/Nemotron-Pretraining-SFT-v1} \\
Math SFT     & 76.31 & 7{,}990{,}170 & \hf{nvidia/Nemotron-Pretraining-SFT-v1}{nvidia/Nemotron-Pretraining-SFT-v1} \\ 
\midrule
\textbf{Total} & \textbf{188.49} & \textbf{21{,}557{,}148} & \\
\bottomrule
\end{tabular}
 
\vspace{2pt}
\end{table}

\begin{table}[t]
\centering
\small
\caption{Long-context phase: number of documents per sequence-length bucket and
domain in the released pool. ``--'' denotes a bucket left unpopulated. Unlike the idealized symmetric schema, the
realized counts do not halve cleanly across buckets.}
\label{tab:lc-docs}
\setlength{\tabcolsep}{4pt}
\begin{tabular}{lrrrrrrr}
\toprule
Seq.\ len. & Web & Code & Math & German & Code SFT & Gen SFT & Math SFT \\
\midrule
4K   & 1{,}535{,}801 & --          & 1{,}781{,}457 &   367{,}330 & --          & --          & --          \\
8K   &   765{,}057 &           3 &   598{,}808 &   178{,}417 & 1{,}002{,}824 & 3{,}307{,}656 & 4{,}091{,}855 \\
16K  &   386{,}188 &   173{,}557 &   123{,}606 &    87{,}248 &   969{,}107 &   851{,}104 & 2{,}991{,}068 \\
32K  &   198{,}283 &    86{,}572 &    22{,}347 &    46{,}283 &   544{,}057 &   225{,}083 &   842{,}927 \\
64K  &    99{,}484 &    43{,}424 &       417 &    23{,}253 &    25{,}908 &     3{,}004 &    64{,}299 \\
128K &    49{,}102 &    22{,}099 &         2 &     9{,}867 & --          & --          &        19 \\
256K & --          &    10{,}840 & --          & --          & --          & --          &         2 \\
512K &    12{,}046 &     5{,}420 & --          &     2{,}523 & --          & --          & --          \\
1M   &     6{,}023 &     2{,}710 & --          &        68 & --          & --          & --          \\
\midrule
\textbf{Total} & 3{,}051{,}984 & 344{,}625 & 2{,}526{,}637 & 714{,}989 & 2{,}541{,}896 & 4{,}386{,}847 & 7{,}990{,}170 \\
\bottomrule
\end{tabular}
\end{table}

\section{Proxy Data-Mixture Ablations}
\label{app:data-ablations}

Before fixing the data recipe used for \ourmodel{}, we ran controlled
small-scale data-mixture ablations. These runs were designed as a proxy study:
their role was to select a robust German--English data family under a much
cheaper setup, not to predict the absolute performance of the final 30B-A3B
hybrid-Mamba MoE model. To make the comparison realistic for the final setting,
we held the English backbone fixed and changed only the German or multilingual
part of the mixture. English therefore remained the dominant language in every
candidate, contributing roughly 90--93\% of the 100B-token proxy budget, while
the remaining budget tested different allocations among German web, PDF,
synthetic, commons, news, and lightly multilingual sources. This isolates the
main design question for \ourmodel{}: how to spend the scarce non-English budget
without confounding the result with changes in English data.

\paragraph{Ablated mixtures.}
The English backbone follows a fixed Nemotron-style composition of web, code,
math, SFT, and PDF sources~\cite{nvidia2025nemotron3}. The ablation identifiers
\textsc{D01}, \textsc{D03}--\textsc{D13} correspond to the non-English mixtures
summarized in \Cref{tab:data-ablation-mixtures}.

The candidates cover several qualitatively different hypotheses. \textsc{D01}
combines all German source families with a strong HPLT-3~\cite{oepen2025hplt3} web anchor and small
amounts of German-Commons~\cite{gienapp2025germancommons} and Genios~\cite{genios2025}. \textsc{D12} uses the same high-level
shares but swaps the German HPLT quality classifier for the Propella-filtered variant~\cite{idahl2026propella1multipropertydocumentannotation}.
\textsc{D04} is the only explicitly multilingual candidate, adding Spanish,
Italian, and French HPLT-3 top-decile data to the otherwise German tail.
\textsc{D08} tests an almost pure German FinePDFs~\cite{finepdfs2025} tail, \textsc{D09} and
\textsc{D10} emphasize MultiSynt~\cite{idahl2026multisyntmttrilliontokenmultiparallelpretraining}, and \textsc{D11} tests a broader HPLT-3
German filter by moving from the top decile to the top two deciles.

\begin{table}[t]
\centering
\scriptsize
\setlength{\tabcolsep}{3.2pt}
\caption{Non-English mixture shares for the evaluated proxy ablations. Entries
are percentages of the 100B-token proxy budget recorded in the ablation sheet.
``0.1q'' denotes the top 10\% HPLT-3 filter, and ``0.2q'' denotes the top 20\%
filter. The ``Other HPLT'' column is non-German HPLT-3 data (Spanish, Italian,
and French) and is nonzero only for the explicitly multilingual ablation
\textsc{D04}.}
\label{tab:data-ablation-mixtures}
\resizebox{\linewidth}{!}{%
\begin{tabular}{llrrrrrrr}
\toprule
\textbf{Run} &  \textbf{HPLT-DE} & \textbf{HPLT-DE} & \textbf{FinePDFs-DE} &
\textbf{MultiSynt-DE} & \textbf{German-Commons} & \textbf{Genios} &
\textbf{Other HPLT} & \textbf{Non-EN total} \\
 & \textbf{filter} & \textbf{\%} & \textbf{\%} & \textbf{\%} &
\textbf{\%} & \textbf{\%} & \textbf{\%} & \textbf{\%} \\
\midrule
\textsc{D01} & 0.1q & 3.04 & 1.77 & 0.93 & 0.72 & 0.80 & 0.00 & 7.26 \\
\textsc{D03} & 0.1q & 3.04 & 1.77 & 2.33 & 0.00 & 0.00 & 0.00 & 7.14 \\
\textsc{D04} & 0.1q & 3.04 & 1.71 & 1.99 & 0.00 & 0.00 & 2.96 & 9.71 \\
\textsc{D05} & 0.1q & 0.30 & 1.77 & 1.40 & 2.16 & 1.20 & 0.00 & 6.83 \\
\textsc{D06} & 0.1q & 1.52 & 3.54 & 1.86 & 0.00 & 0.00 & 0.00 & 6.92 \\
\textsc{D07} & 0.1q & 0.30 & 5.31 & 1.40 & 0.00 & 0.00 & 0.00 & 7.01 \\
\textsc{D08} & 0.1q & 0.00 & 7.08 & 0.00 & 0.00 & 0.00 & 0.00 & 7.08 \\
\textsc{D09} & 0.1q & 0.61 & 1.77 & 4.65 & 0.00 & 0.00 & 0.00 & 7.03 \\
\textsc{D10} & 0.1q & 1.52 & 2.66 & 4.65 & 4.32 & 0.00 & 0.00 & 13.15 \\
\textsc{D11} & 0.2q & 6.09 & 0.00 & 0.93 & 0.00 & 0.00 & 0.00 & 7.02 \\
\textsc{D12} & 0.1q Propella & 3.04 & 1.77 & 0.93 & 0.72 & 0.80 & 0.00 & 7.26 \\
\textsc{D13} & 0.1q & 0.91 & 4.42 & 1.86 & 0.00 & 0.00 & 0.00 & 7.20 \\
\bottomrule
\end{tabular}%
}
\end{table}

\paragraph{Proxy training setup.}
All ablations used the same training configuration and differed only in the
chosen tokenized dataset blend. The proxy model was based on the Qwen3-1.7B
pretraining recipe~\cite{qwen3_2025}. We trained with bf16 mixed precision,
global batch size 256, and micro-batch size 1. The runs were launched on four
Leonardo nodes with four GPUs per node. Checkpoints were written during training and evaluated offline with
the \texttt{lm-evaluation-harness}~\cite{gao2024lmeval} pipeline. We used a training batch size of
2{,}097{,}152 tokens per optimizer step, and the final comparable checkpoint at
step 47{,}683 corresponds to ${\sim}100B$ trained tokens. Checkpoints were evaluated at approximately 10B-token
intervals, giving a learning curve rather than a single endpoint for each
candidate mixture.

\paragraph{Evaluation protocol.}
For mixture selection we used the same English--German harness family as in the
main report. The primary scalar criterion was the average rank over four
suite-level signals: English bits-per-byte, German bits-per-byte, English
normalized rank-choice accuracy, and German normalized rank-choice accuracy.
Bits-per-byte is lower-better; normalized rank-choice accuracy is higher-better.
We preferred ranks over raw-score averaging because the four metrics have
different scales. All terminal averages in Table~\ref{tab:data-ablation-summary}
are computed using only the step-47{,}683 checkpoint. We explicitly ignore the
near-duplicate step-47{,}680 endpoint so that the final evaluation is not
double-counted.

\begin{table}[t]
\centering
\small
\setlength{\tabcolsep}{5.0pt}
\caption{Proxy data-mixture ablations at the terminal 100B-token checkpoint,
computed using only step 47{,}683. ``EN/DE bpb'' are suite-level bits-per-byte
scores (lower is better); ``EN/DE acc.'' are normalized rank-choice accuracies
in percent (higher is better). $R_{100\mathrm{B}}$ is the average rank over
these four terminal suite metrics. Best values are bolded and second-best
values are underlined.}
\label{tab:data-ablation-summary}
\begin{tabular}{lrrrrr}
\toprule
\textbf{Run} & \textbf{EN bpb $\downarrow$} & \textbf{DE bpb $\downarrow$} &
\textbf{EN acc. $\uparrow$} & \textbf{DE acc. $\uparrow$} &
\textbf{$R_{100\mathrm{B}}\downarrow$} \\
\midrule
\textsc{D01} & \textbf{0.800} & 0.796 & \underline{46.8} & \textbf{67.4} & \textbf{1.75} \\
\textsc{D12} & 0.803 & 0.799 & \textbf{46.8} & \underline{67.3} & \underline{3.25} \\
\textsc{D04} & 0.806 & \underline{0.795} & 46.5 & 67.2 & 4.50 \\
\textsc{D09} & 0.805 & 0.796 & 46.6 & 66.4 & 5.00 \\
\textsc{D13} & \underline{0.803} & 0.800 & 46.6 & 66.6 & 5.50 \\
\textsc{D06} & 0.807 & \textbf{0.791} & 46.3 & 66.7 & 6.00 \\
\textsc{D03} & 0.810 & 0.797 & 46.5 & 67.1 & 6.50 \\
\textsc{D11} & 0.808 & 0.807 & 46.4 & 67.1 & 7.50 \\
\textsc{D08} & 0.805 & 0.851 & 46.6 & 65.5 & 8.50 \\
\textsc{D10} & 0.815 & 0.797 & 46.3 & 66.9 & 8.50 \\
\textsc{D07} & 0.814 & 0.800 & 46.1 & 66.4 & 10.50 \\
\textsc{D05} & 0.809 & 0.812 & 46.2 & 66.1 & 10.50 \\
\bottomrule
\end{tabular}
\end{table}

\paragraph{Results.}
The rank-based selection criterion gives a clear winner. \textsc{D01} has the
best terminal average rank ($R_{100\mathrm{B}}=1.75$) and the best average rank
across the full training trace ($R_{\mathrm{all}}=3.55$). At the final
checkpoint it is best on English bits-per-byte and German normalized accuracy,
second on English normalized accuracy, and close to the strongest group on
German bits-per-byte. This balance matters because several alternatives win a
single metric but are less stable overall. \textsc{D06}, for example, achieves
the lowest German bits-per-byte, but its average rank is only sixth at the final
checkpoint and seventh over the full trace. \textsc{D12} is the strongest
runner-up and is slightly better on English normalized accuracy, but it gives
back performance on both bits-per-byte suites and on German normalized accuracy
relative to \textsc{D01}.

The source-level patterns are also informative. A mixture consisting almost
entirely of German FinePDFs (\textsc{D08}) performs competitively on English but
collapses on German bits-per-byte, suggesting that document-style PDF text alone
is too narrow for the German tail. Heavy MultiSynt mixtures improve some German
likelihood tasks but do not produce the best bilingual aggregate
(\textsc{D09}, \textsc{D10}). The explicitly multilingual mixture
(\textsc{D04}) is a strong candidate and obtains the second-best German
bits-per-byte score, but it does not match \textsc{D01} on the combined
English--German selection criterion. The comparison between \textsc{D01} and
\textsc{D12} further suggests that the original HPLT-DE component is preferable
to the Propella-filtered substitute under this proxy setup, even when the
high-level source shares are otherwise unchanged.

Task-group aggregates support the same conclusion. \textsc{D01} is best on the
English likelihood suite, best on the English math and English QA
bits-per-byte aggregates, and best on the German code bits-per-byte and German
grammar-fluency aggregates. Its weaker German QA and German math likelihood
scores are offset by stronger German rank-choice and fluency performance, which
better matches the intended downstream profile of a German--English base model.

\paragraph{Selection for the full run.}
We therefore used the mixture represented by \textsc{D01} as the basis for the
full \ourmodel{} data recipe. The proxy result was not copied mechanically into
the final 26.68T-token curriculum: the final training plan still separates broad
pretraining from high-quality annealing, up-weights German in both phases, and
adds the long-context extension described in Section~\ref{sec:phase3-hparams}.
Nevertheless, the ablations provide the empirical justification for selecting a
balanced German data family---HPLT-DE, German FinePDFs, MultiSynt,
German-Commons, and Genios---rather than optimizing a single benchmark, a single
German source, or a broad multilingual tail in isolation.

\section{Further Dataset Information}
\label{app:genios}
 
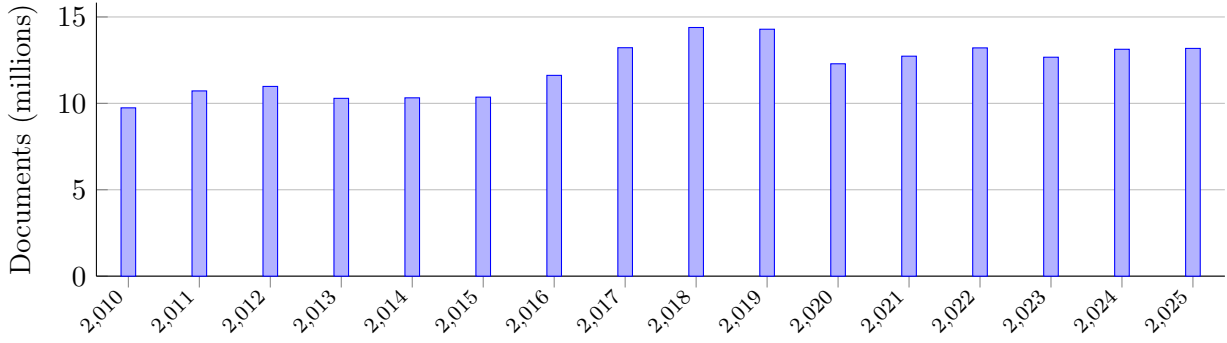
\begin{figure}[t]
\centering
\begin{tikzpicture}
\begin{axis}[
    ybar,
    width=\linewidth, height=5.2cm,
    bar width=5.5pt,
    ymin=0,
    ylabel={Documents (millions)},
    xtick=data,
    x tick label style={rotate=45, anchor=east, font=\scriptsize},
    ytick={0,5,10,15},
    ymajorgrids,
    axis lines*=left,
    enlarge x limits=0.03,
]
\addplot coordinates {
(2010, 9.74) (2011,10.72) (2012,10.98) (2013,10.29)
(2014,10.32) (2015,10.36) (2016,11.62) (2017,13.22)
(2018,14.39) (2019,14.29) (2020,12.29) (2021,12.73)
(2022,13.21) (2023,12.67) (2024,13.13) (2025,13.18)
};
\end{axis}
\end{tikzpicture}
\caption{Temporal distribution of the 193.1M Genios
articles~\cite{genios2025}. Coverage is roughly uniform across
2010--2025 (9.7--14.4M documents per year).}
\label{fig:genios-years}
\end{figure}

\paragraph{Genios.} The Genios corpus~\cite{genios2025} is a commercially
licensed collection of German-language newspaper and trade-press archives
obtained from GBI-Genios. It comprises 916 distinct publications with
193.1M articles ($\sim$57.6B words) spanning 2010--2025, with per-year
volumes between 9.7M and 14.4M documents
(Figure~\ref{fig:genios-years}). The collection is dominated by regional
daily newspapers (e.g.\ \emph{Rheinische Post}, \emph{Rhein-Zeitung},
\emph{Neue Westf\"alische}), complemented by a long tail of national
outlets and specialist trade and academic periodicals
(Table~\ref{tab:genios}). Articles average $\sim$298 words. Because the
data was purchased under a commercial license, it cannot be
redistributed; we therefore report aggregate corpus statistics only.
 
\begin{table}[t]
\centering
\small
\begin{tabular}{@{}lrrl@{}}
\toprule
\textbf{Source} & \textbf{Docs (M)} & \textbf{Words/doc} & \textbf{Years} \\
\midrule
Rheinische Post            & 5.48 & 220 & 2010--2025 \\
Rhein-Zeitung              & 5.32 & 242 & 2010--2025 \\
Neue Westf\"alische        & 5.05 & 208 & 2010--2025 \\
Schw\"abische Zeitung      & 3.15 & 272 & 2011--2025 \\
M\"arkische Allgemeine     & 3.15 & 223 & 2010--2025 \\
Th\"uringer Allgemeine     & 3.07 & 250 & 2010--2025 \\
Passauer Neue Presse       & 2.97 & 303 & 2010--2025 \\
Mitteldeutsche Zeitung     & 2.65 & 309 & 2010--2025 \\
Ostth\"uringer Zeitung     & 2.64 & 259 & 2010--2025 \\
S\"achsische Zeitung       & 2.32 & 312 & 2010--2025 \\
\midrule
Other (906 sources)        & 157.3 & --- & 2010--2025 \\
\midrule
\textbf{Total (916 sources)} & \textbf{193.1} & \textbf{298 (avg.)} & \textbf{2010--2025} \\
\bottomrule
\end{tabular}
\caption{Composition of the Genios corpus~\cite{genios2025}: the ten
largest publications by document count, out of 916 German newspaper and
trade-press sources totalling 193.1M articles and $\sim$57.6B words.}
\label{tab:genios}
\end{table}
 
\section{Further Base Model Evaluations}
\paragraph{Long-context.}\label{para:long-context}

\begin{figure}[ht]
\centering
\includegraphics[width=\textwidth]{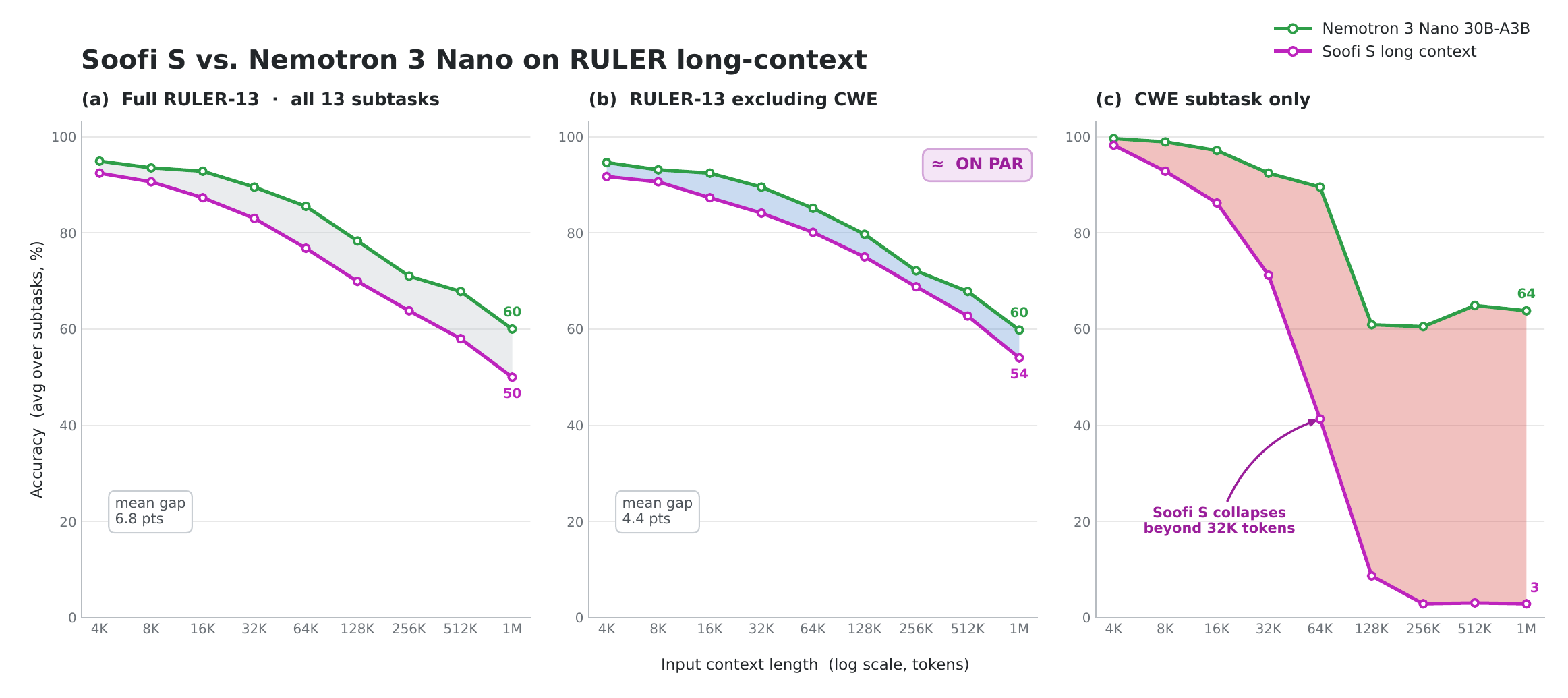}
\caption{RULER accuracy averaged over the selected subtasks, by input context length (4K--1M). (a) all 13 subtasks; (b) all subtasks except CWE;
(c) CWE (common-word extraction) only, where Soofi S degrades sharply past 32K. Shaded band = accuracy gap between the two models.}
\label{fig:ruler}
\end{figure}

We evaluate long-context behaviour with RULER~\cite{hsieh2024ruler} on the
checkpoint after the long-context stage (Section~\ref{sec:phase3-hparams}),
comparing against Nemotron~3~Nano~30B-A3B. Because the two models share an
identical backbone, any difference in long-context accuracy isolates the effect
of the long-context \emph{data} and continuation recipe rather than the
architecture. Figure~\ref{fig:ruler} reports accuracy averaged over RULER
subtasks as a function of input length from 4K to 1M tokens.
 
On the full 13-subtask suite \ourmodel{} trails the reference by a mean of
$6.8$ points across lengths, reaching $50$ versus $60$ at 1M
(Figure~\ref{fig:ruler}a). The gap is almost entirely attributable to a single
subtask, common-word extraction (CWE), which requires aggregating and
reproducing the most frequently occurring words across the entire input rather
than retrieving a localized span. Excluding CWE, the two models track each other
to within a mean of $4.4$ points across all lengths and to within ${\sim}6$
points at 1M ($54$ versus $60$), i.e.\ effectively on par out to the full
context window (Figure~\ref{fig:ruler}b). On CWE itself \ourmodel{} matches the
reference up to 32K but degrades sharply at longer inputs (Figure~\ref{fig:ruler}c),
falling to ${\sim}3\%$ at 256K--1M while Nemotron retains $60$--$64\%$.
 
We attribute this regression to the long-context data mixture rather than the
backbone, for two reasons. First, the two models are architecturally identical,
so the fixed-size Mamba-2 recurrent state and the $6$-of-$52$ attention layers
cannot by themselves explain a gap against the same architecture trained on a
different long-context blend. Second, the reference recipe deliberately targets
this exact class of task: the Nemotron long-context phase devotes its blend to
long-context document-QA data (scaled $3\times$ over the previous generation)
together with a dedicated slice of synthetic \emph{retrieval-focused} data at up
to 256K tokens, added specifically to improve RULER-style subtasks, with the
remainder being down-weighted high-quality pretraining
data~\cite{nvidia2025nemotron3}. Our long-context pool
(Section~\ref{sec:longcontext}), by contrast, is dominated by length-bucketed
SFT and general document text sampled uniformly, and contains no dedicated
retrieval- or aggregation-oriented long-context data. The two runs consume a
comparable long-context token budget (${\sim}100.66$B here versus
${\sim}121$B for the reference~\cite{nvidia2025nemotron3}), so the difference
reflects mixture composition rather than long-context training volume. We flag
CWE beyond 32K as a known limitation and a concrete target for the next
iteration of the long-context data pipeline, specifically, adding retrieval-
and aggregation-style synthetic data in the 32K--1M range, and note that, this
single subtask aside, \ourmodel{} retains near-parity with its
architecture-matched reference across the full 1M-token range.

\section{Checkpoint Merging Ablations}
\label{app:checkpoint-merging}

In addition to selecting a single late-stage checkpoint, we evaluated a series
of post-hoc checkpoint merges over the annealing trajectory. These experiments
were intended to test whether weight-space averaging could reduce checkpoint
noise at the end of training and improve robustness without adding any new
training tokens. In all cases, checkpoints came from the same training run or
from direct continuations of it, so the models had identical architecture,
tokenizer, tensor layout, and optimizer history. The merges therefore average
model weights only; optimizer states were not included.

The relevant baseline for this comparison is \texttt{iter\_1056000}, the
checkpoint selected for the Section~\ref{sec:base-evals} base-model evaluation
and used to initialize the long-context extension. This checkpoint is the model
we would have released without any post-hoc merge. Later final-annealing
checkpoints and their merges are therefore treated as ablations against this
selected checkpoint, not as replacements by default.

\paragraph{Merge implementation.}
All merges were performed in the original distributed-checkpoint layout with a
shared reference checkpoint. Floating-point tensors were accumulated in
\texttt{float32}, while non-floating tensors were copied from the reference
checkpoint. We considered three families of merge weights. First, uniform
averaging assigns equal weight to each checkpoint in a window. Second,
exponential averaging assigns larger weight to later checkpoints, with decay
values $\alpha \in \{0.2, 0.8\}$; larger $\alpha$ keeps the merge closer to the
end of the trajectory. Third, two-checkpoint manual blends mix the penultimate
or earlier endpoint with the final checkpoint using either $0.2/0.8$ or
$0.1/0.9$ weights. All manual weights were normalized and constrained to be
non-negative.

\newcommand{\ub}{\_\allowbreak}  %

\begin{table}[t]
\centering
\small
\setlength{\tabcolsep}{4pt}
\caption{Checkpoint-merge strategies evaluated after annealing. ``Reference''
is the checkpoint whose metadata and non-floating tensors were used for the
merged export.}
\label{tab:merge-strategies}
\begin{tabular}{>{\raggedright\arraybackslash}p{0.30\linewidth}
                >{\raggedright\arraybackslash}p{0.30\linewidth}
                p{0.17\linewidth}
                >{\raggedright\arraybackslash}p{0.15\linewidth}}
\toprule
\textbf{Family} & \textbf{Window / checkpoints} & \textbf{Variants} & \textbf{Reference} \\
\midrule
\texttt{annealing\ub last\ub 1t\ub *} &
12 checkpoints from \texttt{iter\_0952000} through \texttt{iter\_0993410} in
\texttt{the-big-run-4}. &
uniform, exp. $\alpha=0.2$, exp. $\alpha=0.8$ &
\texttt{iter\_0993410} \\
\texttt{annealing\ub last\ub 2\ub *} &
Two-checkpoint tail blend of \texttt{iter\_0992000} and
\texttt{iter\_0993410}. &
$0.2/0.8$, $0.1/0.9$ &
\texttt{iter\_0993410} \\
\texttt{more\ub annealing\ub last\ub 1t\ub *} &
16 checkpoints from \texttt{iter\_0996000} through \texttt{iter\_1056000} in
the \texttt{more/checkpoints} continuation. &
uniform, exp. $\alpha=0.2$, exp. $\alpha=0.8$ &
\texttt{iter\_1056000} \\
\texttt{more\ub annealing\ub last\ub 2\ub *} &
Two-checkpoint tail blend of \texttt{iter\_1052000} and
\texttt{iter\_1056000}. &
$0.2/0.8$, $0.1/0.9$ &
\texttt{iter\_1056000} \\
\texttt{final\ub annealing\ub all\ub 1t\ub *} &
13 checkpoints from \texttt{iter\_1056000\_original} through
\texttt{iter\_1067920}. &
uniform, exp. $\alpha=0.2$, exp. $\alpha=0.8$ &
\texttt{iter\_1067920} \\
\texttt{final\ub annealing\ub all\ub 2\ub *} &
Two-checkpoint tail blend of \texttt{iter\_1067000} and
\texttt{iter\_1067920}. &
$0.2/0.8$, $0.1/0.9$ &
\texttt{iter\_1067920} \\
\texttt{final\ub annealing\ub div4000\ub plus\ub last\ub 1t\ub *} &
Sparse final-window merge over \texttt{iter\_1056000\_original},
\texttt{iter\_1060000}, \texttt{iter\_1064000}, and \texttt{iter\_1067920}. &
uniform, exp. $\alpha=0.2$, exp. $\alpha=0.5$, exp. $\alpha=0.8$ &
\texttt{iter\_1067920} \\
\texttt{final\ub annealing\ub div4000\ub plus\ub last\ub 2\ub *} &
Two-checkpoint endpoint blend of \texttt{iter\_1064000} and
\texttt{iter\_1067920}. &
$0.2/0.8$, $0.1/0.9$ &
\texttt{iter\_1067920} \\
\texttt{top3\ub uniform} &
Uniform average of three evaluation-selected final checkpoints:
\texttt{iter\_1056000\_original}, \texttt{iter\_1060000}, and
\texttt{iter\_1064000}. &
uniform &
\texttt{iter\_1064000} \\
\bottomrule
\end{tabular}
\end{table}

\paragraph{Evaluation.}
We evaluated 22 merged checkpoints with the same benchmark harness used for
late-stage checkpoint selection and compared them directly to the selected
\texttt{iter\_1056000} checkpoint. For the main comparison we used four
aggregate suite metrics: English bits-per-byte, German bits-per-byte, English
normalized rank-choice accuracy, and German normalized rank-choice accuracy.
Bits-per-byte metrics are lower-better, while normalized accuracies are
higher-better. As in the data-mixture ablations, we summarize the trade-off
using an average rank over the four aggregate metrics rather than averaging raw
scores with different scales. One two-checkpoint final merge had incomplete
bits-per-byte coverage in the filtered evaluation file and was therefore
excluded from the rank table.

\begin{table}[t]
\centering
\small
\setlength{\tabcolsep}{4pt}
\caption{Checkpoint merges compared against the selected base checkpoint used
for Section~\ref{sec:base-evals} and long-context initialization. Accuracies
are reported as percentages. $R$ is the average rank over English bpb, German
bpb, English normalized accuracy, and German normalized accuracy, computed over
the selected checkpoint and all complete merge evaluations.}
\label{tab:merge-results}
\begin{tabular}{p{0.43\linewidth}rrrrr}
\toprule
\textbf{Candidate} & \textbf{EN bpb $\downarrow$} & \textbf{DE bpb $\downarrow$} &
\textbf{EN acc. $\uparrow$} & \textbf{DE acc. $\uparrow$} &
\textbf{$R \downarrow$} \\
\midrule
\texttt{final\_annealing\_all\_1t\_uniform} & 0.4396 & \textbf{0.3637} & 77.28 & \textbf{80.34} & \textbf{3.00} \\
\texttt{final\_annealing\_all\_1t\_exp\_avg\_0.8} & 0.4394 & \textbf{0.3637} & 77.25 & \textbf{80.32} & \textbf{3.25} \\
\texttt{final\_annealing\_all\_2\_0.9} & 0.4396 & \textbf{0.3637} & 77.22 & 80.30 & 5.75 \\
\texttt{more\_annealing\_last\_1t\_exp\_avg\_0.8} & 0.4397 & 0.3641 & 77.37 & 80.27 & 6.75 \\
\texttt{top3\_uniform} & 0.4396 & 0.3644 & 77.29 & 80.24 & 7.50 \\
\texttt{final\_annealing\_div4000\_plus\_last\_2\_0.8} & 0.4396 & 0.3638 & 77.19 & 80.26 & 7.75 \\
\texttt{final\_annealing\_all\_1t\_exp\_avg\_0.2} & 0.4395 & 0.3639 & 77.24 & 80.13 & 8.75 \\
\textbf{Selected checkpoint }\texttt{iter\_1056000} & \textbf{0.4390} & 0.3656 & \textbf{77.39} & 80.10 & 9.25 \\
\texttt{final\_annealing\_div4000\_plus\_last\_2\_0.9} & 0.4397 & \textbf{0.3637} & 77.11 & 80.26 & 9.25 \\
\bottomrule
\end{tabular}
\end{table}

\paragraph{Outcome.}
The merge ablation did not reveal a uniformly better model than the selected
\texttt{iter\ub 1056000} checkpoint. The best merged variants were concentrated
in the final annealing window: the uniform average over all 13 final-window
checkpoints had the best aggregate rank, and the late-biased exponential average
with $\alpha=0.8$ was nearly tied. These merges slightly improved the German
suite metrics relative to \texttt{iter\_1056000}: German bits-per-byte improved
from $0.3656$ to $0.3637$, and German normalized accuracy improved from
$80.10$ to $80.34$. However, the selected checkpoint remained better on the
English suite metrics, with English bits-per-byte $0.4390$ and English
normalized accuracy $77.39$, compared with $0.4396$ and $77.28$ for the best
uniform merge.
This trade-off explains why we did not treat checkpoint merging as a decisive
source of additional capability. The final-window merges provide useful evidence
that the end of annealing lies in a stable basin and that small German-side gains
are possible through weight averaging. At the same time, those gains come with
small regressions on the English aggregate metrics and do not clearly dominate
the checkpoint already used for the long-context continuation and the main
Section~\ref{sec:base-evals} evaluation. We therefore report the merge results
for transparency, but keep \texttt{iter\_1056000} as the primary selected base
checkpoint.
\end{document}